\documentclass{article}

\usepackage{arxiv}

\usepackage[utf8]{inputenc} 
\usepackage[T1]{fontenc}    
\usepackage{hyperref}       
\usepackage{url}            
\usepackage{booktabs}       
\usepackage{amsfonts}       
\usepackage{nicefrac}       
\usepackage{microtype}      
\usepackage{subcaption}
\usepackage{enumitem}
\usepackage{lipsum}
\usepackage{amsmath}
\usepackage{cleveref}
\usepackage{algorithm}
\usepackage{algorithmicx}
\usepackage{algpseudocode}
\usepackage{graphicx}
\usepackage{xurl}
\usepackage{multirow}
\usepackage{longtable}
\usepackage{makecell}
\graphicspath{ {./images/} }

\title{VR-FuseNet: A Fusion of Heterogeneous Fundus Data and Explainable Deep Network for Diabetic Retinopathy Classification}

\author{
 Shamim Rahim Refat \\
  Software Engineer (AI/ML)\\
  Technohaven Company Ltd.\\
  Dhaka, Bangladesh\\
  \texttt{n.a.refat2000@gmail.com}\\
  \And
 Ziyan Shirin Raha\\
  Department of Computer Science and Engineering\\
  Ahsanullah University of Science and Technology\\
  Dhaka, Bangladesh\\
  \texttt{ziyanraha@gmail.com}\\
  \And
 Shuvashis Sarker\\
  Department of Computer Science and Engineering\\
  Ahsanullah University of Science and Technology\\
  Dhaka, Bangladesh\\
  \texttt{shuvashisofficial@gmail.com}\\
  \And
 Faika Fairuj Preotee\\
  Department of Computer Science and Engineering\\
  Ahsanullah University of Science and Technology\\
  Dhaka, Bangladesh\\
  \texttt{faikafairuj2001@gmail.com}\\
  \And
 MD. Musfikur Rahman\\
  Department of Computer Science and Engineering\\
  Ahsanullah University of Science and Technology\\
  Dhaka, Bangladesh\\
  \texttt{mush9031@gmail.com}\\
  \And
 Tashreef Muhammad \\
  Department of Computer Science and Engineering\\
  Southeast University\\
  Dhaka, Bangladesh\\
  \texttt{tashreef.muhammad@seu.edu.bd}\\
  \And
 Mohammad Shafiul Alam \\
  Department of Computer Science and Engineering\\
  Ahsanullah University of Science and Technology\\
  Dhaka, Bangladesh\\
  \texttt{shafiul.cse@aust.edu}\\
}

\begin{document}
\maketitle

\begin{abstract}
Diabetic retinopathy is a severe eye condition caused by diabetes where the retinal blood vessels get damaged and can lead to vision loss and blindness if not treated. Early and accurate detection is key to intervention and stopping the disease progressing. For addressing this disease properly, this paper presents a comprehensive approach for automated diabetic retinopathy (DR) detection by proposing a new hybrid deep learning model called VR-FuseNet. Diabetic retinopathy is a major eye disease and leading cause of blindness, especially among diabetic patients, so accurate and efficient automated detection methods are required. To address the limitations of existing methods including dataset imbalance, diversity and generalization issues this paper presents a hybrid dataset created from five publicly available diabetic retinopathy datasets: APTOS 2019, DDR, IDRiD, Messidor 2, and Retino. Essential preprocessing techniques such as Synthetic Minority Over-sampling Technique (SMOTE) for class balancing and Contrast Limited Adaptive Histogram Equalization (CLAHE) for image enhancement are applied systematically to the dataset to improve the robustness and generalizability of the dataset. The proposed VR-FuseNet model combines the strengths of two state-of-the-art convolutional neural networks—VGG19 which captures fine-grained spatial features and ResNet50V2 which is known for its deep hierarchical feature extraction. This fusion improves the diagnostic performance and achieves an accuracy of 91.824\%, precision of 92.612\%, recall of 92.233\%, F1-score of 92.392\% and AUC of 98.749\%. The model outperforms individual architectures on all performance metrics demonstrating the effectiveness of hybrid feature extraction in Diabetic Retinopathy (DR) classification tasks. To make the proposed model more clinically useful and interpretable, this paper incorporates multiple Explainable Artificial Intelligence (XAI) techniques including Gradient-weighted Class Activation Mapping (Grad-CAM), Grad-CAM++, Score-CAM, Faster Score-CAM and Layer-CAM. These techniques generate visual explanations that clearly indicate the retinal features affecting the model's prediction such as microaneurysms, hemorrhages and exudates so that clinicians can interpret and validate. 
\end{abstract}


\section{Introduction}
The human eye is an amazing sensory organ that is the gateway to visual perception. It detects light and sends signals to the brain through the optic nerve to form images, colors and depth perception \cite{yadav2019classification}. This complex process allows us to navigate, recognize faces and do many activities that require visual input. But many medical conditions can impair the eye and cause significant visual loss. Retinopathy is a condition that affects the retina and is often a result of systemic diseases like hypertension, chronic kidney disease, heart disease and most importantly diabetes. Diabetes is a chronic metabolic disorder characterized by high blood sugar, it is a global health problem affecting millions of people worldwide. It occurs when the body either doesn’t produce insulin (Type 1 diabetes) or can’t use insulin properly (Type 2 diabetes) and results in consistently high blood sugar levels. Over time, uncontrolled diabetes results in complications affecting multiple organ systems including cardiovascular, renal and nervous systems. Among these complications ocular diseases like Diabetic Retinopathy (DR) is a major threat to vision and can be irreversible if left untreated \cite{saeedi2019global, aschner2021international}. The prevalence of diabetes is alarming, 537 million people are affected globally as of 2021, projected to rise to 578 million by 2030 and 700 million by 2045 according to the report of International Diabetic Federation \cite{idf2021}. This burden is disproportionately affecting low and middle income countries where 75\% of diabetes cases are reported. In these regions lack of healthcare access, limited awareness and inadequate screening programs contribute to the high prevalence of diabetes related complications including DR \cite{wong2020strategies, choo2021review}.

Among the many complications of diabetes, Diabetic Retinopathy (DR) is the leading cause of vision loss and blindness in people with diabetes. As of 2020, 103 million adults worldwide had DR, and by 2045 that will increase to 161 million. In 2020 alone, 1.07 million (95\% UI: 0.76–1.51 million) were blind from DR and 3.28 million (95\% UI: 2.41–4.34 million) had moderate to severe visual impairment (MSVI) \cite{curran2024global}. In the United States of America (USA), 9.6 million (26.43\%) people with diabetes had DR in 2021, and 1.84 million (5.06\%) had Vision-Threatening Diabetic Retinopathy (VTDR) \cite{lundeen2023prevalence}. Regionally, the highest burden of DR blindness and MSVI was in Latin America \& the Caribbean (6.95\%, 95\% UI: 5.08–9.51). North America and the Caribbean had high prevalence rates (33.30\%), while South and Central America had the lowest rates (13.37\%). Elsewhere, Africa had the highest DR prevalence at 35.90\%, followed by North Africa \& the Middle East (2.12\%, 95\% UI: 1.55–2.79) \cite{teo2021global}. Ethnic groups also varied in their susceptibility, with Hispanics (OR: 2.92; 95\% CI: 1.22–6.98) and Middle Easterners (OR: 2.44; 95\% CI: 1.51–3.94) more likely to develop DR than Asians \cite{salpea2023call}.

\begin{figure}[htpb]
    \centering
    \begin{subfigure}[b]{0.45\textwidth}
        \centering
        \includegraphics[width=\textwidth]{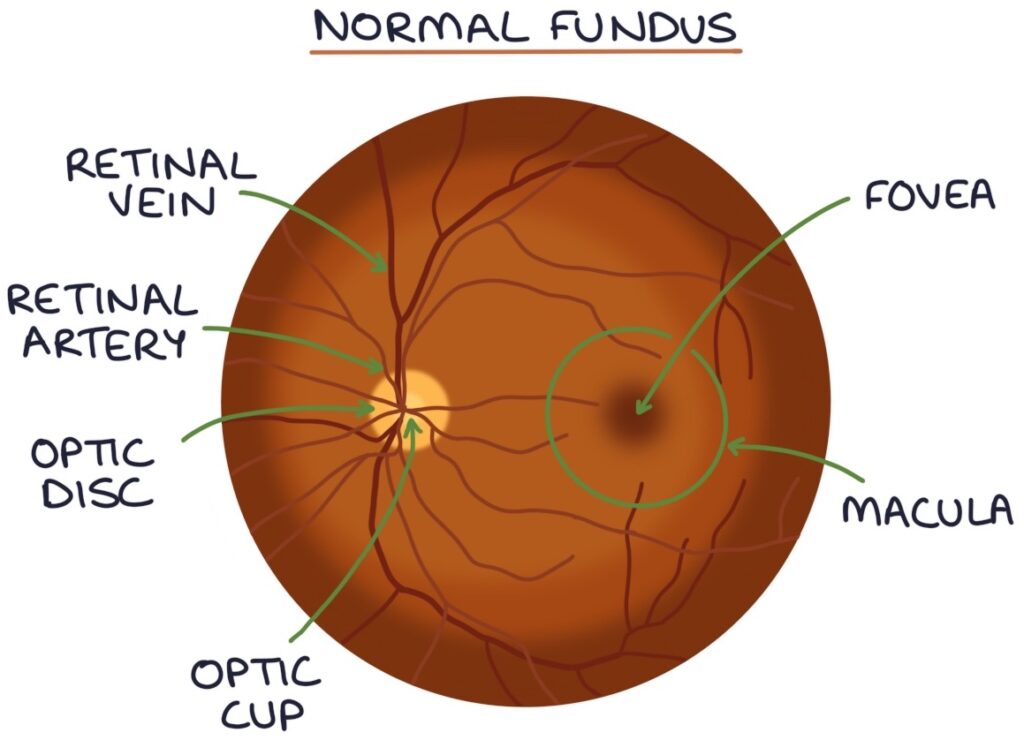}
        \caption{Normal Fundus}
    \end{subfigure}
    \hfill
    \begin{subfigure}[b]{0.52\textwidth}
        \centering
        \includegraphics[width=\textwidth]{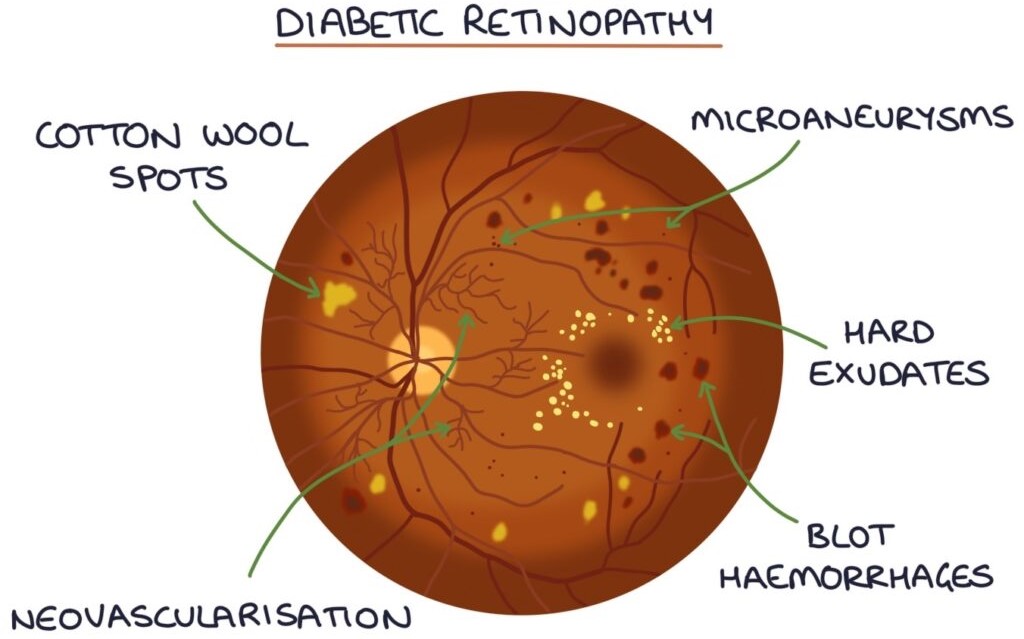}
        \caption{Diabetic Retinopathy}
    \end{subfigure}
    \caption{Differences between Normal Fundus Image and Diabetic Retinopathy Fundus Image. \cite{diabeticretinopathy}}
    \label{Fundus_Comparison}
\end{figure}

Diabetic retinopathy happens when high blood sugar over time damages the small blood vessels in the retina causing microvascular problems like microaneurysms, haemorrhages, hard exudates and neovascularization. Advanced stages are called Proliferative Diabetic Retinopathy (PDR) where new abnormal blood vessels grow and the risk of retinal detachment and total blindness increases. People with diabetes are also prone to other eye problems like glaucoma and cataracts. As shown in \Cref{Fundus_Comparison}, the pathological changes in a diabetic retina compared to a normal one are microaneurysms, cotton wool spots, hard exudates and neovascularization which leads to progressive visual impairment. Symptoms can range from blurry vision and dark spots to total blindness as the disease progresses. Since DR is the leading cause of blindness among working age adults, early detection through regular eye check ups, prompt medical attention and good diabetes management is key to prevent further progression of the disease.

Machine learning (ML) has changed Diabetic Retinopathy (DR) diagnosis by automated and accurate detection, reducing manual assessments and healthcare burden \cite{lakhwani2020impact,bhandari2023literature}. Traditional diagnosis using retinal fundus imaging is time consuming, expert dependent and variable, making early detection challenging; hence researchers use ML and Deep Learning (DL) for stage classification, lesion detection and severity grading \cite{tajudin2022deep}. Hybrid models combining ML, metaheuristic optimization and computational methods have improved performance and generalization \cite{bhandari2023literature,dayana2023comprehensive}. Large public datasets have enabled successful binary and multi-class classifications, but multi-class approaches face challenges like data imbalance, imaging quality issues and suboptimal feature extraction \cite{shakibania2024dual,muthusamy2024deep,akhtar2024diabetic,arora2024ensemble,youldash2024early,olatunji2023comprehensible,yao2024novel,harisha2024deep}. To address these issues, preprocessing methods like contrast enhancement, generative modeling, attention based extraction have improved diagnostic accuracy \cite{alwakid2023deep}. Ensemble methods, transfer learning and hybrid classifications have advanced automated detection of critical DR indicators like microaneurysms, hemorrhages and lesion segmentation. Optimized classifiers and advanced architectures have further improved accuracy and performance in multi-view joint learning using public datasets \cite{xu2024computer}. Despite these improvements, issues like dataset imbalance, noise in fundus images and overfitting still persist, affecting model generalization while computational demands pose challenge for real world clinical integration. But ML driven DR diagnosis has revolutionized ophthalmic diagnostics, making early screening more accessible, efficient disease monitoring and overall patient care outcomes \cite{moustari2024two}.

Despite the progress in Machine learning based Diabetic Retinopathy (DR) detection, we need to address the current limitations. Existing models although work well in many controlled settings, fail to generalise across different datasets and real world clinical environments. This is mainly due to issues like data imbalance, limited diversity in training datasets and inability to capture temporal and contextual features which are important for detecting subtle disease progression. So we focus on improving model generalisation across different populations, incorporating sequential dependencies in retinal features to better represent disease evolution over time and developing consistent, robust and clinically interpretable explainable AI (XAI) methods to support clinical decision making. We also aim to reduce the dependency on large annotated datasets through efficient learning strategies and ensure our models are computationally efficient and scalable for deployment in resource constrained environments. By using advanced computational techniques, state of the art learning frameworks and robust validation processes, we hope to set a new benchmark in DR detection. To achieve this we propose the following new contributions:

\begin{enumerate}[label=\roman*]
    \item \textbf{Hybrid Dataset:} Combines five public datasets to generalize, reduce biases and increase data diversity for robust performance across different imaging conditions.
    \item \textbf{Model Evaluation Across Individual Datasets:} Conducts an extensive evaluation of individual deep learning models, including VGG16, VGG19, ResNet50V2, MobileNetV2, and Xception, on each dataset independently before applying the combined hybrid dataset, to ensure accurate performance assessment and robust generalization.    
    \item \textbf{VR-FuseNet Model:} Proposes and develops VR-FuseNet, a hybrid deep learning model inspired by the VGG19 and ResNet50V2 architectures, effectively leveraging their complementary strengths for improved diabetic retinopathy classification performance.  
    \item \textbf{Explainable AI (XAI) Methods:} Uses Grad-CAM, Grad-CAM++, Layer-CAM, Score-CAM, Faster Score-CAM to evaluate and find the best approach to highlight DR specific features.
    \item \textbf{Clinical Interpretability Heatmaps:} Shows multiple XAI heatmaps comparison, highlighting microaneurysms, exudates and hemorrhages and provides a clinician friendly interface for model validation.
\end{enumerate}   

By addressing these challenges, our approach gives better generalization, better interpretability and better accuracy, hence a more robust and scalable DR detection framework for clinical use. This paper is structured as follows: Section \ref{Review_of_the_Literature} reviews the existing research and methods for DR classification. Section \ref{Approaches_to_Research} outlines the approaches to research, including the techniques and strategies utilized throughout the study. Section \ref{Result_Analysis} presents the evaluation metrics and results of the proposed model. Section \ref{Explainable Artificial Intelligence} shows various Explainable AI (XAI) techniques for DR diagnosis. Section \ref{Limitations and Future Works} summarizes the findings, limitations and future works. Section \ref{Conclusion} concludes the paper.

\section{Review of the Literature}
\label{Review_of_the_Literature}
The automation of DR detection has emerged as a critical area of research due to the rising global prevalence of diabetes and the urgent need for early diagnosis to prevent vision loss. In response, numerous studies have explored a wide range of computational techniques—including traditional machine learning, deep learning, and advanced image analysis—to enhance diagnostic accuracy and clinical applicability. These efforts have involved diverse strategies such as innovative data preprocessing, sophisticated model architectures, and the integration of interpretability frameworks.

Saxena et al. \cite{saxena2020improved} used InceptionV3 and InceptionResNetV2 on EyePACS and Messidor datasets and got AUC values ranging from 0.92 to 0.958. Deep convolutional networks are powerful for DR presence detection. But their model was only for binary classification and not for severity levels of DR—a limitation same as Yi et al. \cite{yi2024midc} who used transfer learning CNNs and improved performance by dataset cleaning and got accuracy from 71.18\% to 85.13\%. But their model was also only for presence/absence classification and not for clinical granularity. Mukherjee et al. \cite{mukherjee2023comparing} targeted this gap by integrating ROI extraction, Z-score normalization and EfficientNet B1 to do severity grading. Their system got impressive results with AUC values of 0.98 for referable DR screening and DR severity grading. But their interpretability framework was limited as they only used Grad-CAM without benchmarking it against other explanation methods. Xu et al. \cite{xu2024computer} proposed RT2Net with robust preprocessing techniques like CLAHE and background removal and got 85.4\%–88.2\% accuracy on EyePACS and APTOS. But their fusion of multi-view features increased computational complexity and made class imbalance more challenging. Tariq et al. \cite{tariq2022transfer} highlighted the importance of choosing the right classifier by comparing SVM and DenseNet121 and got a huge jump in accuracy from 52.57\% to 93\%. But absence of oversampling details limited replication and reliability.

To address class imbalance, Alwakid et al. \cite{alwakid2023deep} and Alwakid et al. \cite{alwakid2023enhancement} conducted experiments on APTOS and DDR datasets using DenseNet121 and advanced image enhancement techniques like ESRGAN, histogram equalization and CLAHE. They reported accuracy of 98.7\% which shows the potential of high resolution image synthesis in improving the classification accuracy. But they didn’t compare with other deep models so the merit of their approach is unclear. Same issue found in Mohanty et al. \cite{mohanty2023using} who developed a hybrid model with VGG16, XGBoost and DenseNet121 and got 97.3\% accuracy but didn’t compare with more recent architectures. Wong et al. \cite{wong2023diabetic} tested how the performance changes across DR grading scales using ShuffleNet and ResNet18 and found significant drop in accuracy with finer class granularity—highlighting the need for more complex multi-class training pipelines. They used only one optimizer and froze the model layers which may limit the generalization. Mutawa et al. \cite{mutawa2023transfer} broadened the architectural comparison by testing VGG16, MobileNetV2 and DenseNet121 but didn’t use image enhancement or denoising which may compromise the robustness in real world clinical settings. Nahiduzzaman et al. \cite{nahiduzzaman2023diabetic} used PCNN-ELM and got decent accuracy on EyePACS and APTOS but didn’t benchmark with CNNs which dominate the domain. Murugappan et al. \cite{murugappan2022novel} introduced DRNet and got 99.7\% in binary and 98.1\% in multi-class classification but their reliance on limited datasets and Grad-CAM alone makes it difficult to evaluate the generalizability and model transparency.

More recent studies have explored interpretability, hybrid modeling and preprocessing customization for DR detection. Abdel Maksoud et al. \cite{abdelmaksoud2022computer} introduced E-DenseNet with HEBPDS and median filtering and got 91.2\% average accuracy across multiple datasets but underperformed for normal DR cases. Bilal et al. \cite{bilal2022ai} used U-Net and CNN-SVD hybrids with green channel transformations and got nearly 98\% on EyePACS but didn’t defined oversampling protocol to handle imbalance. Beham et al. \cite{beham2023optimized} integrated InceptionV3, custom CNN and PBIL and scored high sensitivity and specificity (88.13\% and 96.56\% respectively) but didn’t provide cross comparison with standard models. Butt et al. \cite{butt2022diabetic} combined GoogleNet and ResNet18 with SVM and got 97.8\% binary classification but reduced DR classes from 5 to 3 which may undermine nuanced diagnosis. Abbood et al. \cite{abbood2022hybrid} used Gaussian blur, foreground identification and DRRNet and scored up to 93.6\% but struggled to detect mild DR due to weak supervision and illumination inconsistencies. Similarly Ali et al. \cite{ali2023hybrid} applied IR-CNN with histogram equalization and normalization and got 96.85\% but evaluated only 2 deep models (ResNet50 and InceptionV3). Raiaan et al. \cite{raiaan2023lightweight} used RetNet-10 with comprehensive set of preprocessing methods (CLAHE, morphological opening, ROI extraction) and got 98.65\% but limited availability of annotated DR grade samples reduced reliability. Menauer et al. \cite{menaouer2022diabetic} implemented hybrid CNN+VGG16+VGG19 architecture and got 90.6\% but didn’t provide baseline comparison and mentioned need for longer training epochs. Mercaldo et al. \cite{mercaldo2023diabetic} compared CNNs like VGG16, MobileNet and AlexNet for both binary and severity classification and got up to 97\% but struggled with class imbalance and didn’t specify oversampling details. Vasireddi et al.\cite{vasireddi2024dr} used DFNN with LOA and got 98.04\% but oversimplified DR progression and used only LIME for explainability. Alghamdi et al. \cite{alghamdi2022towards} tested VGG16, DenseNet121 and ResNet18 and got limited performance in multiclass (48.43\%) and used only Grad-CAM which restricted interpretability depth. Many works adopted only a single explainable AI technique. In this context, \Cref{Comparison_of_Related_Works} provides the findings and methodologies from the 24 research papers that were systematically reviewed as part of this study.

\begin{longtable}[c]{|p{0.08\textwidth}|p{0.08\textwidth}|p{0.14\textwidth}|p{0.13\textwidth}|p{0.1\textwidth}|p{0.12\textwidth}|p{0.16\textwidth}|}
\caption{Comparative Analysis of Previous Studies} 
\label{Comparison_of_Related_Works} \\ 
\hline
\makecell{\multicolumn{1}{c}{\textbf{Author}}} & \makecell{\multicolumn{1}{c}{\textbf{Dataset}}} & \makecell{\multicolumn{1}{c}{\textbf{Preprocessing}} \\ \multicolumn{1}{c}{\textbf{Techniques}}} & \makecell{\multicolumn{1}{c}{\textbf{Methodology}}} & \makecell{\multicolumn{1}{c}{\textbf{Result}} \\ \multicolumn{1}{c}{\textbf{Analysis}}} & \makecell{\multicolumn{1}{c}{\textbf{Explainable}} \\ \multicolumn{1}{c}{\textbf{AI}}} & \makecell{\\\multicolumn{1}{c}{\textbf{Limitation}}} \\
\hline
\endfirsthead
\hline
\makecell{\multicolumn{1}{c}{\textbf{Author}}} & \makecell{\multicolumn{1}{c}{\textbf{Dataset}}} & \makecell{\multicolumn{1}{c}{\textbf{Preprocessing}} \\ \multicolumn{1}{c}{\textbf{Techniques}}} & \makecell{\multicolumn{1}{c}{\textbf{Methodology}}} & \makecell{\multicolumn{1}{c}{\textbf{Result}} \\ \multicolumn{1}{c}{\textbf{Analysis}}} & \makecell{\multicolumn{1}{c}{\textbf{Explainable}} \\ \multicolumn{1}{c}{\textbf{AI}}} & \makecell{\\\multicolumn{1}{c}{\textbf{Limitation}}} \\
\endhead
\hline
\multicolumn{7}{|r|}{Table continues to next page} \\
\hline
\endfoot
\hline
\endlastfoot

Saxena et al. \cite{saxena2020improved} & EyePACS, Messidor 1, Messidor 2 & Cropping and Centering Data Augmentation & InceptionV3, InceptionResNetV2 & AUC of 0.927 on EyePACS, AUC of 0.958 on Messidor 1, AUC of 0.92 on Messidor 2 & Model name wasn't mentioned & Only checked if DR were present or not, didn't separate different levels of DR severity \\ \hline

Tariq et al. \cite{tariq2022transfer} & EyePACS, APTOS 2019 & Image Resizing & ResNet50, DenseNet121, SVM & EyePACS: SVM 52.57\% APTOS 2019: DenseNet121 93\% & \multicolumn{1}{c|}{\textbf{X}} & Class imbalance, exact amount of data oversampling details were missing \\ \hline

Mukherjee et al. \cite{mukherjee2023comparing} & EyePACS, Hybrid (APTOS 2019 + Messidor 1+ Messidor 2) & ROI Extraction, Ben Graham Approach, Z-score Normalization, Data Augmentation & Transfer Learning CNNs & EfficientNet B1 recorded an accuracy of 0.9511 in DR severity grading, 0.9934 in DR-Screening and 0.9798 in referable DR & Grad-CAM & Exact amount of data oversampling details were missing, only Grad-CAM was used for explanation, without comparing other methods \\ \hline

Yi et al. \cite{yi2024midc} & APTOS 2019, Messidor 2 & Didn't mention any preprocessing technique & Transfer Learning CNNs & Binary Classification for the APTOS 2019 Dataset: average accuracy after cleaning increased from 71.18\% to 85.13\% & \multicolumn{1}{c|}{\textbf{X}} & Classified only DR as only present or absent, didn't distinguish between different severity levels \\ \hline

Xu et al. \cite{xu2024computer} & APTOS 2019, EyePACS & Image Resizing, Removing Background, Image Gray-scaling, CLAHE & RT2Net & APTOS 2019: 85.4\%, EyePACS: 88.2\% & \multicolumn{1}{c|}{\textbf{X}} & Class imbalance, computational complexity due to multi-view feature fusion \\ \hline

Alwakid et al. \cite{alwakid2023deep} & APTOS 2019, DDR & CLAHE, ESRGAN, Data Augmentation & DenseNet121 & APTOS 2019: 98.7\% , DDR: 79.67\% & \multicolumn{1}{c|}{\textbf{X}} & Small sample size and dataset biases, only DenseNet121 was used, without testing other models \\ \hline

Mohanty et al. \cite{mohanty2023using} & APTOS 2019 & Image Resizing, Gaussian Blur, Ben Graham Approach & Hybrid Model (VGG16 + XGBoost Classifier), DenseNet121 & Hybrid Model: 79.50\%, DenseNet 121: 97.30\% & \multicolumn{1}{c|}{\textbf{X}} & Only VGG16 and DenseNet 121 were tested, no comparison with other deep learning models \\ \hline

Moustari et al. \cite{moustari2024two} & APTOS 2019 & Laplacian of Gaussian (LoG), CLAHE & DenseNet121 & 98.42\% & Grad-CAM & Imbalance Dataset, only DenseNet121 was used, limiting model comparison, only Grad-CAM was used without comparing other interpretation methods \\ \hline

Alwakid et al. \cite{alwakid2023enhancement} & APTOS 2019 & Histogram, CLAHE, ESRGAN, Data Augmentation & DenseNet121 & 98.36\% & \multicolumn{1}{c|}{\textbf{X}} & Only DenseNet121 was used, without testing other models, ESRGAN-based image enhancement requires high computation, extensive hyperparameter tuning was needed for optimization \\ \hline

Wong et al. \cite{wong2023diabetic} & APTOS 2019, Hybrid (EyePACS+ Messidor 2) & CLAHE & ShuffleNet, ResNet18 & APTOS 2019 2 Class Grading: 96\%, APTOS 2019 5 Class Grading: 82\%, Hybrid Dataset 3 Class Grading: 75\% & \multicolumn{1}{c|}{\textbf{X}} & ADE was the only optimizer tested, without comparing others, frozen layers were used from only two transfer learning models \\ \hline

Nahiduz-zaman et al. \cite{nahiduzzaman2023diabetic} & EyePACS, APTOS 2019 & CLAHE & PCNN-ELM & EyePACS: 91.78\%, APTOS 2019: 97.27\% & \multicolumn{1}{c|}{\textbf{X}} & Imbalanced dataset, used ML techniques for evaluation metrics, no comparison with deep learning models \\ \hline

Mutawa et al. \cite{mutawa2023transfer} & APTOS 2019, EyePACS, ODIR & Data Augmentation & VGG16, InceptionV3, MobileNetV2, DenseNet 121 & By using hybrid dataset, DenseNet 121: 98.97\%, VGG16: 98.79\%, MobileNetV2: 98.51\%, InceptionV3: 97.21\% & \multicolumn{1}{c|}{\textbf{X}} & Didn't used any image enhancement, noise removal preprocessing technique \\ \hline

Murug-appan et al. \cite{murugappan2022novel} & APTOS 2019 & Image Resizing & DRNet & Binary Classification: 99.73\%, Multiclass Classification: 98.18\% & \textbf{Grad-CAM} & No benchmarked datasets for proper few-shot classification evaluation, 100-episode limit may exclude some training samples, only Grad-CAM was used, without comparing other methods \\ \hline

Abdel Maksoud et al. \cite{abdelmaksoud2022computer} & APTOS 2019, IDRiD, EyePACS, Messidor & Histogram Equalization for Brightness Preservation based on a Dynamic Stretching (HEBPDS) and Median Filter, Data Augmentation & Hybrid E-DenseNet & Average 91.2\% & \multicolumn{1}{c|}{\textbf{X}} & Low AUC for the normal class (35\% in most cases) except higher AUC (67\%) only in the MESSIDOR dataset \\ \hline

Bilal et al. \cite{bilal2022ai} & Messidor 2, EyePACS 1, DIARETDB0 & Green Channel Image, Top-Bottom Hat Transformation, Data Augmentation & U-Net models, Hybrid CNN-SVD & In InceptionV3, EyePACS 1: 97.92\%, Messidor 2: 94.59\%, DIARETDB0: 93.52\% & \multicolumn{1}{c|}{\textbf{X}} & Class imbalance was present, lack of defined oversampling counts \\ \hline

Beham et al. \cite{beham2023optimized} & EyePACS & Didn't mention any preprocessing technique & InceptionV3, Custom CNN, PBIL & PBIL Sensitivity: 88.13\%, PBIL Specifivity: 96.56\% & \multicolumn{1}{c|}{\textbf{X}} & Only specific CNN architectures were used without comparing other models \\ \hline

Butt et al. \cite{butt2022diabetic} & APTOS 2019 & Normalization, Image Resizing & (GoogleNet and ResNet18) +SVM & Binary Classification: 97.80\%, Multiclass Classification: 89.29\% & \multicolumn{1}{c|}{\textbf{X}} & Classes were reduced from five to three to manage data imbalance \\ \hline

Abbood et al. \cite{abbood2022hybrid} & EyePACS, Messidor & Foreground Identification and Circle Crop, Gaussian Blur & DRRNet & EyePACS: 92\% , Messidor: 93.6\% & \multicolumn{1}{c|}{\textbf{X}} & Limited image level supervision affects DR feature detection, difficulty in distinguishing Mild DR due to the similarity with normal grades, illumination issues remain unresolved \\ \hline

Ali et al. \cite{ali2023hybrid} & EyePACS & Histogram Equalization, Intensity Normalization & IR-CNN & 96.85\% & \multicolumn{1}{c|}{\textbf{X}} & Class imbalance, Only ResNet50 and InceptionV3 were used, without testing other architectures \\ \hline

Raiaan et al. \cite{raiaan2023lightweight} & APTOS 2019, Messidor 2, IDRiD & OTSU Thresholding, Contour Detection and Sorting, ROI Extraction, Morphological Opening, NLMD, CLAHE, Data Augmentation & RetNet-10 & 98.65\% & \multicolumn{1}{c|}{\textbf{X}} & Few images available for DR grades three and four after combining datasets, needed better preprocessing to handle noisy images \\ \hline

Menaouer et al. \cite{menaouer2022diabetic} & APTOS 2019 & Data Augmentation & Hybrid Model (CNN+ VGG16+ VGG19) & 90.6\% & \multicolumn{1}{c|}{\textbf{X}} & Only hybrid deep learning features were used, without comparing other techniques limited dataset size, more training epochs are needed to improve accuracy \\ \hline

Alghamdi et al. \cite{alghamdi2022towards} & APTOS 2019 & Image Resizing & VGG16, DenseNet121, ResNet18 & VGG16 for, Binary Classification: 73.04\%, Multiclass Classification: 48.43\% & Grad-CAM & Only Grad-CAM was used, without comparing other methods \\ \hline

Mercaldo et al. \cite{mercaldo2023diabetic} & Hybrid (EyePACS+ APTOS 2019) & Gaussian Filter, Cropping, Data Augmentation & StandardCNN, VGG16, AlexNet, MobileNet & Binary Classification for, Presence of DR: VGG16 97\%, Severity of DR: VGG16 and MobileNet 91\% & Grad-CAM, Score-CAM & Class imbalance was present, lack of defined oversampling counts \\ \hline

Vasireddi et al. \cite{vasireddi2024dr} & Messidor & AHE, Gamma Correction, CLAHE & DFNN with LOA & 98.04\% & LIME & Assumption of constant learned features ignoring DR progression, low amount of data, only LIME was used, without comparing other interpretability methods \\ \hline

\end{longtable}

\section{Approaches to Research}
\label{Approaches_to_Research}

\begin{figure}[htpb]
    \centering
    \includegraphics[width=\linewidth]{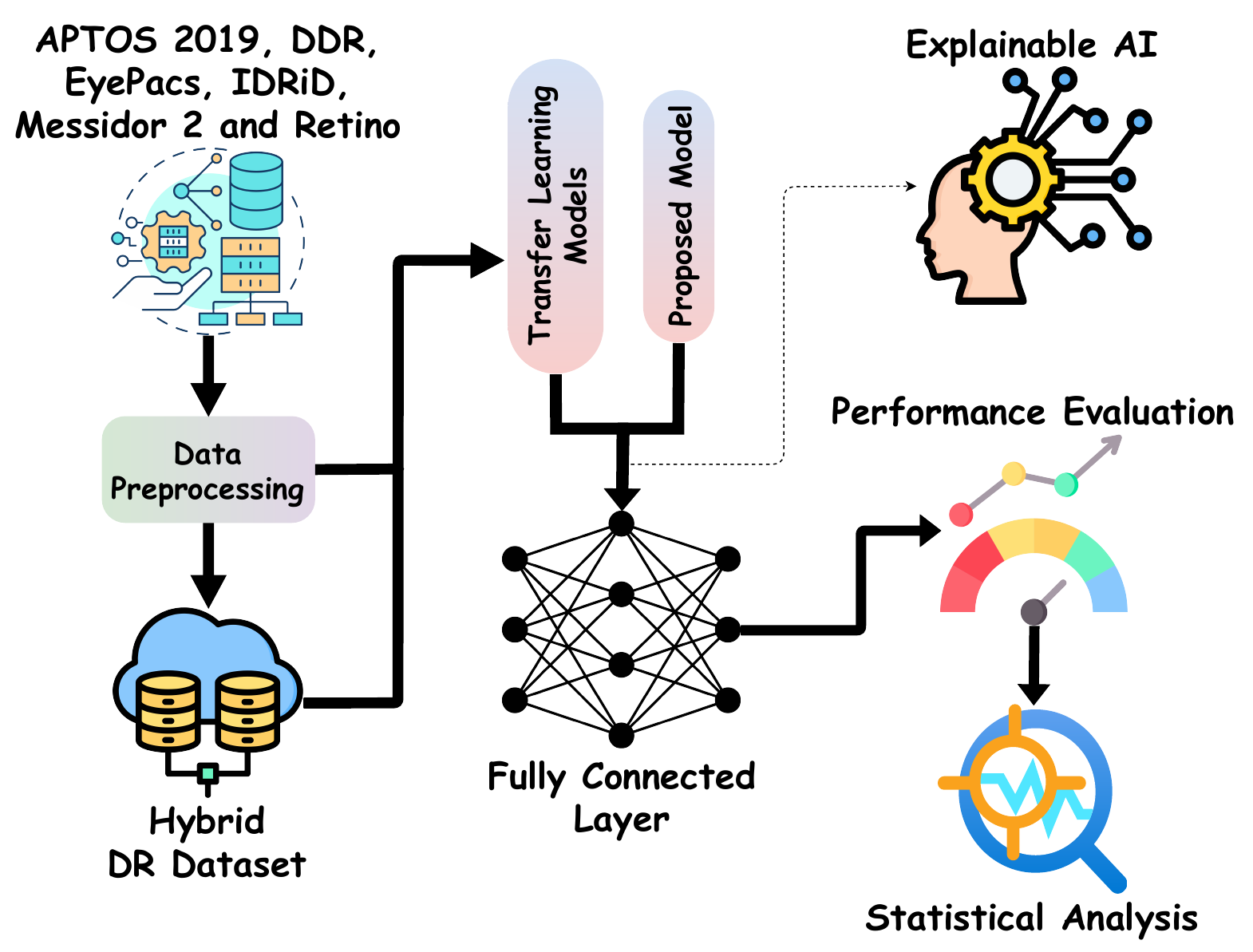}
    \caption{Proposed Methodology}
    \label{Proposed_Methodology}
\end{figure}

Our proposed methodology uses a systematic approach to detect diabetic retinopathy using advanced deep learning techniques. \Cref{Proposed_Methodology} shows the workflow of our method. We selected five public datasets to get a big and representative dataset. After that we did comprehensive preprocessing including Random Undersampling and SMOTE for class imbalance, CLAHE for contrast enhancement, normalization for same data scale and resizing images to a standard size for model input. Each preprocessed dataset was evaluated individually using popular transfer learning models like VGG16, VGG19, ResNet50V2, MobileNetV2, and Xception to get baseline metrics. Then we combined all the datasets into one hybrid dataset to test the transfer learning and our proposed models robustness and generalizability across different retinal images. To make it more transparent and interpretable we applied five gradient based Explainable AI (XAI) methods using the final convolutional layers of our trained models. We did comprehensive performance evaluation and statistical analysis to quantify the improvements and validate our approach.

\subsection{Dataset}
\begin{figure}[htpb]
    \centering
    \includegraphics[width=\linewidth]{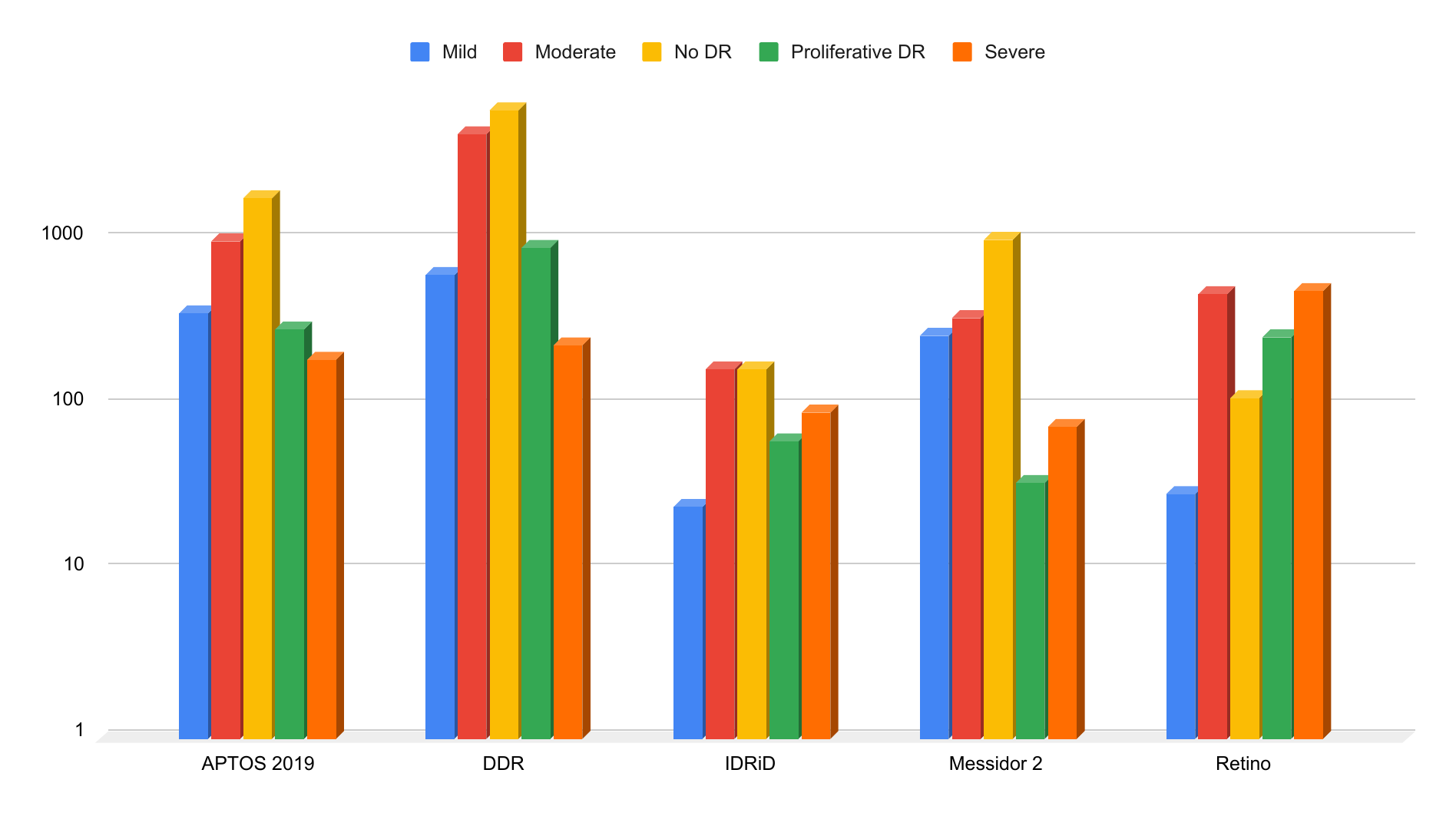}
    \caption{Class Distribution Across Different Datasets Visualized on a Logarithmic Scale}
    \label{Imbalance_Datasets_Graph}
\end{figure}

\begin{figure}[htpb]
    \centering
    \includegraphics[width=\linewidth]{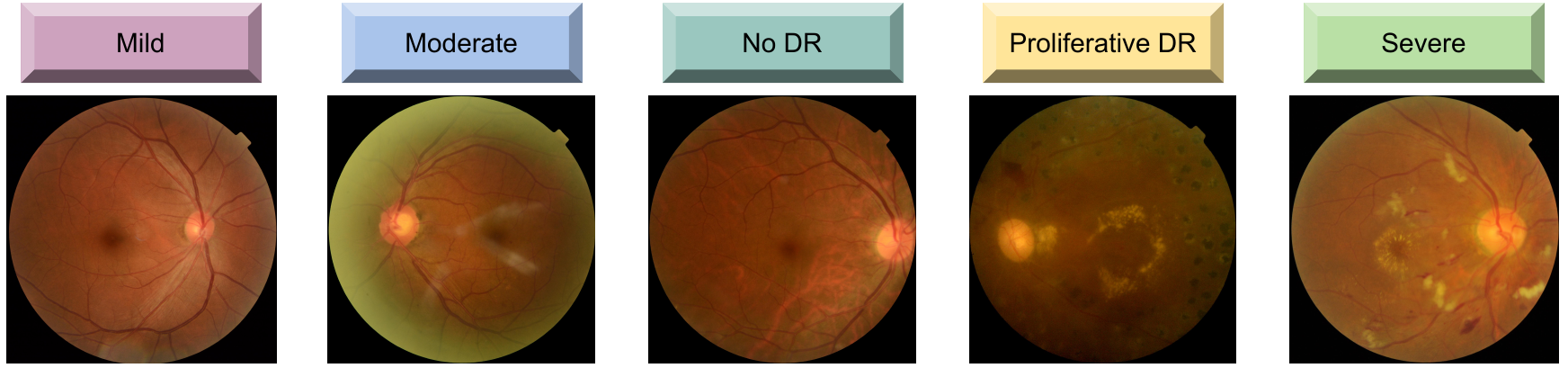}
    \caption{Sample Images of the Dataset}
    \label{Sample_Images}
\end{figure}

To ensure a robust and generalized model for diabetic retinopathy (DR) detection, we utilize five publicly available datasets: APTOS 2019, DDR, IDRiD, Messidor 2 and Retino. The integration of multiple datasets enhances diversity in imaging conditions, camera settings, and population demographics, reducing biases and improving model performance across different clinical scenarios. \Cref{Distribution_of_Datasets} and \Cref{Imbalance_Datasets_Graph} presents the distribution of various Diabetic Retinopathy stages across different datasets, with \Cref{Imbalance_Datasets_Graph}  employing a logarithmic scale to enhance the visibility of class imbalances .\Cref{Sample_Images} dispalys the sample images of various datasets.

\begin{table}[htpb]
    \centering
    \caption{Distribution of Diabetic Retinopathy Stages Across Different Datasets}
    \label{Distribution_of_Datasets}
    \renewcommand{\arraystretch}{1.4} 
    \begin{tabular}{|c|c|c|c|c|c|c|}
        \hline
        \textbf{Dataset Name} & \textbf{Mild} & \textbf{Moderate} & \textbf{No DR} & \textbf{Proliferative DR} & \textbf{Severe} \\ 
        \hline
        APTOS 2019  & 370  & 999  & 1805  & 295  & 193  \\ 
        \hline
        DDR  & 630  & 4477  & 6266  & 913  & 236  \\ 
        \hline
        IDRiD  & 25  & 168  & 168  & 62  & 93  \\ 
        \hline
        Messidor 2  & 270  & 347  & 1017  & 35  & 75  \\ 
        \hline
        Retino  & 30  & 480  & 112  & 265  & 505  \\ 
        \hline
    \end{tabular}
\end{table}

\subsubsection{Description of used dataset}
\textbf{APTOS 2019:} The APTOS 2019 Blindness Detection Dataset, a publicly available Kaggle dataset \cite{karthik2019}, consists of 3,662 high-resolution retinal fundus images, categorized into five severity levels of diabetic retinopathy. The dataset includes images labeled as no DR, mild DR, moderate DR, severe DR, and proliferative DR, with 1805, 370, 999, 193, and 295 images, respectively. The images, with a resolution of 3216 × 2136 pixels, were collected from various clinics using different imaging devices, introducing natural variations in image quality and lighting conditions, which help create a more diverse and challenging dataset.

\textbf{DDR:} The DDR dataset \cite{li2019diagnostic} has six stages; however, for uniformity with other datasets, we employ just five stages of diabetic retinopathy (DR), from stage 0 (no DR) to stage 4 (proliferative DR), omitting the sixth stage. This dataset, accessible on Github \cite{nkicsl2019}, comprises 12,522 high-resolution retinal images gathered from 147 universities across 23 provinces in China. Furthermore, all photos have been subjected to preprocessing to eliminate black backgrounds, thereby providing cleaner and more dependable inputs for model training.

\textbf{IDRiD:} The Indian Diabetic Retinopathy Image Dataset (IDRiD) \cite{porwal2018indian}, available on IEEE DataPort \cite{porwal2018}, consists of 516 fundus images, including both disease grading and lesion segmentation annotations. The images were collected from Shiva Netralaya, Uttar Pradesh, India. IDRiD mainly focuses on moderate to severe DR cases, making it particularly valuable for studying disease progression and lesion localization. This dataset provides detailed ground truth information for developing deep learning models that not only classify DR severity but also detect important retinal abnormalities.

\textbf{Messidor 2:} The Messidor 2 dataset is an extension of the original Messidor dataset \cite{decenciere2014feedback}, consisting of 1,748 high-resolution macula-centered retinal fundus images from 874 patient examinations, sourced from a website \cite{adcis}. The dataset is divided into two parts: Messidor-Original, which contains 1,058 images, and Messidor-Extension, which contains 690 images. The images were captured using a Topcon TRC NW6 non-mydriatic fundus camera at Brest University Hospital in France between 2009 and 2010. While the original dataset lacks official DR severity labels, we utilize an annotated version obtained from a third-party Kaggle dataset \cite{googlebrain2018}, proposed by Krause et al. \cite{krause2018grader}, providing detailed DR grading for classification tasks. Out of the total 1,748 images available for classification, 4 images lacked labels and were therefore excluded from analysis. Consequently, the classification experiments were conducted utilizing the remaining 1,744 labeled images. Messidor-2 is widely used in medical imaging research due to its high-quality fundus images, making it suitable for developing and testing deep learning models for DR detection.

\textbf{RETINO:} The RETINO dataset \cite{wang2023real}, available on figshare \cite{wang2023}, is for automated DR grading using retinal fundus images. It has 1,392 images, each for one patient, from Shanghai Tenth People’s Hospital using Canon non-mydriatic fundus cameras. Images are labeled with the 5 DR levels, from no DR to proliferative DR. The average image size is 2736 × 1824, and all images were re-checked by an expert ophthalmologist with over 10 years of experience to ensure the labels are correct. This dataset is very useful for evaluating automated screening methods for DR, reducing the dependence on manual diagnosis and early detection.

\subsection{Hybrid Dataset}

\begin{figure}[htpb]
    \centering
    \includegraphics[width=0.6\linewidth]{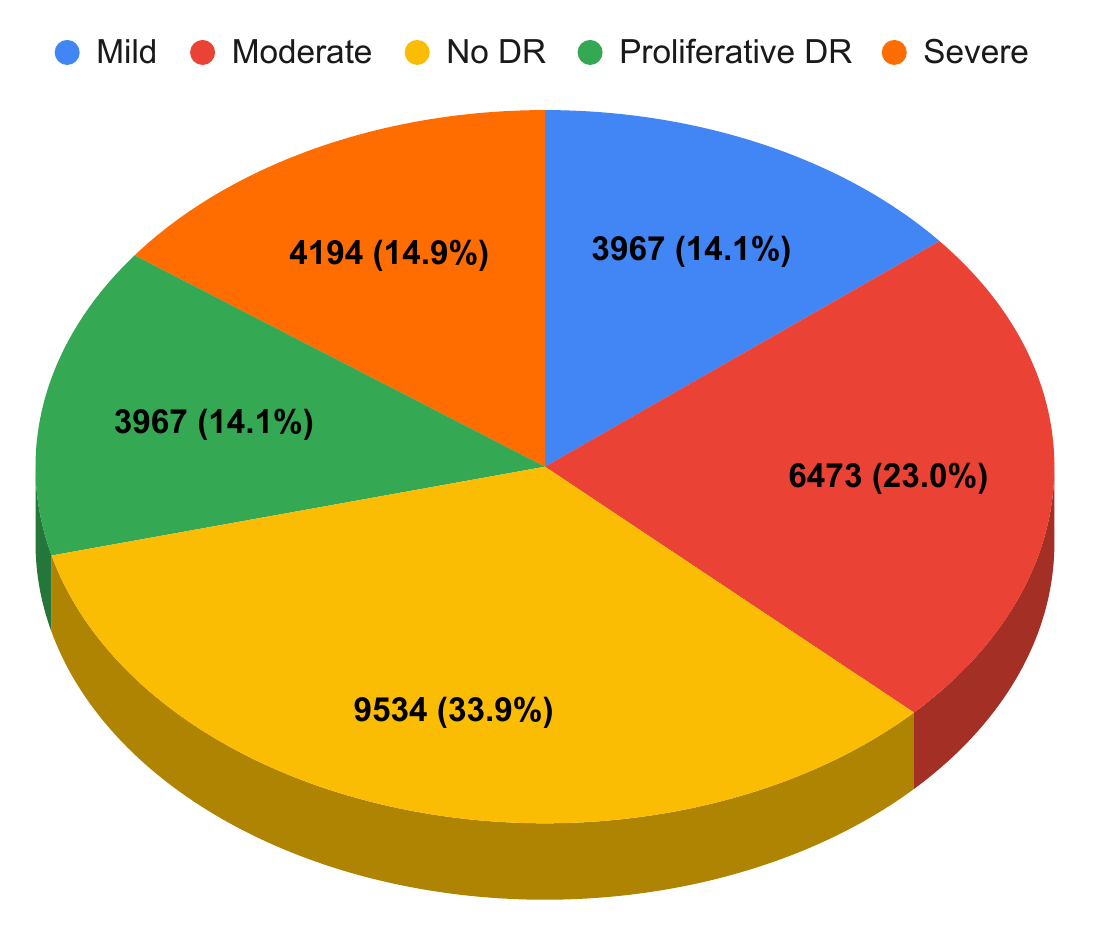}
    \caption{Distribution of Sample Images in the Hybrid Dataset}
    \label{Hybrid_Dataset}
\end{figure}

The hybrid dataset is a combination of 5 publicly available diabetic retinopathy datasets: APTOS 2019, DDR, IDRiD, Messidor 2 and Retino. Each of these datasets is publicly available and curated independently by different sources, clinical settings and research institutions. They together provide a wide range of diabetic retinopathy conditions and severities. The integration process was to merge these datasets into one. This resulted in a very diverse dataset with huge variability in terms of image quality, resolution, clinical severity and diagnostic characteristics. \Cref{Hybrid_Dataset} shows the distribution of Sample Images in the Hybrid Dataset.

The hybrid dataset used for training deep learning models in Diabetic Retinopathy (DR) classification has images in multiple severity levels. It has 3,967 images as Mild DR, 4,194 images as Moderate DR, 9,534 images as No DR, 3,967 images as Proliferative DR, and 6,473 images as Severe DR. By combining multiple publicly available datasets, this hybrid dataset covers the clinical and imaging variability that is seen in real world. The images are from different sources so the dataset takes care of the image quality, lighting conditions and retinal structures from different patient population. This diversity is very important for generalizability of deep learning models so they can adapt to different clinical settings and imaging conditions. So this hybrid dataset is a good and reliable base to train, validate and test deep learning models for accurate and robust DR diagnosis across all severity levels.

\subsection{Data Preprocessing}
\begin{figure}[htpb]
    \centering
    \includegraphics[width=\linewidth]{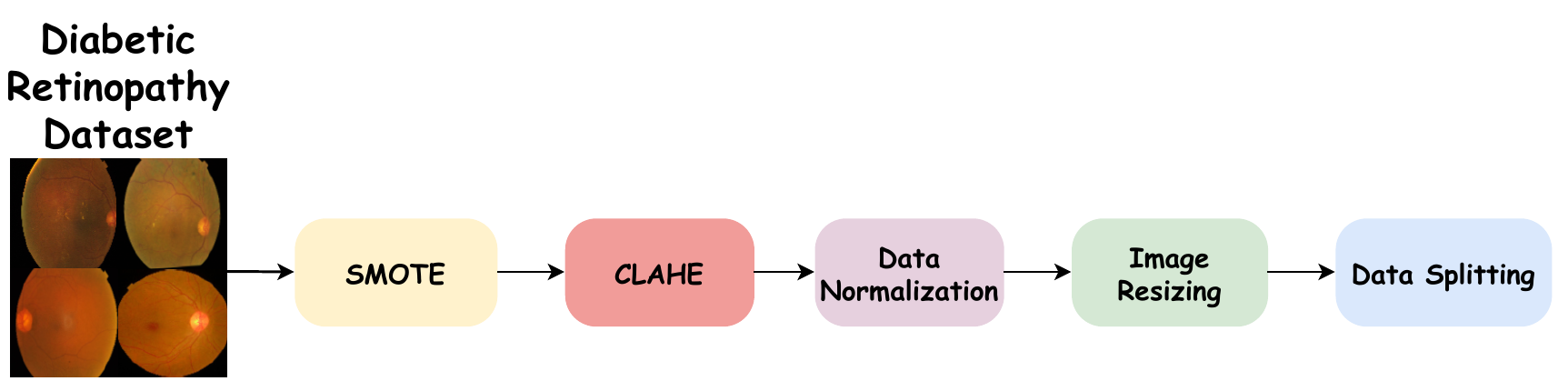}
    \caption{Workflow of the pre-processing techniques}
    \label{Figure pre-processing}
\end{figure}

Image preprocessing plays a vital role in medical image analysis, as the effectiveness of classification models heavily depend on the quality of preprocessing applied to the images. \Cref{Figure pre-processing} shows the workflow of the pre-processing techniques. In this case, initially, class imbalance is handled through SMOTE oversampling techniques.
CLAHE is applied to enhance the contrast of the retinal images. Following contrast enhancement, this preprocessing steps involve implementing image normalization and resizing techniques. Normalization initially adjusts pixel intensity values, generally within the range of 0 to 1, to enhance consistency in visual representation. Images are subsequently downsized to a specified dimension of 128×128 pixels, maintaining uniformity and keeping essential visual qualities. These preprocessing approaches are crucial for diminishing computational complexity and improving performance in subsequent picture analysis or model training activities. Finally, the dataset is partitioned into training, validation and test sets in an 80:10:10 ratio, designating 80\% for training, 10\% for validation and 10\% for testing \cite{shuvo2024advanced}.

\subsubsection{Synthetic Minority Over-Sampling Technique (SMOTE)}
Synthetic Minority Over-sampling Technique also known as SMOTE is a resampling method that addresses class imbalance by generating synthetic data points rather than duplicating existing samples. This was initially inspired by handwritten character recognition where rotation and skewing was applied to create more training data \cite{chawla2002smote}. Unlike those domain specific augmentations, SMOTE generates new synthetic samples in the feature space by interpolating between real data points to create a balanced dataset and expand the decision boundary of the minority class. This technique helps expand the feature space representation of the minority class, preventing overfitting and improving the generalization capability of the classifier. The synthetic instance $x_{\text{new}}$ is generated using the following interpolation formula:
\begin{equation}
x_{\text{new}} = x_i + \delta \cdot (x_{zi} - x_i) 
\label{eq:smote-basic}
\quad \text{\cite{chawla2002smote}}
\end{equation}
where $x_i$ is a minority class sample, $x_{zi}$ is one of its $k$-nearest neighbors, and $\delta \in [0, 1]$ is a random number.
To generalize this process when generating $N$ synthetic samples per minority class instance, the interpolation can be repeated across different neighbors and values of $\delta$:

\begin{equation}
x_{\text{new}}^{(n)} = x_i + \delta_n \cdot (x_{zi}^{(n)} - x_i), \quad n = 1, 2, ..., N 
\label{eq:smote-multiple}
\quad \text{\cite{chawla2002smote}}
\end{equation}

SMOTE can also be applied to image datasets particularly in medical imaging tasks. Since images are high dimensional data, applying SMOTE directly on raw pixel values can distort meaningful medical patterns. Instead, the process involves first converting images to feature vectors using a deep learning based embedding model. These feature vectors represent the underlying structure and characteristics of the images so we can apply SMOTE in a more meaningful way. After extracting these features, SMOTE identifies k-nearest neighbors of each minority class image and generates synthetic feature vectors through interpolation. Then reshape the synthetic feature vectors back to image to use for training. By ensuring the synthetic samples have clinically relevant features, this technique helps improve classification accuracy while minimizing the effect of class imbalance in the dataset \cite{joloudari2023effective, dablain2022deepsmote}.

\begin{algorithm}
\caption{SMOTE-Based Image Oversampling}
\begin{algorithmic}[1]
\State \textbf{Input:} Minority class samples $T$, oversampling percentage $N$, nearest neighbors $k$
    \If{$N < 100\%$}  
        \State Randomly select a subset of $T$ samples  
    \EndIf  
    \State Compute number of synthetic samples:  
    \State $num\_synthetic = (N / 100) * len(T)$
    \For{each sample $x_i$}  
        \State Compute Euclidean distance to k-nearest neighbors:
        \State $d(x_i, x_{NN}) = \sqrt{\sum (x_i - x_{NN})^2}$
    \EndFor
    \For{each sample $x_i$}  
        \State Select a random neighbor $x_{NN}$  
        \State Compute difference:  
        \State $diff = x_{NN} - x_i$
        \State Generate a random gap $\lambda$ between 0 and 1:  
        \State $lambda = random(0,1)$
        \State Compute synthetic sample:  
        \State $x\_syn = x_i + lambda * diff$
        \State Store $x\_syn$ in dataset  
    \EndFor
\State \textbf{Repeat Until Balance is Achieved:} Continue generating synthetic samples until $N\%$ oversampling is met.
\State \textbf{Return New Balanced Dataset:} Merge synthetic samples with the original dataset.
\label{SMOTE}
\end{algorithmic}
\end{algorithm}

\begin{figure}[htpb]
    \centering
    \includegraphics[width=\linewidth]{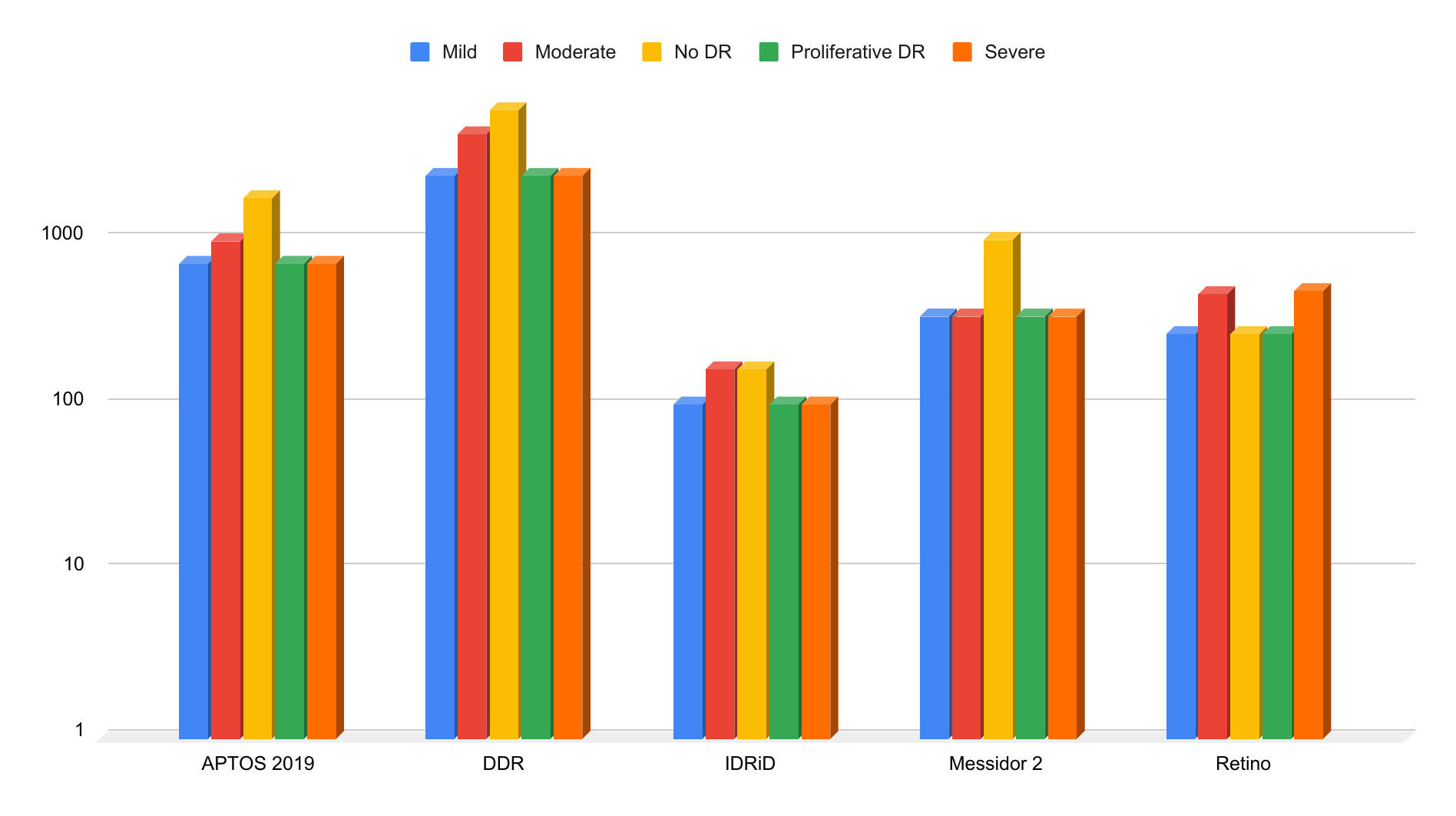}
    \caption{Comparison of Different Datasets by Categories after SMOTE using Log Scale Plot}
    \label{Figure SMOTE}
\end{figure}

This \Cref{SMOTE} is applied to image datasets by first converting images into feature vectors that represent the patterns and feature distributions of the images. Since direct pixel level interpolation can distort images, SMOTE operates in feature space where each image is represented as a high dimensional numerical vector. After finding k-nearest neighbors for each minority class image, a random neighbor is selected and a new synthetic feature vector is generated by interpolating between the original vector and its neighbor with a random gap value $\lambda$. The synthetic feature vector is then reshaped back into image to make sure the newly generated images preserve the meaningful medical features like microaneurysms, hemorrhages and exudates. Finally, the synthetic images are added to the original dataset to improve class balance, reduce bias and enhance the model’s ability to classify diabetic retinopathy cases \cite{belharar2024}. Following this process, SMOTE has been applied across the entire five datasets. In every dataset, the oversampling process is conducted by computing the mean value among the five stages, ensuring representative distribution of DR stages while mitigating the risks of class bias. \Cref{SMOTE_distribution} and \Cref{Figure SMOTE}  illustrate the distribution of different stages across various datasets following the application of SMOTE, with a logarithmic scale applied to improve the clarity of class imbalance visualization.

\begin{table}[htpb]
\centering
\caption{Distribution of Diabetic Retinopathy Stages Across Different Datasets after SMOTE}
\label{SMOTE_distribution}
\renewcommand{\arraystretch}{1.4} 
\begin{tabular}{|c|c|c|c|c|c|c|}
\hline
\textbf{Dataset Name} & \textbf{Mild} & \textbf{Moderate} & \textbf{No DR} & \textbf{Proliferative DR} & \textbf{Severe} \\ \hline
APTOS 2019                                                      & 733           & 999               & 1805           & 733                                                                  & 733           \\ \hline
DDR                                                             & 2504          & 4477              & 6266           & 2504                                                                 & 2504          \\ \hline
EyePACS                                                         & 3025          & 5292              & 5810           & 3025                                                                 & 3025          \\ \hline
IDRiD                                                           & 103           & 168               & 168            & 103                                                                  & 103            \\ \hline
Messidor 2                                                      & 349           & 347               & 1017           & 349                                                                  & 349           \\ \hline
Retino                                                          & 278           & 480               & 278            & 278                                                                  & 505           \\ \hline
\end{tabular}
\end{table}

\subsubsection{Contrast Limited Adaptive Histogram Equalization (CLAHE)}
Contrast Limited Adaptive Histogram Equalization (CLAHE) \cite{zuiderveld1994contrast} is an improved version of Adaptive Histogram Equalization (AHE) \cite{pizer1987adaptive} which enhances contrast without amplifying noise in uniform areas. Unlike global histogram equalization which applies uniform contrast to the whole image, AHE works on local regions, so it’s more suitable for images with non uniform lighting. But AHE tends to over enhance noise in uniform areas and produces artifacts. CLAHE overcomes this by introducing a clip limit to prevent excessive contrast changes. The CLAHE process starts by dividing the image into non overlapping tiles (sub-regions), usually 8×8 for a 512×512 image, to ensure local contrast enhancement. Each tile does histogram computation, where pixel intensity distribution is computed independently. The contrast clip limit is then applied, which is a multiple of the average histogram contents, so no histogram bin exceeds a specified threshold. If there are excess pixels beyond the clip limit, they are redistributed across the whole histogram, keeping the total pixel count balanced. Then a Cumulative Distribution Function (CDF) is computed to get intensity mapping. Since equalizing each tile separately produces artificial boundaries, CLAHE applies bilinear interpolation to blend adjacent tiles smoothly, so there’s no gap between regions.

To formalize the mapping function, for each region, the grayscale mapping is obtained by estimating the histogram density function. The mapping is calculated as:
\begin{equation}
f_{i,j}(n) = \frac{(N-1)}{M} \sum_{k=0}^{n} h_{i,j}(k)
\end{equation}

where \( M \) is the total number of pixels in the region, \( N \) is the total number of grayscale levels, and \( h_{i,j}(n) \) is the histogram count for intensity level \( n \) \cite{reza2004realization}. To limit the contrast, the clip factor \( \beta \) is introduced:

\begin{equation}
\beta = \frac{M}{N} \left( 1 + \frac{\alpha}{100} (s_{\text{max}} - 1) \right)
\end{equation}

where \( \alpha \) is the clip factor percentage, and \( s_{\text{max}} \) controls the maximum allowable histogram slope \cite{reza2004realization}.

After CLAHE modifies the histogram it redistributes excess histogram counts across lower bins to balance the transformation. The new grayscale value for a pixel is determined by the 4 nearest neighbours using the following equation:

\begin{equation}
p_{\text{new}} = \frac{s}{r+s} \left( \frac{y}{x+y} f_{i-1,j-1}(p_{\text{old}}) + \frac{x}{x+y} f_{i,j-1}(p_{\text{old}}) \right) 
+ \frac{r}{r+s} \left( \frac{y}{x+y} f_{i-1,j}(p_{\text{old}}) + \frac{x}{x+y} f_{i,j}(p_{\text{old}}) \right)
\end{equation}

where \( x, y, r, s \) represent the relative distances between the pixel and the centers of the neighboring regions \cite{reza2004realization}.

The redistribution algorithm ensures no pixel value overshoots so no artificial contrast spikes. This is repeated until all excess counts are evenly spread across the histogram bins. The final step is to merge the enhanced tiles back into a full image so the contrast enhancement is globally consistent.

\begin{figure}[htpb]
    \centering
    \includegraphics[width=\linewidth]{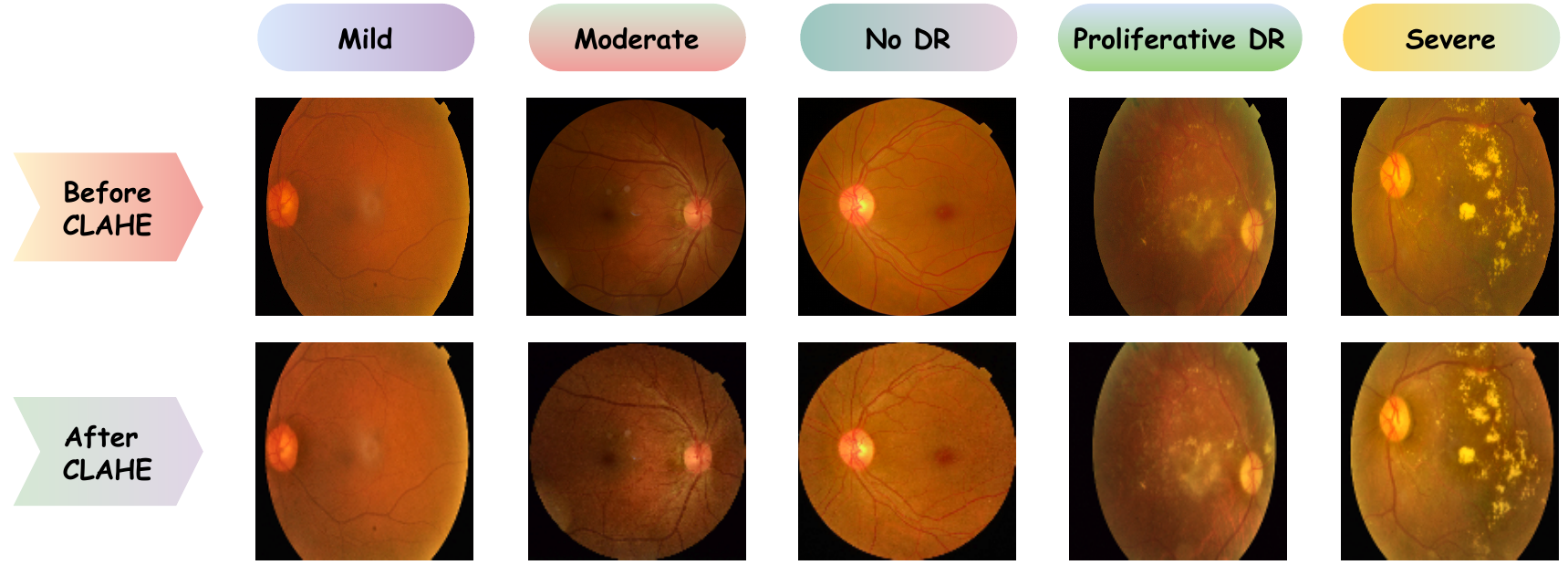}
    \caption{Comparison of Retinal Images Before and After CLAHE Enhancement.}
    \label{Figure CLAHE}
\end{figure}

CLAHE is used in medical imaging  where fine details such as microaneurysms, hemorrhages, exudates in diabetic retinopathy detection need to be preserved while making them visible. By applying CLAHE enhances critical medical features without blowing up bright regions or amplifying noise making it a useful tool in computer aided diagnosis (CAD) and medical image analysis. Here, this approach converts the image to the LAB color space, processing only the L (luminance) channel to maintain color integrity. The CLAHE algorithm is utilized on the L channel with a clipping limit of 0.5 and a tile grid size of (8,8), enabling local contrast improvement while preventing excessive brightness fluctuations. Following enhancement, the modified L channel is re-integrated with the A and B channels, and the image is reverted to the BGR color space. \Cref{Figure CLAHE} presents a visual comparison of retinal images before and after the application of CLAHE.

\subsection{Transfer Learning}
Several popular convolutional neural network architectures were used in this study because of their good performance in image classification tasks. These are VGG16, VGG19, ResNet50V2, MobileNetV2, Xception \cite{sarker2024comprehensive, hasan2024deep}. VGG16 and VGG19 mainly use 3×3 convolutional layers and are commonly used for feature extraction, VGG19 has additional convolutional layers to detect more complex features. ResNet50V2 has residual connections to mitigate the vanishing gradient problem and to improve training stability and speed. Xception uses depthwise separable convolutions to further improve computational efficiency. Finally, MobileNetV2 is optimized for environments with limited computational resources, uses inverted residual structures and bottleneck layers to balance model performance and computational cost.

The convolutional neural networks (CNNs) architectures VGG16, VGG19, ResNet50V2, MobileNetV2, Xception are used with their initial weights from ImageNet pre-trained \cite{sarker2024exploratory}. These models are fine-tuned with the Diabetic Retinopathy dataset, so the pre-learned features are adapted to this specific task. The modified model structure has a flattening layer, dense layers of 1024 and 512 neurons, dropout layers to prevent overfitting and a softmax output layer for classification. Fine-tuning these pre-trained models on the DR disease training dataset improves model performance and reduces the overall training time by leveraging the learned representations from ImageNet initialization. The final dense layer with softmax activation function classifies the input data into 5 classes.

\subsection{Feature Fusion Method}
\begin{figure}[htpb]
    \centering
    \includegraphics[width=\linewidth]{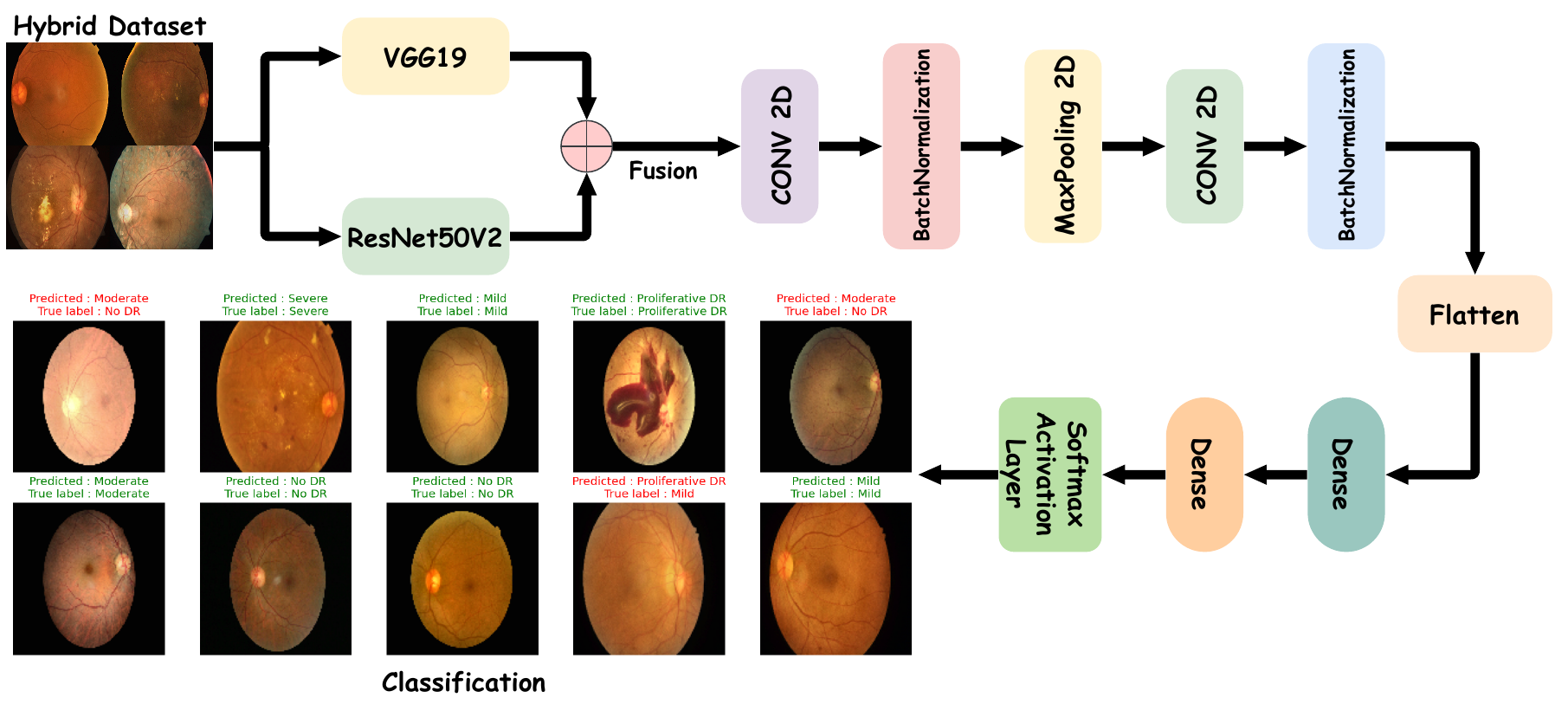}
    \caption{Model Architecture of VR-FuseNet Model}
    \label{Figure VR-FuseNet}
\end{figure}

Feature Fusion is a key technique in pattern recognition applications as it expands the feature space and improves classification accuracy and object detection \cite{khan2021covid}. The combination of multiple feature maps results in a denser and more representative feature space, which improves the discriminative power of the model. However, concatenating feature vectors directly can introduce redundant features and increase computational complexity \cite{mohammad2022deep}. To address this issue a novel feature concatenation approach is used in our method. \Cref{Figure VR-FuseNet} shows the model architecture of our proposed VR-FuseNet model. In our study two pre-trained deep convolutional neural networks (CNNs) VGG19 and ResNet50V2 are used for feature extraction. Let the extracted deep CNN feature vectors from VGG19 and ResNet50V2 be $m_1$ and $m_2$ respectively. The dimensions of these feature vectors are $q \times r$ and $q \times s$ where $q$ is the number of images and $r$ and $s$ are the respective lengths of the feature attributes. In our case, both $r$ and $s$ are 512 and 2048, so the feature dimensions are $q \times 512$ and $q \times 2048$ for both networks. To pad feature space dimensions before concatenation mean is computed and added with it. The transformation of feature vectors into time series can be expressed as follows :
\begin{equation}
z_1 = 1^T m_1
\end{equation}
\begin{equation}
z_2 = 2^T m_2
\end{equation}

The maximum covariance of $m_1$ and $m_2$ is defined as:

\begin{equation}
\hat{f} = \text{Cov}[z_1, z_2]
\end{equation}
\begin{equation}
\hat{f} = \text{Cov} (1^T m_1, 1^T m_1)
\end{equation}
\begin{equation}
\hat{f} = \frac{1}{n-1} \text{Cov} (1^T m_1, 1^T m_1)
\end{equation}

The covariance of the final concatenated feature space is given by:

\begin{equation}
\hat{f} = \frac{1}{(F_{1,2})} 2
\end{equation}

where:

\begin{equation}
F_{1,2} = \frac{1}{n-1} (m_1 m_2^T)
\end{equation}
Here $F_{1,2}$ is the covariance between the two feature maps with features indexed by $i$ and $j$. The final concatenated feature space using parallel maximum covariance strategy gives a feature vector of dimension $q \times 2560$. This dense representation helps in feature selection and avoid feature redundancy. By using this feature fusion approach our model benefits from the strengths of both VGG19 and ResNet50V2 feature extractors and gets better performance and efficiency.

After analysis we select VGG19 and ResNet50V2 for feature fusion as they have complementary strengths and perform better than other models. VGG19 with its deep yet simple architecture of stacked convolutional layers with small 3x3 filters is good at capturing fine grained local features. ResNet50V2 with its residual connections allows training of deeper networks by mitigating the vanishing gradient problem and is good at capturing abstract high level features. By fusing features from both models we leverage the local detail extraction capability of VGG19 and the global contextual understanding of ResNet50V2 and create a more robust and comprehensive feature representation. This fusion outperforms individual models or other combinations as it integrates both depth and residual learning and helps the model to generalize across diverse retinal images and improve classification accuracy.

The VR-FuseNet Model is a hybrid deep learning model which combines two convolutional neural networks (CNNs) VGG19 and ResNet50V2 to leverage their complementary feature extraction capabilities. ResNet50V2 with its deep residual learning captures high level abstract features and VGG19 captures fine grained spatial features in retinal images. The input hybrid dataset images are passed through both the networks separately, each network extracts unique and meaningful features. The model then uses fusion mechanism to combine the features from both the networks so that the best of both the architectures is used for better classification.

After fusion, the fused feature maps undergo further refinement through additional Conv2D layers, batch normalization, and max-pooling operations to enhance spatial hierarchies. Then the model passes the combined features through fully connected (dense) layers which refines the learned representations. Specifically it has a flatten layer followed by 2 dense layers of 256 and 64 neurons with dropout layers to prevent overfitting so the model can generalize to different datasets.  The last dense layer of the model uses softmax activation function which assigns probabilities to different DR severity levels and  classifies the input data into 5 classes. This way the model can distinguish different stages of DR, mild, moderate, severe, proliferative DR and normal cases with no DR. By combining two powerful CNN architectures, VR-FuseNet improves feature extraction and hence better generalization and higher accuracy. The model can learn from diverse set of retinal images and is a promising solution for automated DR diagnosis which can help ophthalmologists in early detection and treatment planning.

\section{Result Analysis}
\label{Result_Analysis}
\subsection{Evaluation Metrices}
To evaluate the proposed VR-FuseNet model for diabetic retinopathy classification, we used a set of well-known performance metrics. These metrics give insights into different aspects of the model’s predictability, such as overall correctness, true cases detection and balance between precision and recall. Below are the mathematical definitions and explanations of each metric used in this study.

\begin{itemize}
    \item \textbf{Accuracy}: The proportion of true results (both true positives and true negatives) in the population, computed as:
    \begin{equation}
        \text{Accuracy} = \frac{TP + TN}{TP + TN + FP + FN}
        \label{accuracy}
    \end{equation}
    
    \item \textbf{Precision}: The ratio of true positive to total predicted positives, calculated as:
    \begin{equation}
        \text{Precision} = \frac{TP}{TP + FP}
        \label{Precision}
    \end{equation}
    
    \item \textbf{Recall}: Also known as true positive rate, the ratio of true positive to all actual positives, expressed as:
    \begin{equation}
        \text{Recall} = \frac{TP}{TP + FN}
        \label{Recall}
    \end{equation}
    
    \item \textbf{F1-Score}: The harmonic mean of precision and recall, formulated as:
    \begin{equation}
        \text{F1-Score} = 2 \times \frac{\text{Precision} \times \text{Recall}}{\text{Precision} + \text{Recall}}
        \label{F1-Score}
    \end{equation}

\item \textbf{Receiver Operating Characteristic Curve}: ROC curve is a fundamental tool to evaluate the performance of classification models, especially in binary classification, such as presence or absence of a disease. It provides a graphical representation by plotting the True Positive Rate (TPR or Recall) against the False Positive Rate (FPR) at different threshold settings \cite{jocelyn2018}. The FPR is calculated as:

\begin{equation} \text{FPR} = \frac{FP}{FP + TN} \label{ROC} \end{equation}

Originally developed for military radar signal detection in 1940s, ROC curve has become a standard metric in machine learning and medical diagnostics. It helps to balance sensitivity (TPR) and specificity (1 - FPR), giving insights into the model’s discriminative ability. The Area Under the ROC Curve (AUC) is a quantitative measure of this performance. AUC values range from 0.5 (random guessing) to 1.0 (perfect classification), with higher values indicating better model reliability and accuracy. In this study, AUC-ROC is used to evaluate the proposed model’s ability to detect diabetic retinopathy. ROC curves closer to the upper left corner of the plot space indicates better classification performance, the model is able to correctly identify both positive and negative cases at different thresholds \cite{hoo2017roc}.

\end{itemize}

\subsection{Comparative Analysis Across Datasets}
Results of several deep learning models on 5 publicly available diabetic retinopathy (DR) datasets—APTOS 2019, DDR, IDRiD, Messidor 2, and Retino—are shown in \Cref{tab:Tableresult}. Models are evaluated in terms of accuracy, precision, recall and F1-score. Models are trained with a learning rate of 1e-5 to ensure convergence during optimization. Different datasets require different batch sizes; 16, 32 and 128 are used based on computational constraints and dataset complexity. 50 to 110 epochs are used to avoid overfitting and learn features.

\begin{table}[htbp]
\caption{Performance Metrics for Different Models}
\label{tab:Tableresult}
\centering
\renewcommand{\arraystretch}{1.3}
\begin{tabular}{|c|c|c|c|c|c|c|c|}
\hline
\textbf{Dataset} & \textbf{Model} & \textbf{\begin{tabular}[c]{@{}c@{}}Batch\\ Size\end{tabular}} & \textbf{Epoch} & \textbf{\begin{tabular}[c]{@{}c@{}}Accuracy\\ (\%)\end{tabular}} & \textbf{\begin{tabular}[c]{@{}c@{}}Precision\\ (\%)\end{tabular}} & \textbf{\begin{tabular}[c]{@{}c@{}}Recall\\ (\%)\end{tabular}} & \textbf{\begin{tabular}[c]{@{}c@{}}F1 Score\\ (\%)\end{tabular}} \\ \hline
\multirow{5}{*}{\textbf{APTOS 2019}} & VGG16 & \multirow{5}{*}{32} & 70 & 95.400 & 95.628 & 94.511 & 95.019 \\ \cline{2-2} \cline{4-8} 
 & \textbf{VGG19} & & 80 & \textbf{96.000} & \textbf{96.135} & \textbf{95.164} & \textbf{95.620} \\ \cline{2-2} \cline{4-8} 
 & ResNet50V2 & & 50 & 92.400 & 92.173 & 91.348 & 91.678 \\ \cline{2-2} \cline{4-8} 
 & MobileNetV2 & & \multirow{2}{*}{70} & 92.200 & 91.562 & 90.489 & 90.870 \\ \cline{2-2} \cline{5-8} 
 & Xception & & & 92.000 & 91.605 & 90.720 & 91.076 \\ \cline{1-3} \cline{4-8} 
\multirow{5}{*}{\textbf{DDR}} & \textbf{VGG16} & \multirow{5}{*}{128} & \multirow{5}{*}{70} & \textbf{90.630} & \textbf{93.023} & \textbf{93.227} & \textbf{93.122} \\ \cline{2-2} \cline{5-8} 
 & VGG19 & & & 90.027 & 92.630 & 93.015 & 92.727 \\ \cline{2-2} \cline{5-8} 
 & ResNet50V2 & & & 87.726 & 90.567 & 90.365 & 90.354 \\ \cline{2-2} \cline{5-8} 
 & MobileNetV2 & & & 90.301 & 92.742 & 92.051 & 92.264 \\ \cline{2-2} \cline{5-8} 
 & Xception & & & 89.151 & 91.567 & 91.387 & 91.441 \\ \cline{1-3} \cline{4-8} 
\multirow{5}{*}{\textbf{IDRiD}} & VGG16 & \multirow{5}{*}{16} & \multirow{3}{*}{70} & 79.688 & 80.770 & 79.692 & 79.844 \\ \cline{2-2} \cline{5-8} 
 & \textbf{VGG19} & & & \textbf{81.250} & \textbf{84.902} & \textbf{82.082} & \textbf{82.924} \\ \cline{2-2} \cline{5-8} 
 & ResNet50V2 & & & 76.562 & 78.452 & 77.042 & 77.600 \\ \cline{2-2} \cline{4-8} 
 & MobileNetV2 & & 100 & 78.125 & 82.094 & 78.581 & 79.340 \\ \cline{2-2} \cline{4-8} 
 & Xception & & 50 & 73.438 & 72.873 & 72.686 & 72.655 \\ \cline{1-3} \cline{4-8} 
\multirow{5}{*}{\textbf{Messidor 2}} & VGG16 & \multirow{5}{*}{16} & \multirow{3}{*}{60} & 90.871 & 87.272 & 87.561 & 87.383 \\ \cline{2-2} \cline{5-8} 
 & \textbf{VGG19} & & & \textbf{91.286} & \textbf{88.038} & \textbf{88.190} & \textbf{88.010} \\ \cline{2-2} \cline{5-8} 
 & ResNet50V2 & & & 88.797 & 84.869 & 84.737 & 84.561 \\ \cline{2-2} \cline{4-8} 
 & MobileNetV2 & & 90 & 87.552 & 82.655 & 83.295 & 82.298 \\ \cline{2-2} \cline{4-8} 
 & Xception & & 60 & 87.344 & 82.847 & 83.128 & 82.664 \\ \cline{1-3} \cline{4-8} 
\multirow{5}{*}{\textbf{Retino}} & VGG16 & \multirow{5}{*}{16} & \multirow{2}{*}{70} & 92.265 & 94.903 & 94.545 & 94.607 \\ \cline{2-2} \cline{5-8} 
 & \textbf{VGG19} & & & \textbf{93.923} & \textbf{95.698} & \textbf{95.157} & \textbf{95.357} \\ \cline{2-2} \cline{4-8} 
 & ResNet50V2 & & \multirow{2}{*}{110} & 87.845 & 90.928 & 90.802 & 90.834 \\ \cline{2-2} \cline{5-8} 
 & MobileNetV2 & & & 87.845 & 91.078 & 91.048 & 91.025 \\ \cline{2-2} \cline{4-8} 
 & Xception & & 70 & 88.398 & 91.481 & 91.682 & 91.554 \\ \hline
\end{tabular}
\end{table}

\begin{figure}[htbp]
    \centering
    
    \begin{subfigure}[b]{0.48\textwidth}
        \centering
        \includegraphics[width=\textwidth, keepaspectratio]{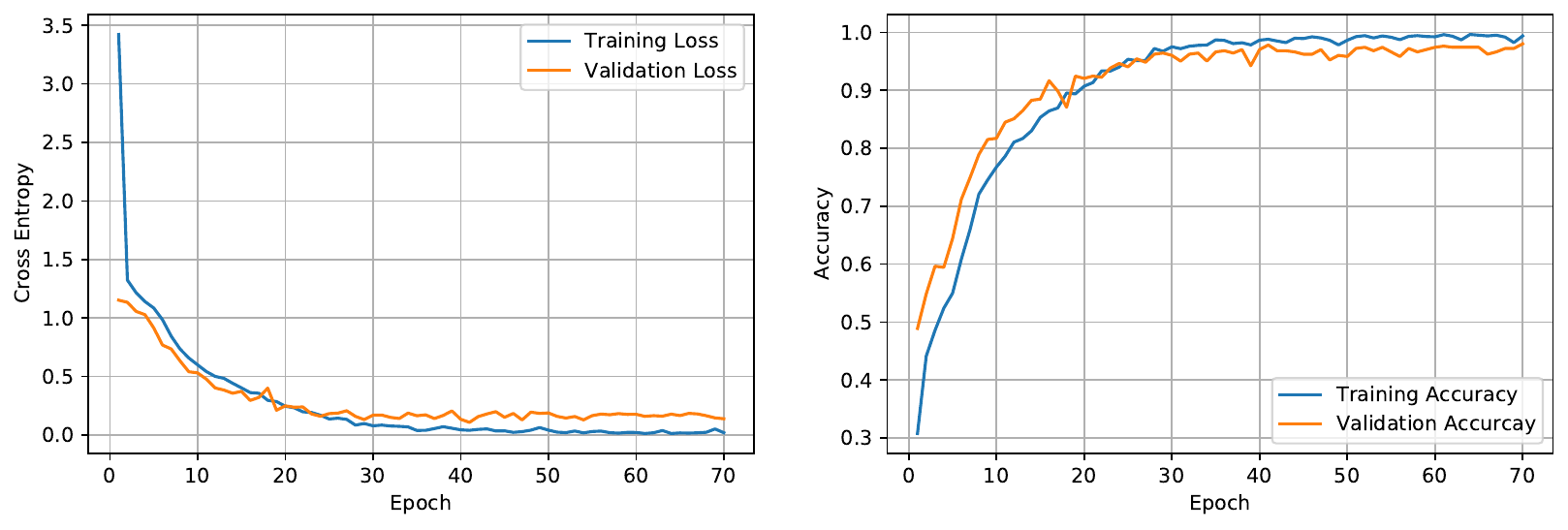} 
        \caption{APTOS 2019 (VGG16)}
    \end{subfigure}
    \hspace{0.02\textwidth} 
    \begin{subfigure}[b]{0.48\textwidth}
        \centering
        \includegraphics[width=\textwidth, keepaspectratio]{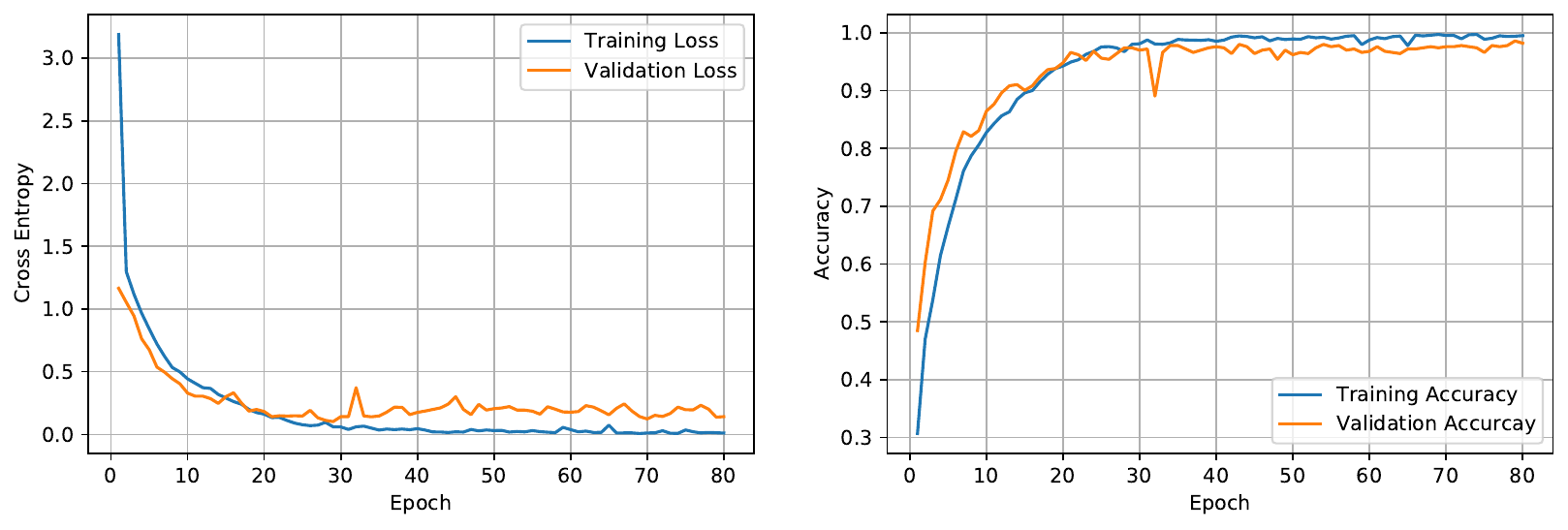} 
        \caption{APTOS 2019 (VGG19)}
    \end{subfigure}
    
    \begin{subfigure}[b]{0.48\textwidth}
        \centering
        \includegraphics[width=\textwidth, keepaspectratio]{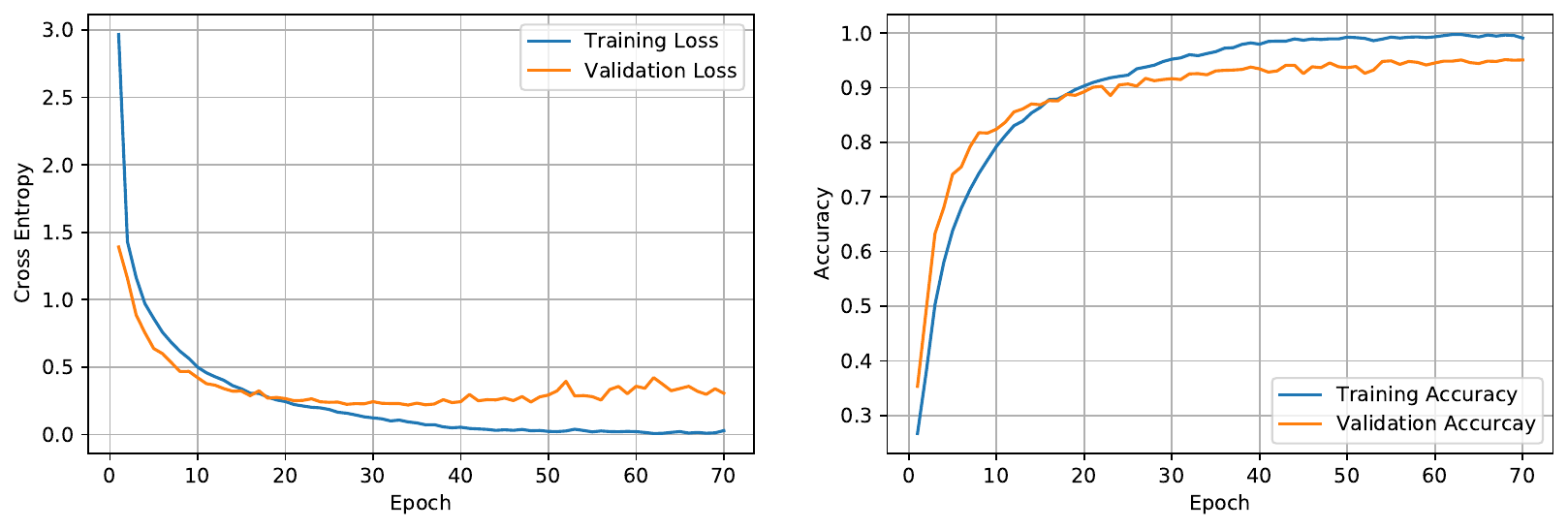} 
        \caption{DDR (VGG16)}
    \end{subfigure}
    \hspace{0.02\textwidth} 
    \begin{subfigure}[b]{0.48\textwidth}
        \centering
        \includegraphics[width=\textwidth, keepaspectratio]{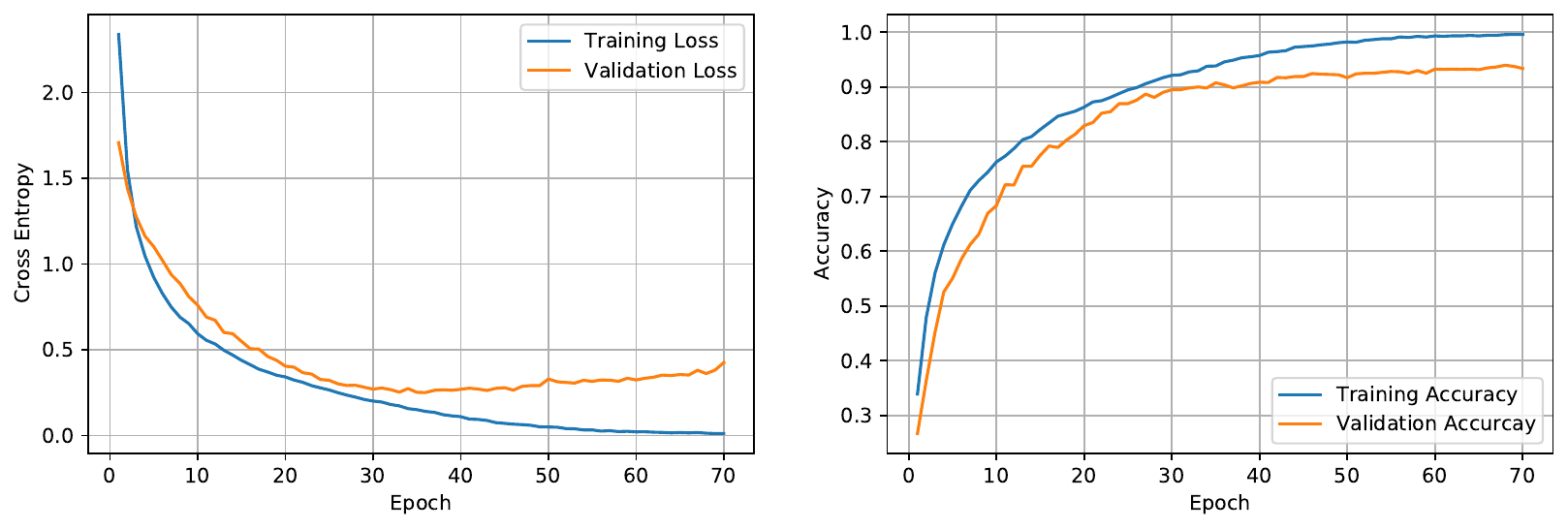} 
        \caption{DDR (MobileNetV2)}
    \end{subfigure}

    \begin{subfigure}[b]{0.48\textwidth}
        \centering
        \includegraphics[width=\textwidth, keepaspectratio]{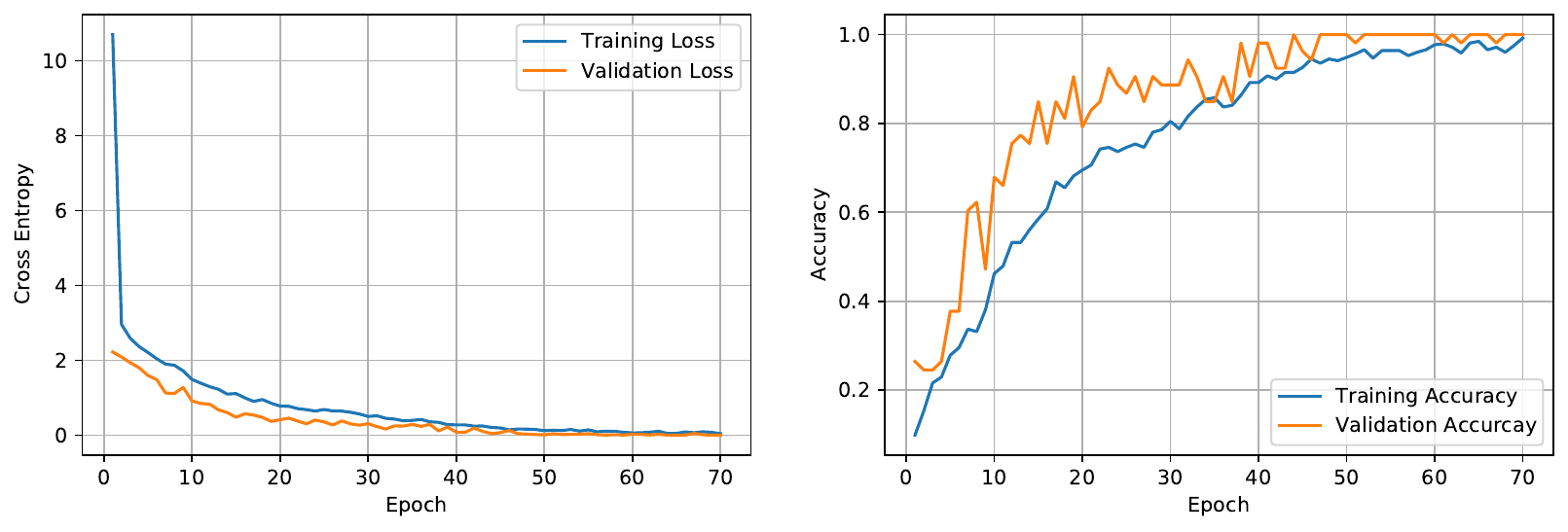} 
        \caption{IDRiD (VGG16)}
    \end{subfigure}
    \hspace{0.02\textwidth} 
    \begin{subfigure}[b]{0.48\textwidth}
        \centering
        \includegraphics[width=\textwidth, keepaspectratio]{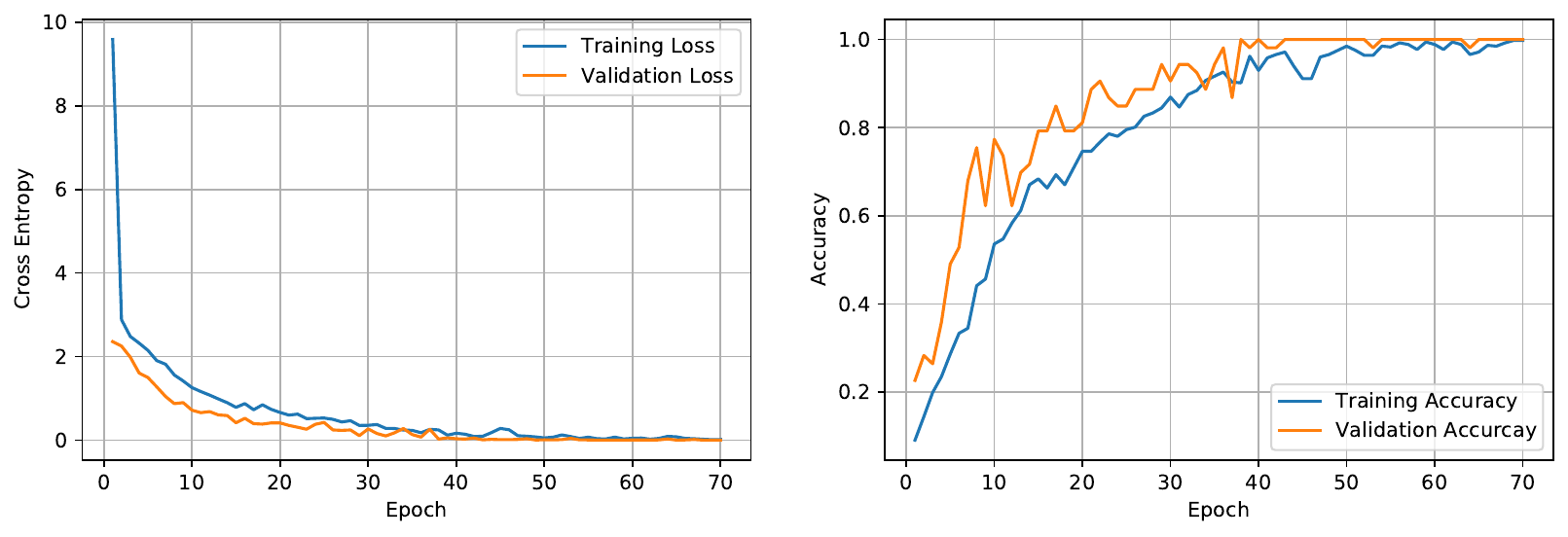} 
        \caption{IDRiD (VGG19)}
    \end{subfigure}

    \begin{subfigure}[b]{0.48\textwidth}
        \centering
        \includegraphics[width=\textwidth, keepaspectratio]{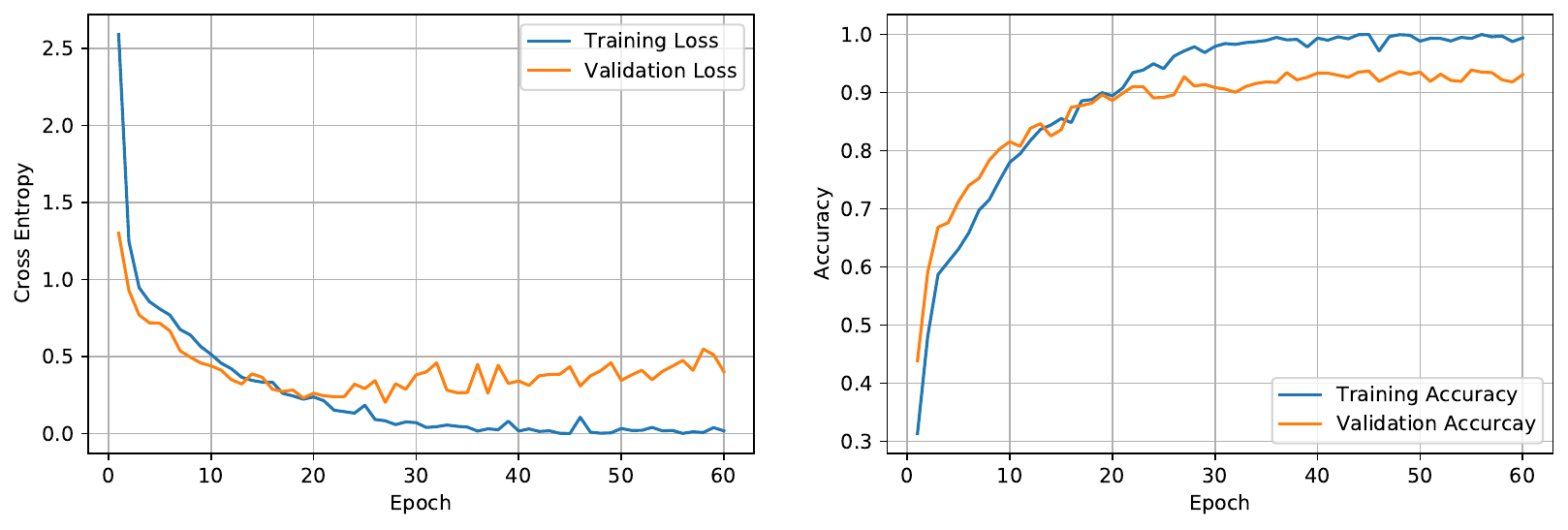} 
        \caption{Messidor 2 (VGG16)}
    \end{subfigure}
    \hspace{0.02\textwidth} 
    \begin{subfigure}[b]{0.48\textwidth}
        \centering
        \includegraphics[width=\textwidth, keepaspectratio]{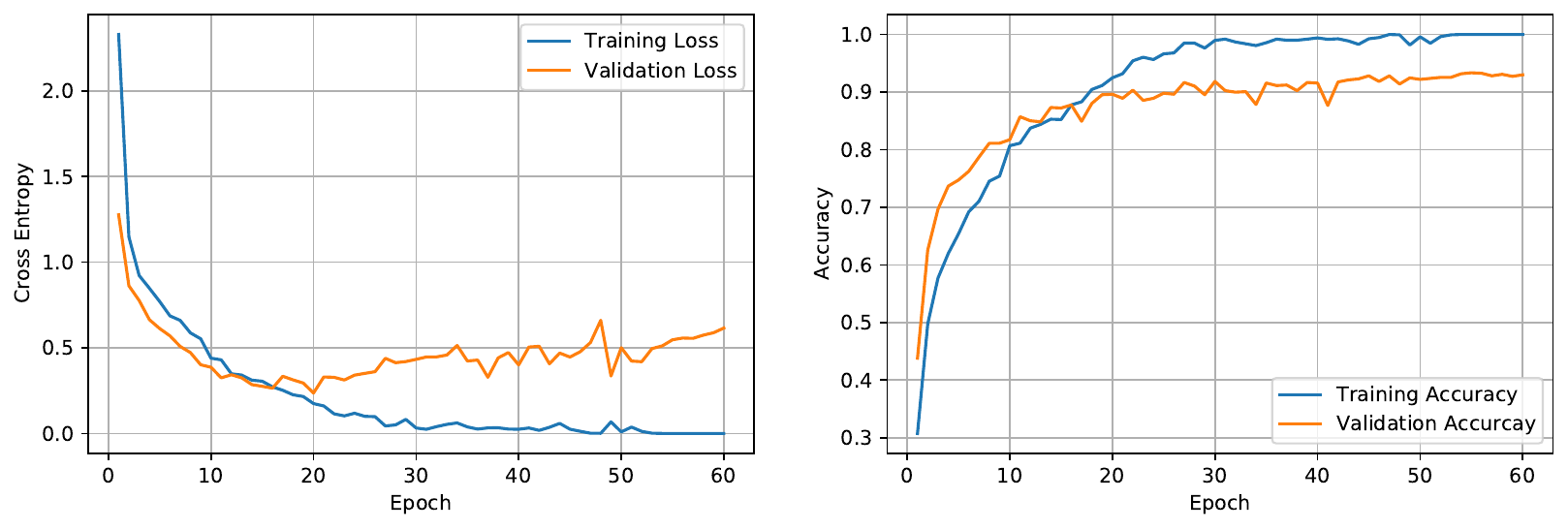} 
        \caption{Messidor 2 (VGG19)}
    \end{subfigure}

    \begin{subfigure}[b]{0.48\textwidth}
        \centering
        \includegraphics[width=\textwidth, keepaspectratio]{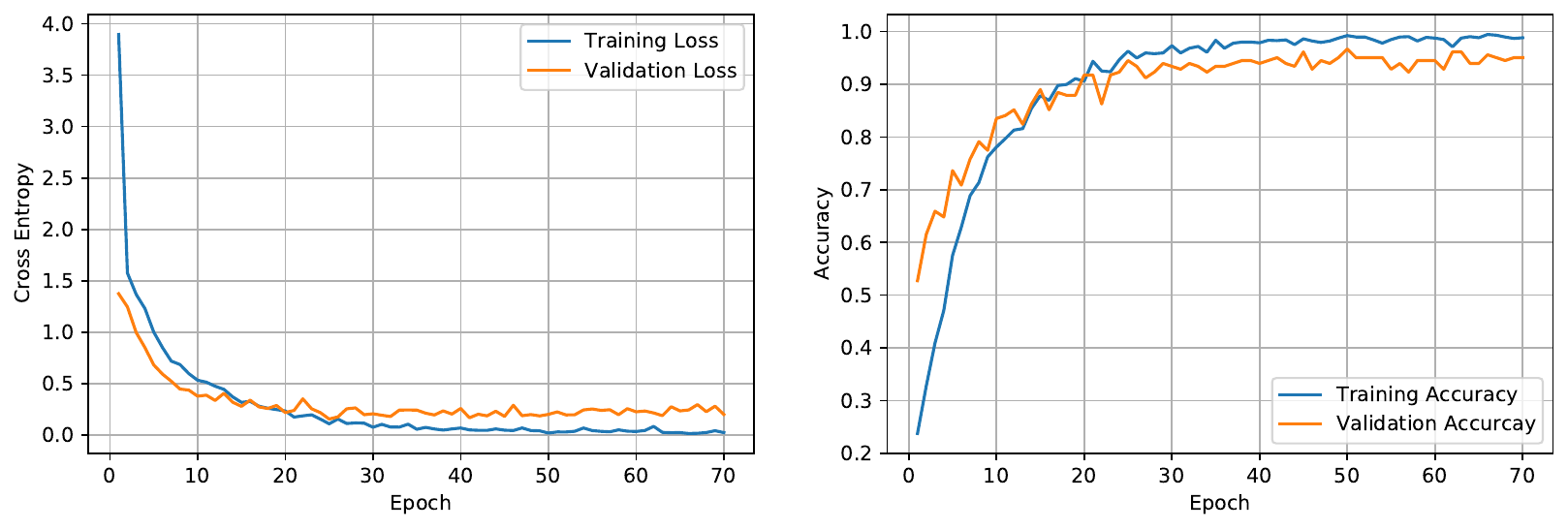} 
        \caption{Retino (VGG16)}
    \end{subfigure}
    \hspace{0.02\textwidth} 
    \begin{subfigure}[b]{0.48\textwidth}
        \centering
        \includegraphics[width=\textwidth, keepaspectratio]{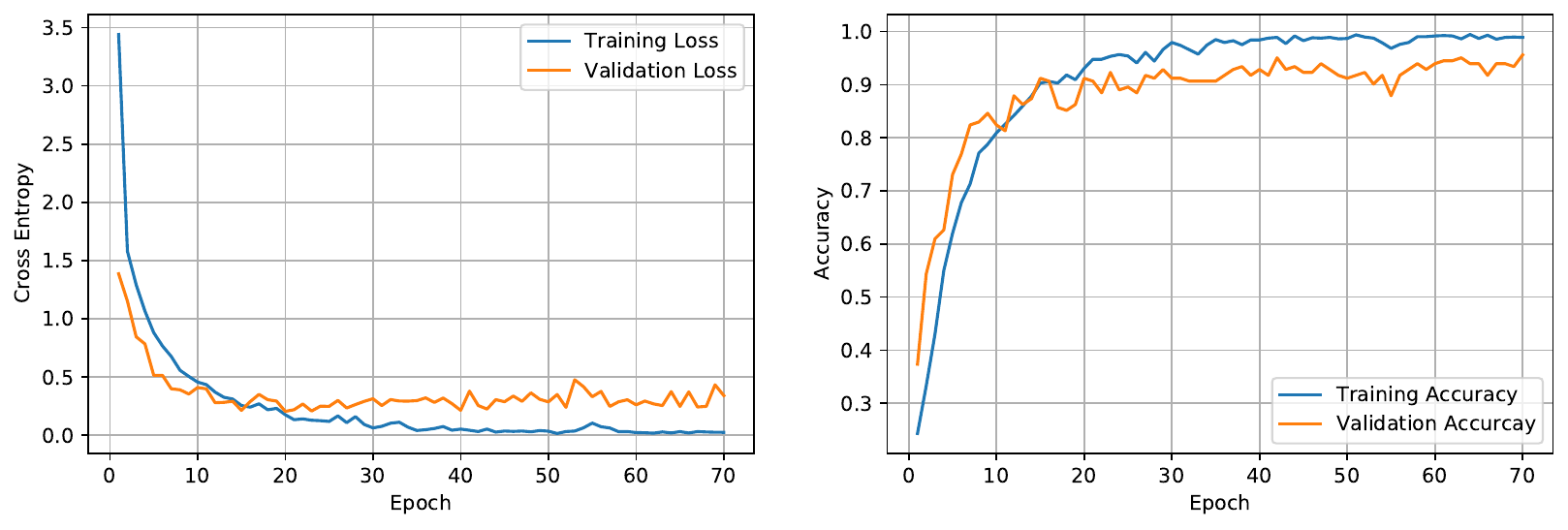} 
        \caption{Retino (VGG19)}
    \end{subfigure}
    
    \caption{Individual Dataset Loss and Accuracy Curve}
    \label{Individual_Dataset_Loss_and_Accuracy_Curve}
\end{figure}

\begin{figure}[htbp]
    \centering
    
    \begin{subfigure}[b]{0.3\textwidth}
        \centering
        \includegraphics[scale = 0.3]{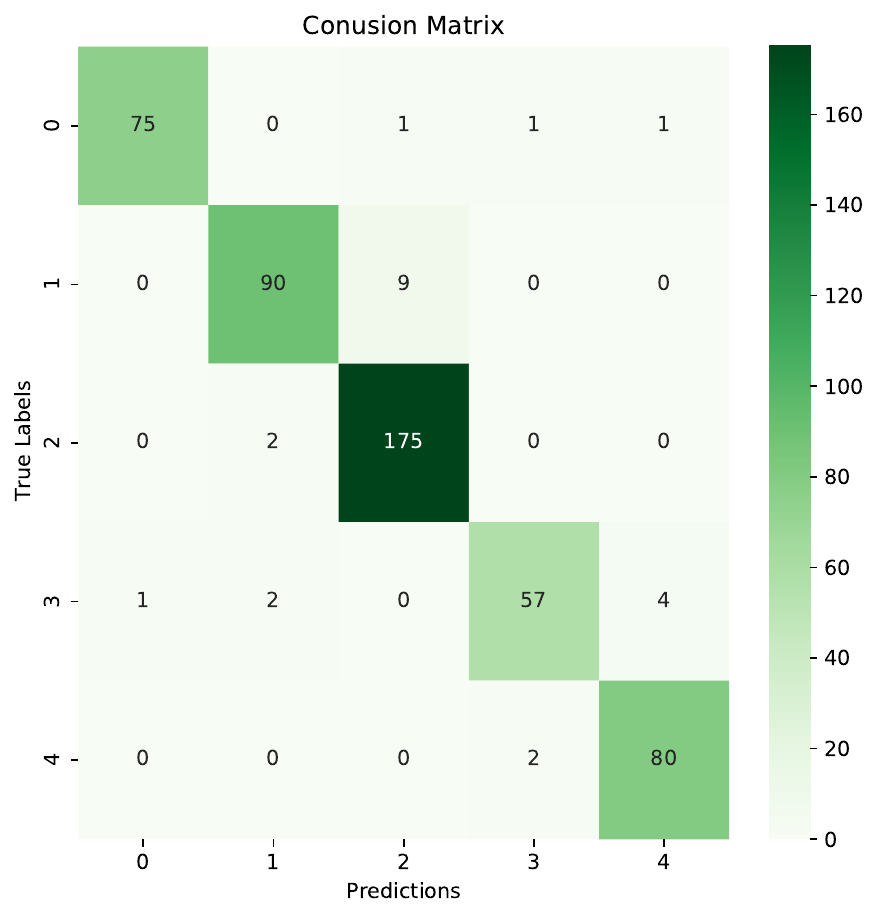} 
        \caption{APTOS 2019 (VGG16)}
    \end{subfigure}
    \hspace{0.02\textwidth} 
    \begin{subfigure}[b]{0.3\textwidth}
        \centering
        \includegraphics[scale = 0.3]{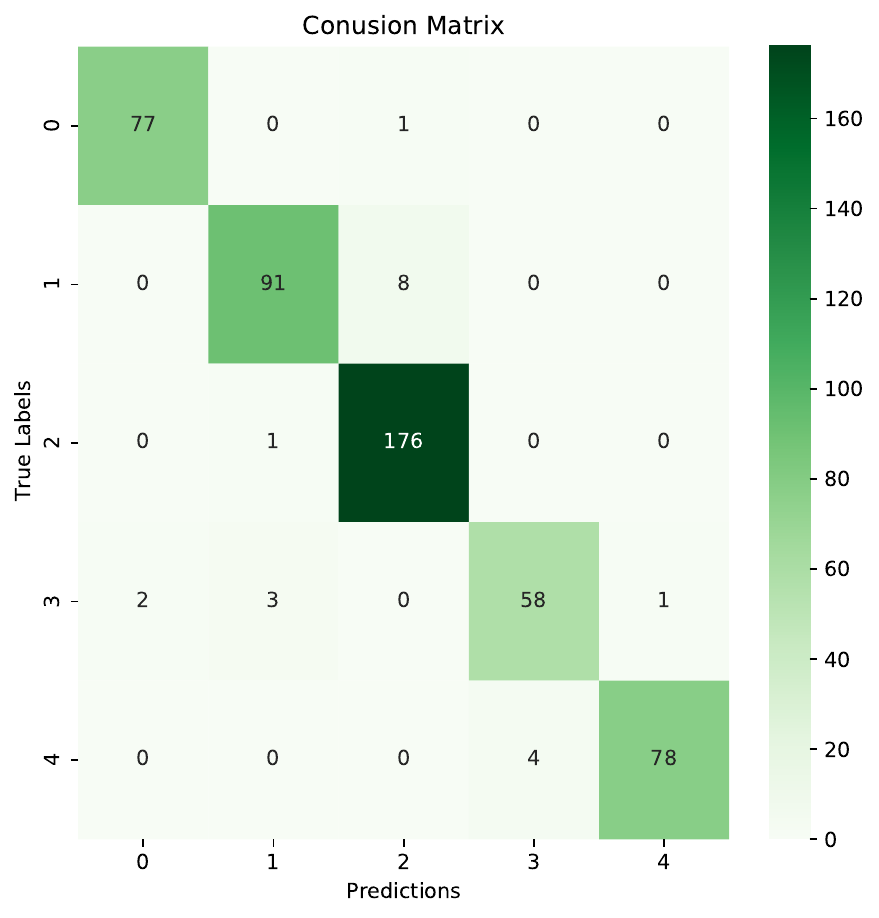} 
        \caption{APTOS 2019 (VGG19)}
    \end{subfigure}
    \hspace{0.02\textwidth} 
    \begin{subfigure}[b]{0.3\textwidth}
        \centering
        \includegraphics[scale = 0.3]{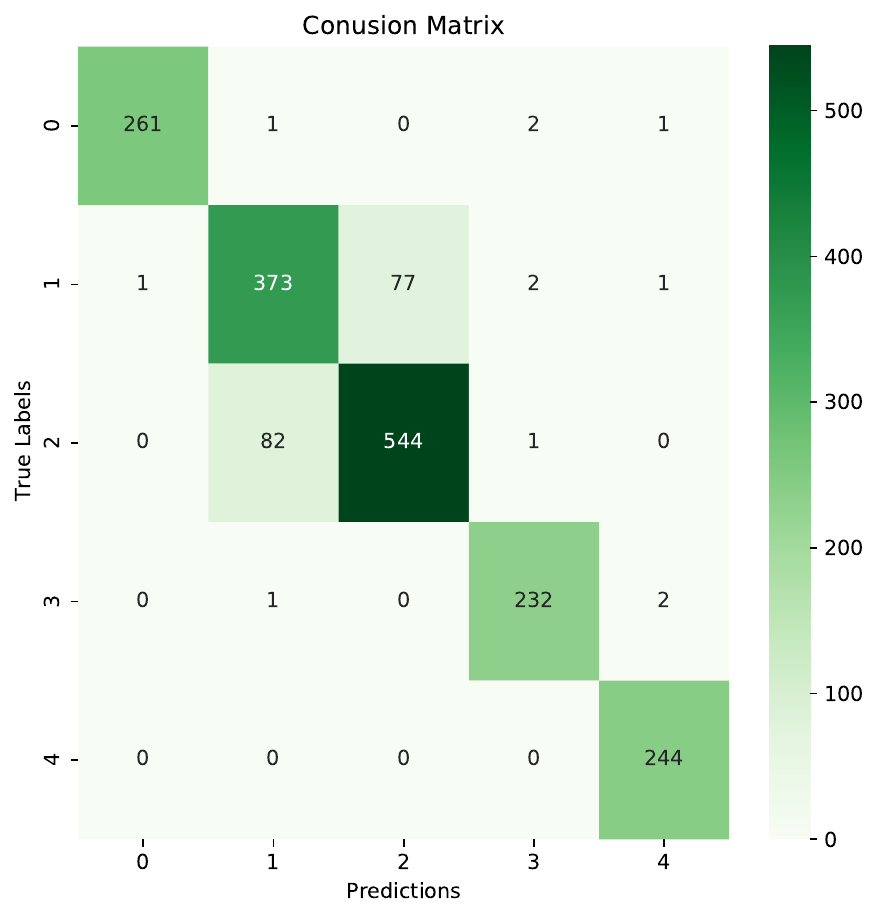} 
        \caption{DDR (VGG16)}
    \end{subfigure}
    \hspace{0.02\textwidth} 
    \begin{subfigure}[b]{0.3\textwidth}
        \centering
        \includegraphics[scale = 0.3]{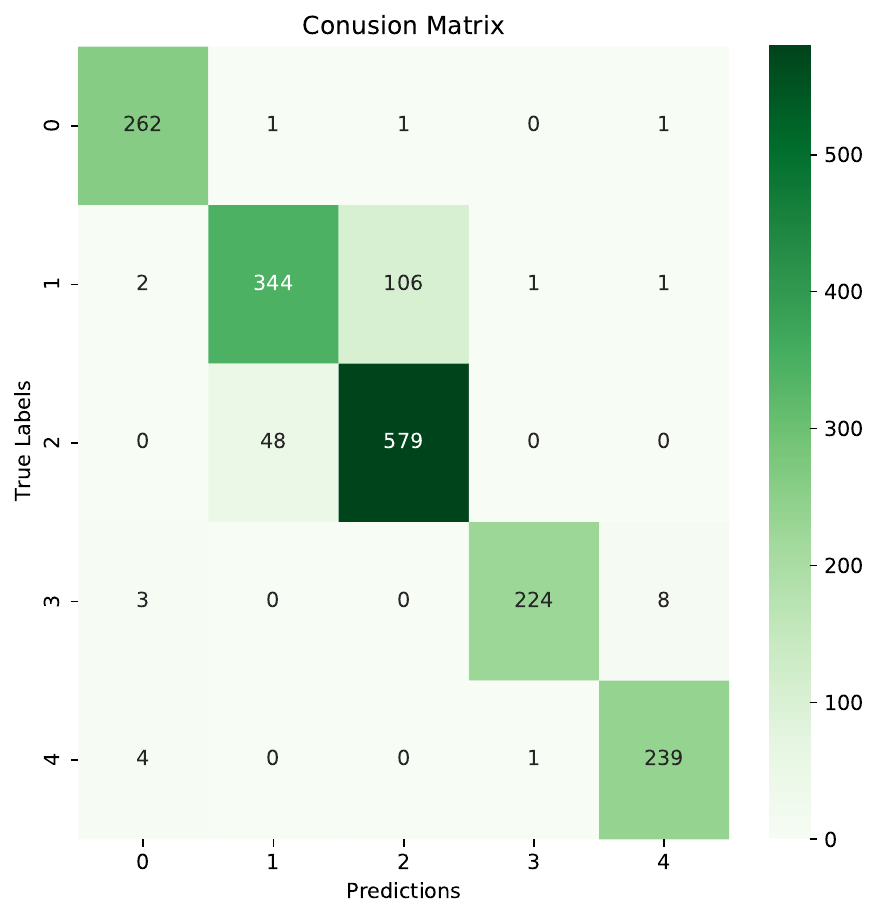} 
        \caption{DDR (MobileNetV2)}
    \end{subfigure}
    \hspace{0.02\textwidth} 
    \begin{subfigure}[b]{0.3\textwidth}
        \centering
        \includegraphics[scale = 0.3]{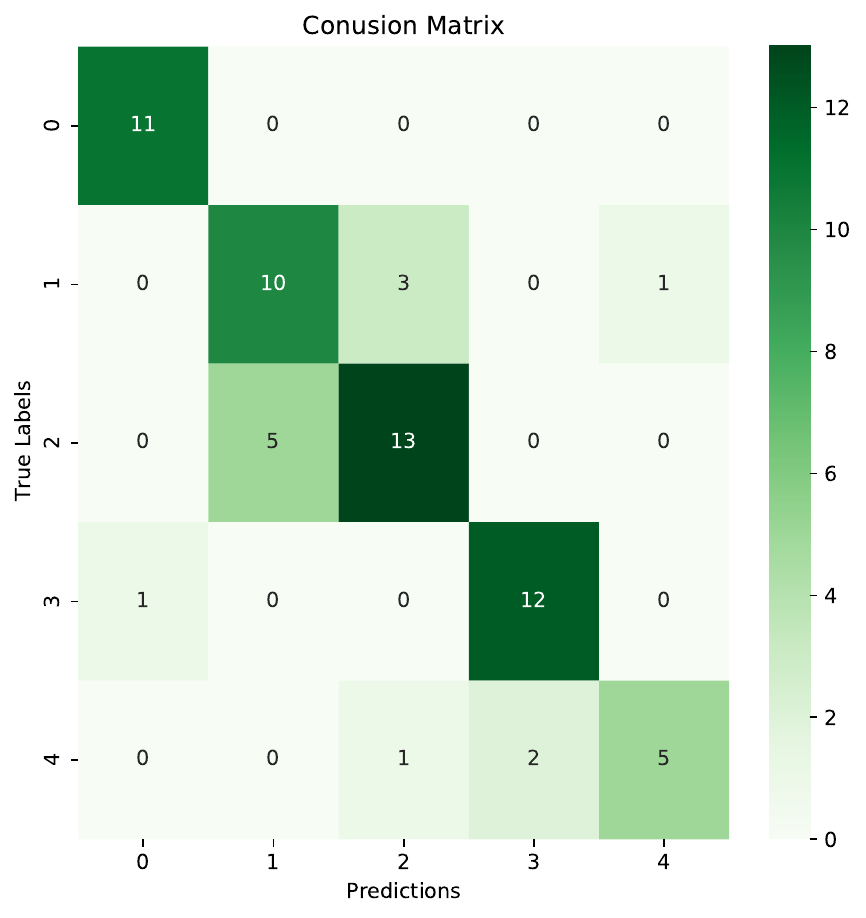} 
        \caption{IDRiD (VGG16)}
    \end{subfigure}
    \hspace{0.02\textwidth} 
    \begin{subfigure}[b]{0.3\textwidth}
        \centering
        \includegraphics[scale = 0.3]{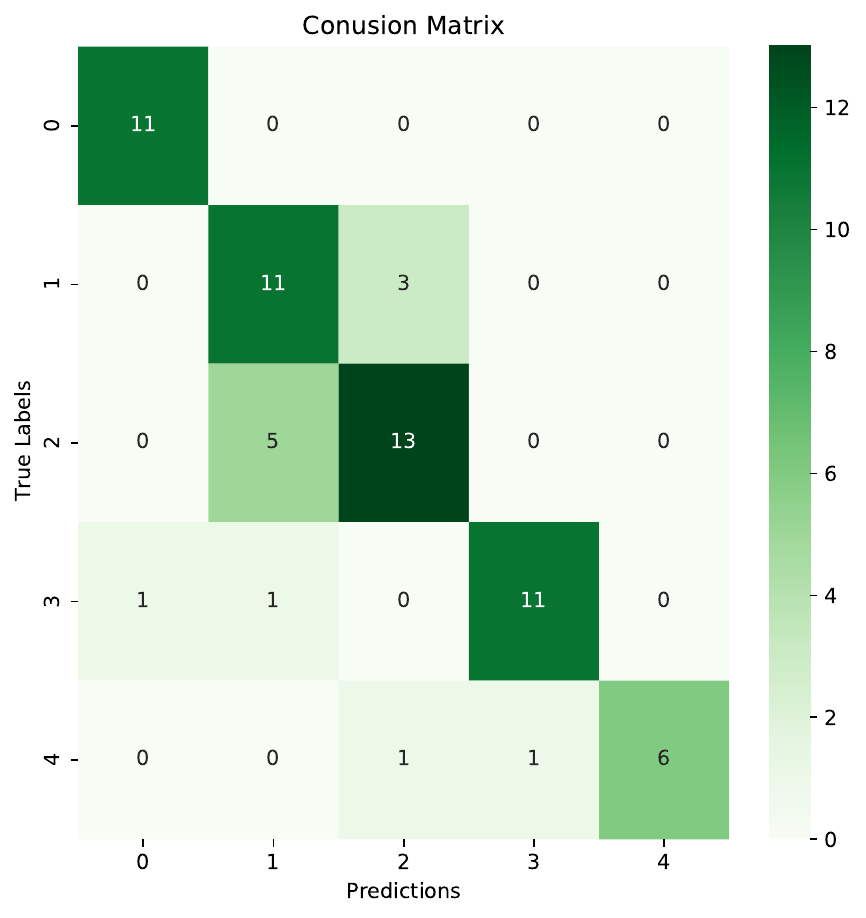} 
        \caption{IDRiD (VGG19)}
    \end{subfigure}
    \hspace{0.02\textwidth} 
    \begin{subfigure}[b]{0.3\textwidth}
        \centering
        \includegraphics[scale = 0.3]{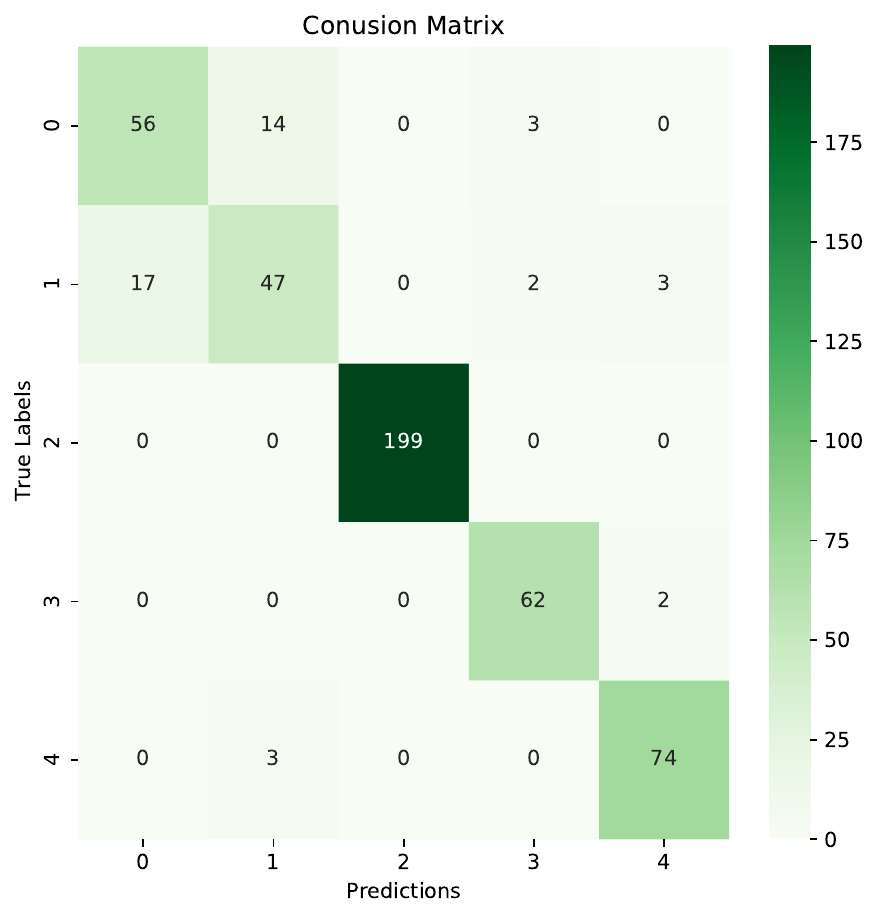} 
        \caption{Messidor 2 (VGG16)}
    \end{subfigure}
    \hspace{0.02\textwidth} 
    \begin{subfigure}[b]{0.3\textwidth}
        \centering
        \includegraphics[scale = 0.3]{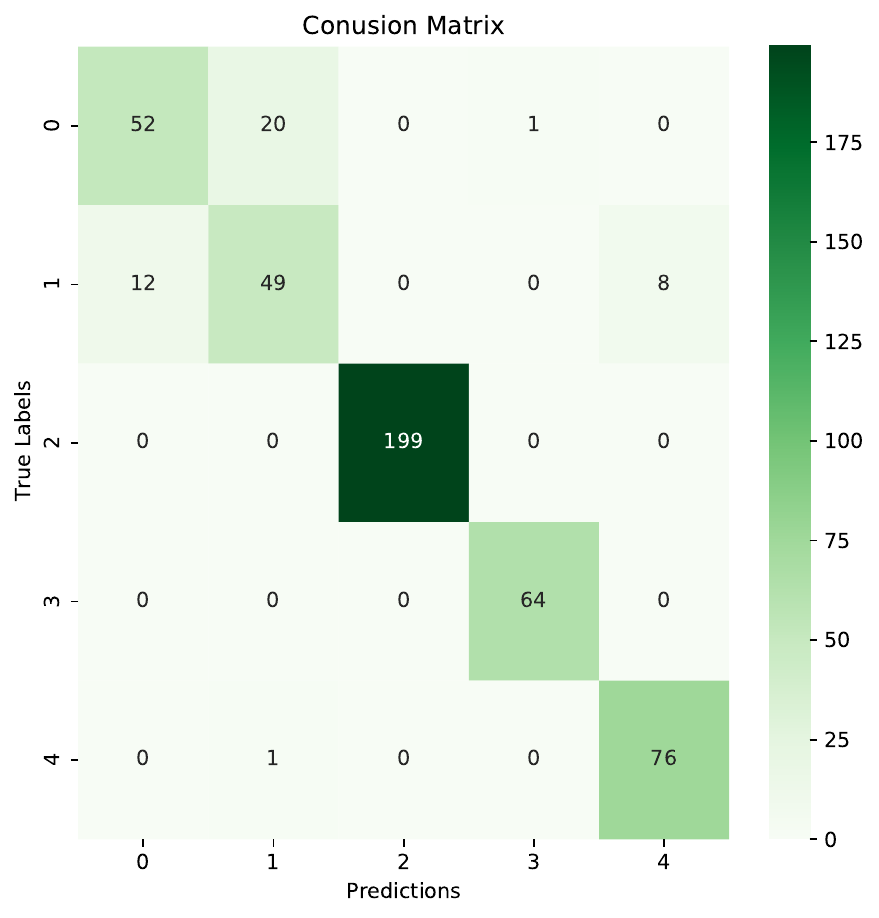} 
        \caption{Messidor 2 (VGG19)}
    \end{subfigure}
    \hspace{0.02\textwidth} 
    \begin{subfigure}[b]{0.3\textwidth}
        \centering
        \includegraphics[scale = 0.3]{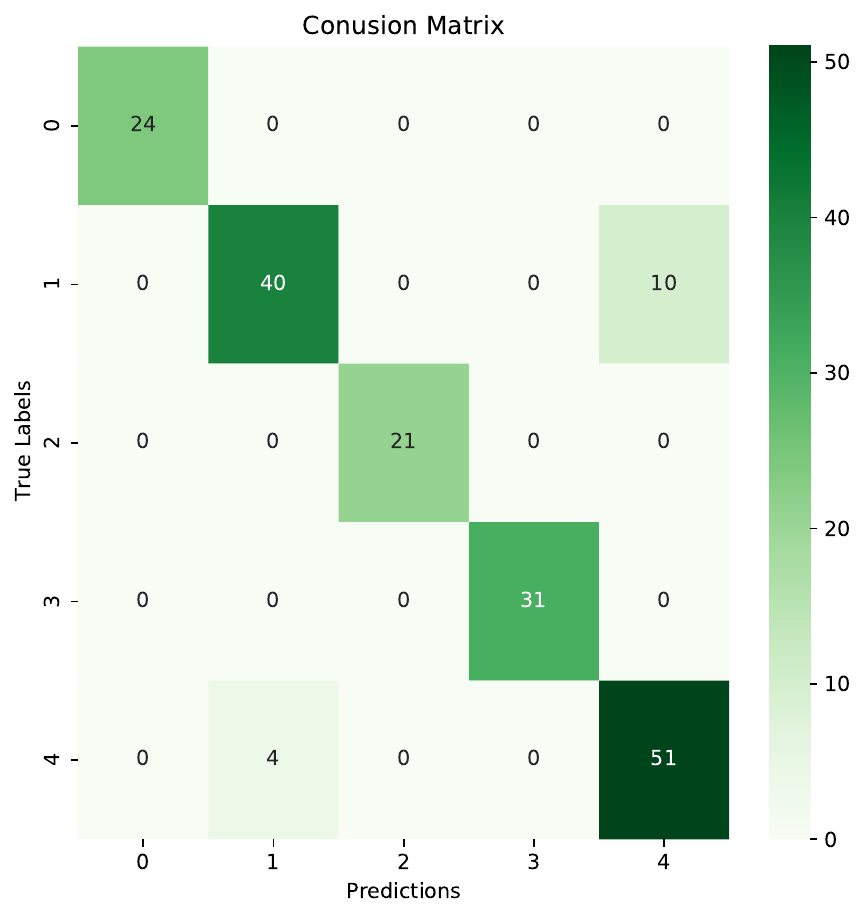} 
        \caption{Retino (VGG16)}
    \end{subfigure}
    \hspace{0.02\textwidth} 
    \begin{subfigure}[b]{0.3\textwidth}
        \centering
        \includegraphics[scale = 0.3]{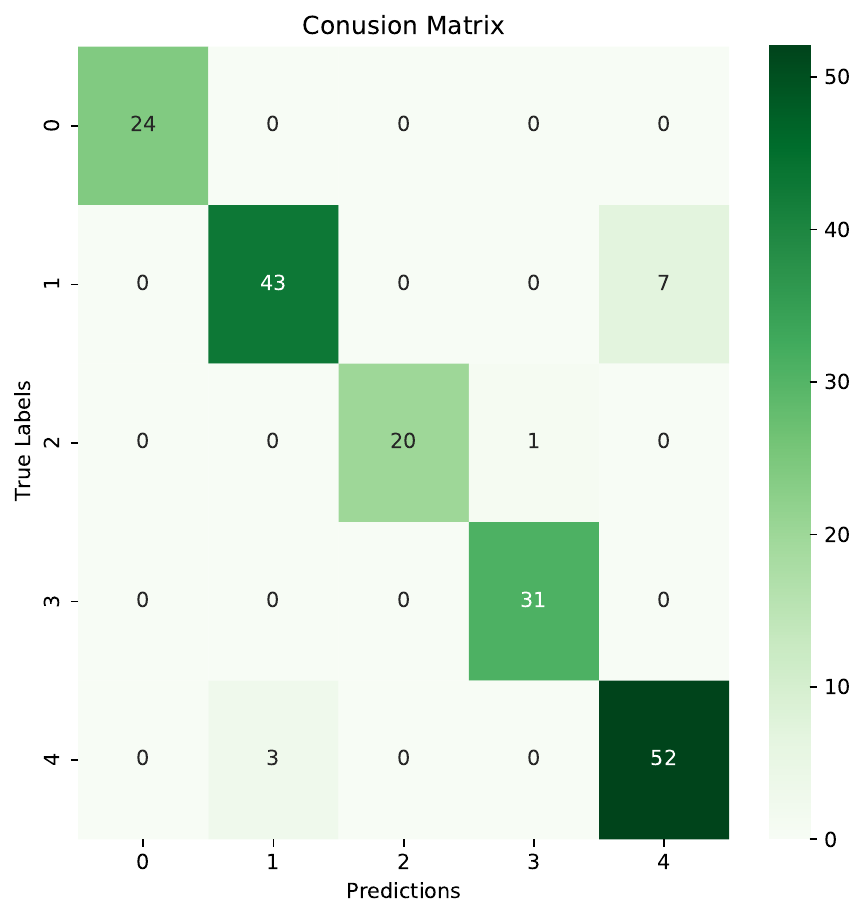} 
        \caption{Retino (VGG19)}
    \end{subfigure}
    
    \caption{Individual Dataset Confusion Matrix}
    \label{Individual_Dataset_Confusion_Matrix}
\end{figure}

The APTOS 2019 dataset shows strong performance across all models, with VGG19 achieving the highest accuracy at 96.000\%. VGG16 follows closely at 95.400\%. The ResNet50V2 model achieved 92.400\%, while MobileNetV2 and Xception models showed slightly lower accuracies at 92.200\% and 92.000\% respectively. This dataset appears to be well-suited for the VGG architecture, with both VGG variants outperforming the other models by a notable margin of approximately 3-4\%.
For the DDR dataset, VGG16 leads with an accuracy of 90.630\%, followed by VGG19 at 90.027\%. MobileNetV2 performs well at 90.301\%, with Xception slightly behind at 89.151\%. ResNet50V2 shows the lowest accuracy at 87.726\%. The performance gap between models is less pronounced on this dataset compared to APTOS 2019, suggesting that the DDR dataset presents challenges that are handled somewhat similarly by the different architectures.

The IDRiD dataset shows generally lower accuracy scores across all models compared to other datasets. VGG19 achieves the highest accuracy at 81.250\%, followed by VGG16 at 79.688\%. MobileNetV2 reaches 78.125\%, while ResNet50V2 and Xception trail at 76.562\% and 73.438\% respectively. The consistent drop in performance across all models suggests that this dataset presents more challenging classification problems compared to the others.
On the Messidor 2 dataset, VGG19 leads with 91.286\% accuracy, followed by VGG16 at 90.871\%. The remaining models show a pronounced gap, with ResNet50V2 at 88.797\%, MobileNetV2 at 87.552\%, and Xception at 87.344\%. This dataset again demonstrates the strength of the VGG architecture, particularly VGG19, which outperforms other architectures by approximately 2-4\%.Retino dataset shows the best overall performance, VGG19 is the highest with 93.923\%, followed by VGG16 with 92.265\%. Xception is 88.398\%, and both ResNet50V2 and MobileNetV2 are 87.845\%. VGG models are way ahead of the others by 4-6\%.

Across all datasets, VGG19 is the best, followed by VGG16. The newer architectures like ResNet50V2, MobileNetV2, Xception are not as good as expected despite their advantages in other domains.\Cref{Individual_Dataset_Loss_and_Accuracy_Curve} and \Cref{Individual_Dataset_Confusion_Matrix} are loss and accuracy curves and confusion matrices for training and validation data across multiple datasets used for diabetic retinopathy classification. Each row is for a specific dataset, APTOS 2019, DDR, IDRiD, Messidor 2, Retino, with best two models for that dataset. Blue curve is training, orange curve is validation. These curves show how well the models converge during training, difference in learning stability, overfitting and generalization across datasets. At the same time, each confusion matrix shows the number of correct and incorrect instances across different diabetic retinopathy severity levels. This helps to compare the performance of different models on different datasets and understand their strengths and weaknesses in detecting diabetic retinopathy.

\subsection{Impact of Hybrid dataset on Model Performance}

Analyzing the performance results presented in \Cref{tab:Table1} for the hybrid dataset which integrates all the individual datasets reveals valuable insights into the generalizability and robustness of deep learning models in diagnosing diabetic retinopathy using a fixed batch size of 128, a learning rate of 1e-5 and 60 training epochs across diverse imaging conditions. The hybrid dataset combines different imaging standards, resolutions and qualities from multiple sources to test model adaptability and overcome dataset specific biases and limitations. Training models on a comprehensive hybrid dataset encourages the development of generalized features and the resulting models are highly applicable to real world clinical scenarios. When looking at individual models for the hybrid dataset, VGG19 is performing very strong and consistent, with the highest accuracy of 90.935\%. VGG16 is close second with 90.473\% reflecting the strength of VGG architectures in learning features across different datasets. ResNet50V2 is slightly lower at 90.153\% indicating it has strong generalization but slightly less than VGG models, there is room to tune feature extraction. MobileNetV2 although valued for its efficiency and lightweight is moderate at 90.082\% indicating challenges in capturing diverse and complex features in the large hybrid dataset. Xception are lower at 88.482\% showing their limitations in generalizing across the dataset’s heterogeneous imaging conditions.

\begin{figure}[htbp]
    \centering
    
    \begin{subfigure}[b]{0.48\textwidth}
        \centering
        \includegraphics[width=\textwidth, keepaspectratio]{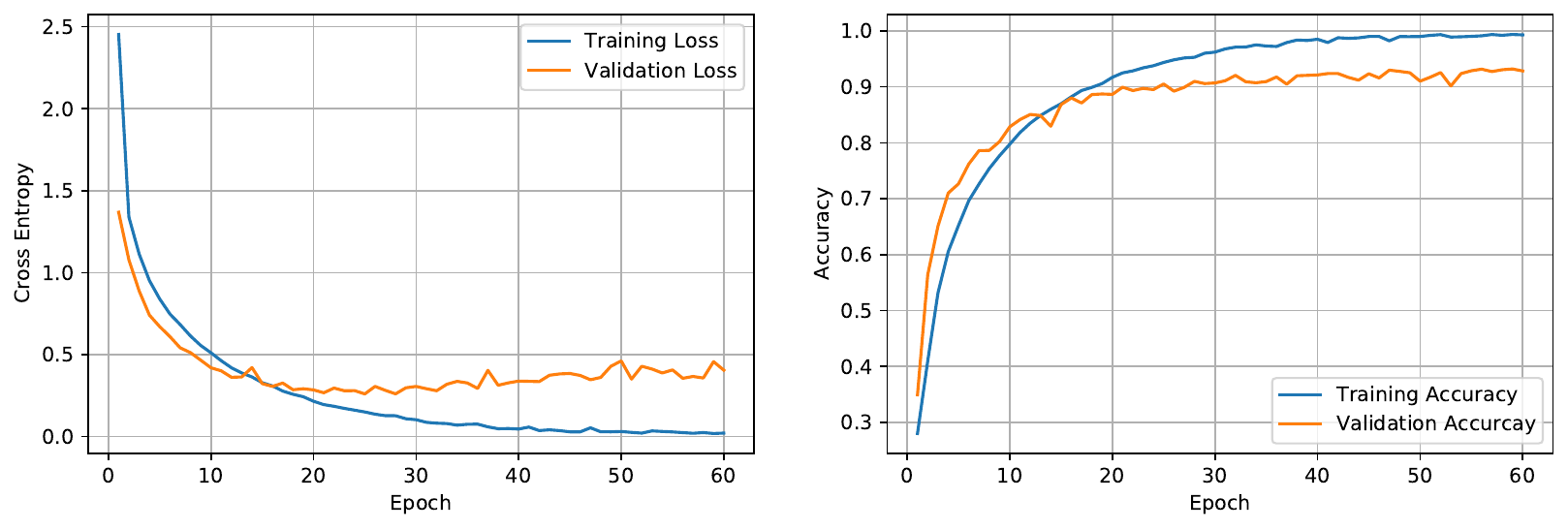} 
        \caption{VGG16}
    \end{subfigure}
    \hspace{0.02\textwidth} 
    \begin{subfigure}[b]{0.48\textwidth}
        \centering
        \includegraphics[width=\textwidth, keepaspectratio]{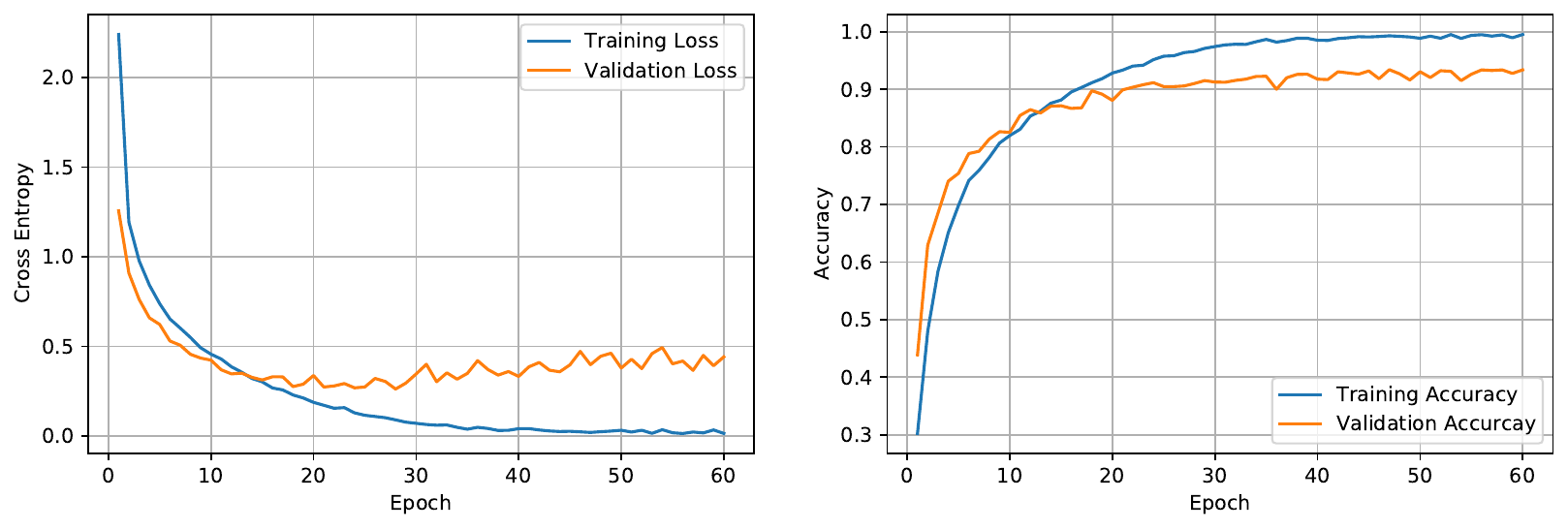} 
        \caption{VGG19}
    \end{subfigure}
    
    \begin{subfigure}[b]{0.48\textwidth}
        \centering
        \includegraphics[width=\textwidth, keepaspectratio]{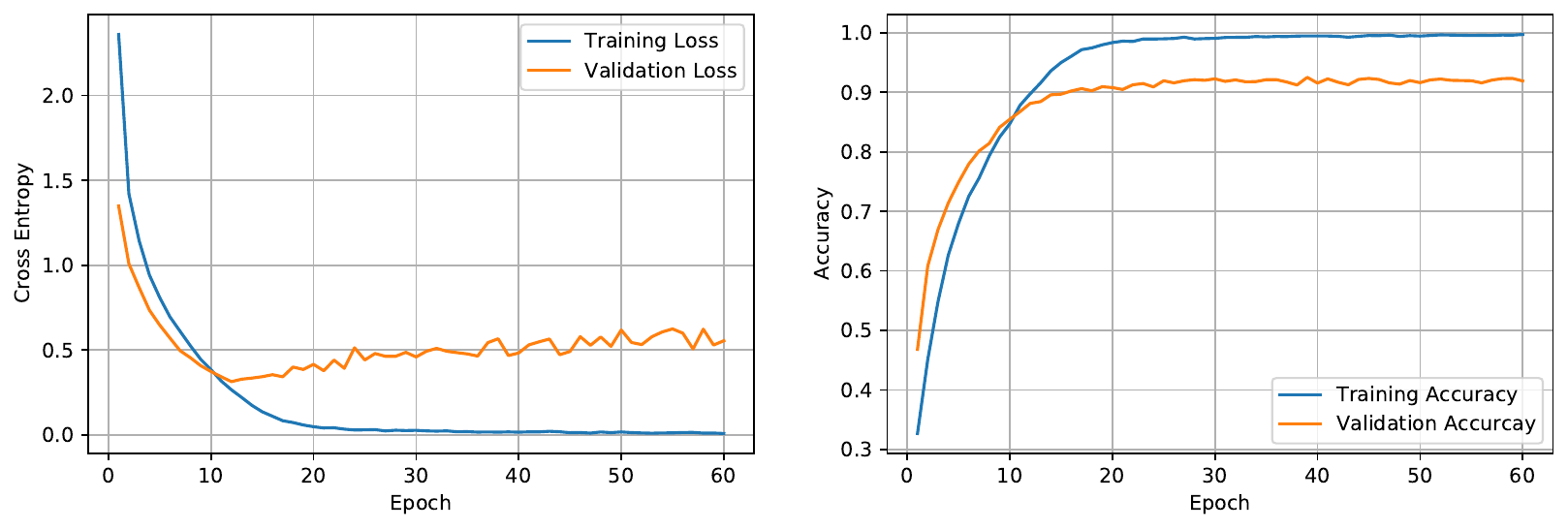} 
        \caption{ResNet50V2}
    \end{subfigure}
    \hspace{0.02\textwidth} 
    \begin{subfigure}[b]{0.48\textwidth}
        \centering
        \includegraphics[width=\textwidth, keepaspectratio]{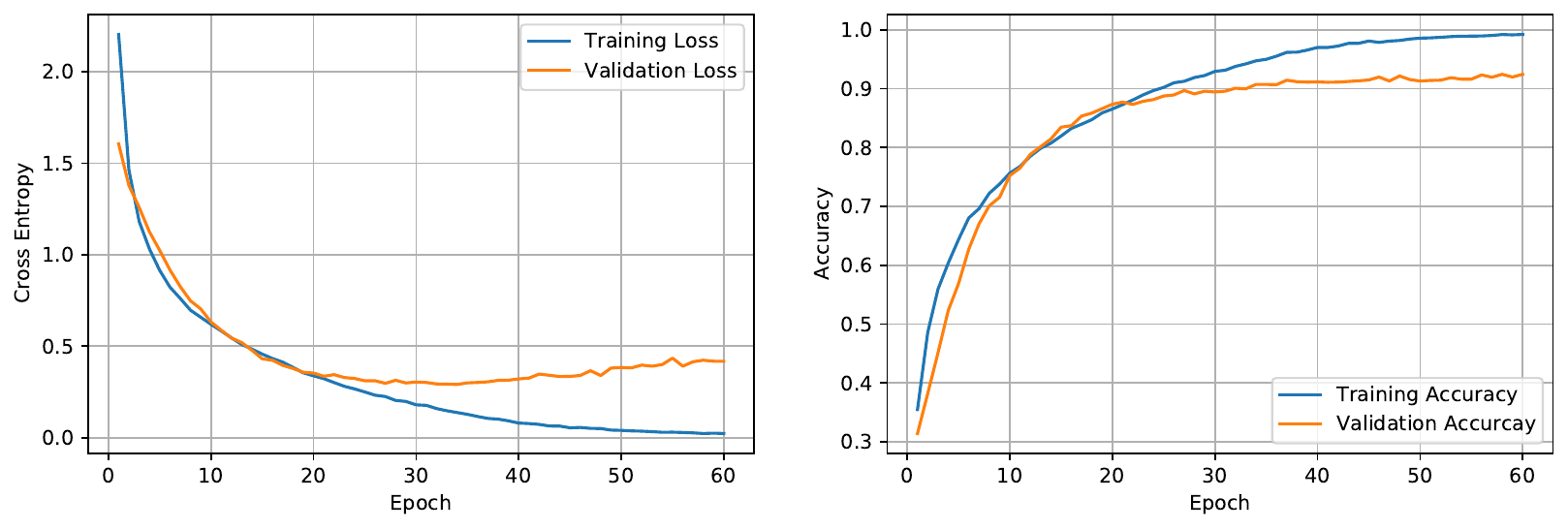} 
        \caption{MobileNetV2}
    \end{subfigure}

    \begin{subfigure}[b]{0.48\textwidth}
        \centering
        \includegraphics[width=\textwidth]{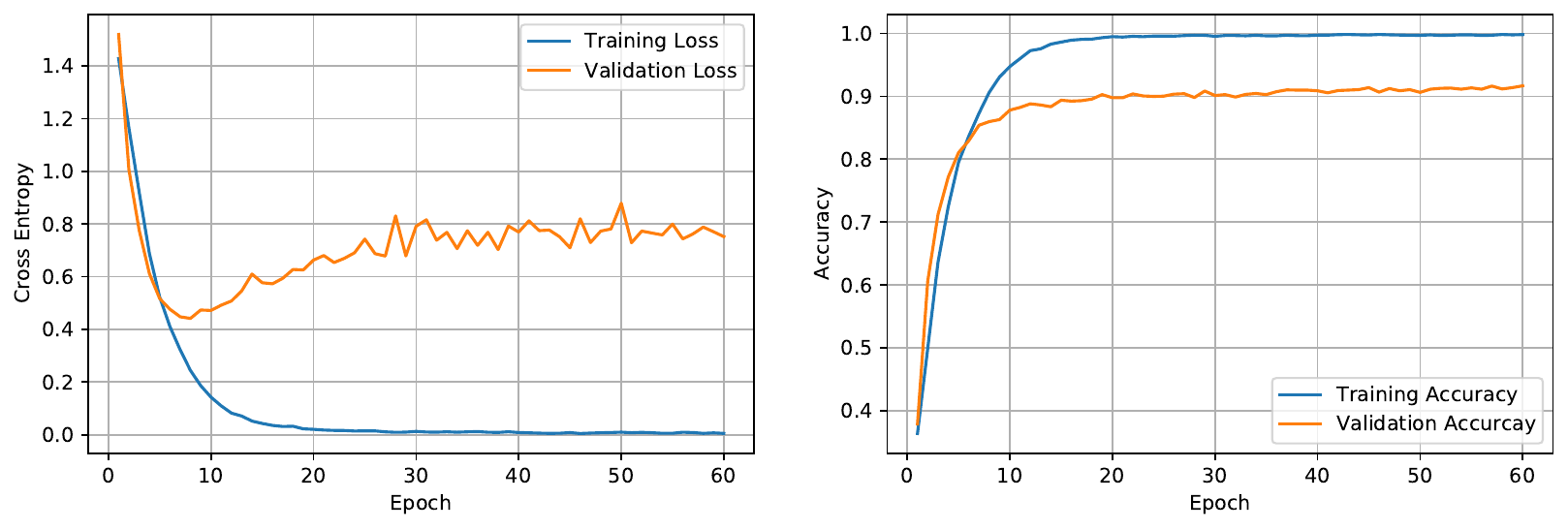} 
        \caption{Xception}
    \end{subfigure}
    \hspace{0.02\textwidth} 
    \begin{subfigure}[b]{0.48\textwidth}
        \centering
        \includegraphics[width=\textwidth, keepaspectratio]{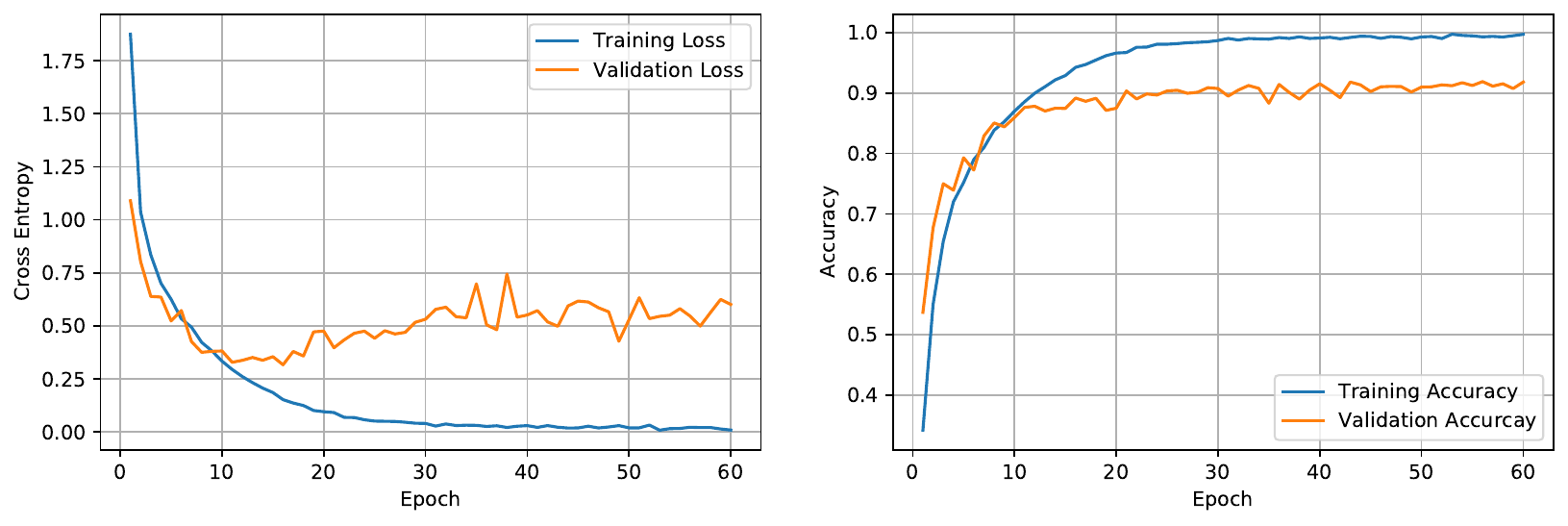} 
        \caption{VR-FuseNet}
    \end{subfigure}
 
    \caption{Hybrid Dataset Loss vs Accuracy Graph}
    \label{Hybrid_Loss_Graph}
\end{figure}

\begin{figure}[htbp]
    \centering
    
    \begin{subfigure}[b]{0.3\textwidth}
        \centering
        \includegraphics[width=\textwidth, keepaspectratio, scale = 0.7]{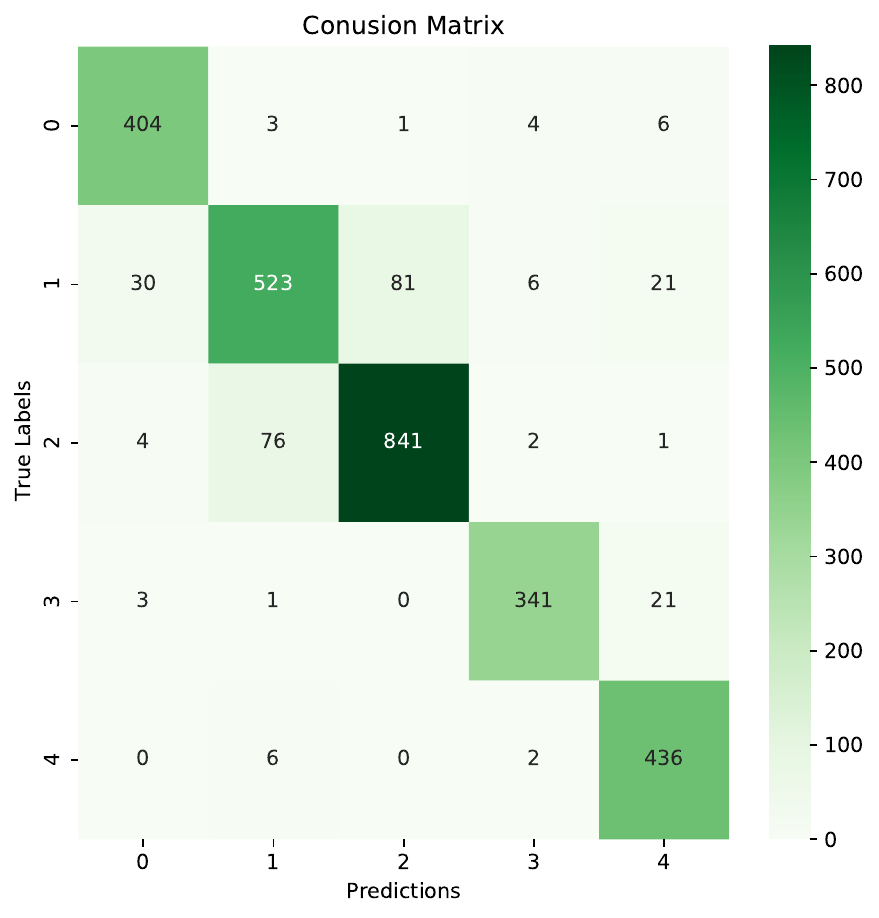} 
        \caption{VGG16}
    \end{subfigure}
    \hspace{0.02\textwidth} 
    \begin{subfigure}[b]{0.3\textwidth}
        \centering
        \includegraphics[width=\textwidth, keepaspectratio, scale = 0.7]{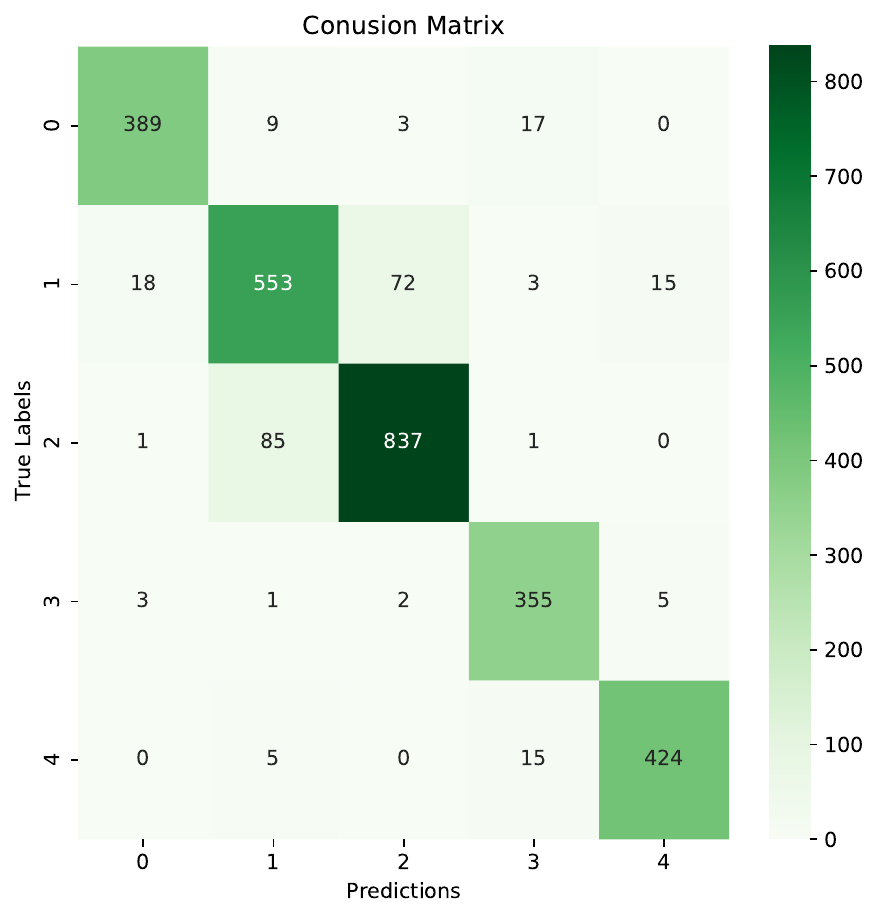} 
        \caption{VGG19}
    \end{subfigure}
    \hspace{0.02\textwidth} 
    \begin{subfigure}[b]{0.3\textwidth}
        \centering
        \includegraphics[width=\textwidth, keepaspectratio, scale = 0.7]{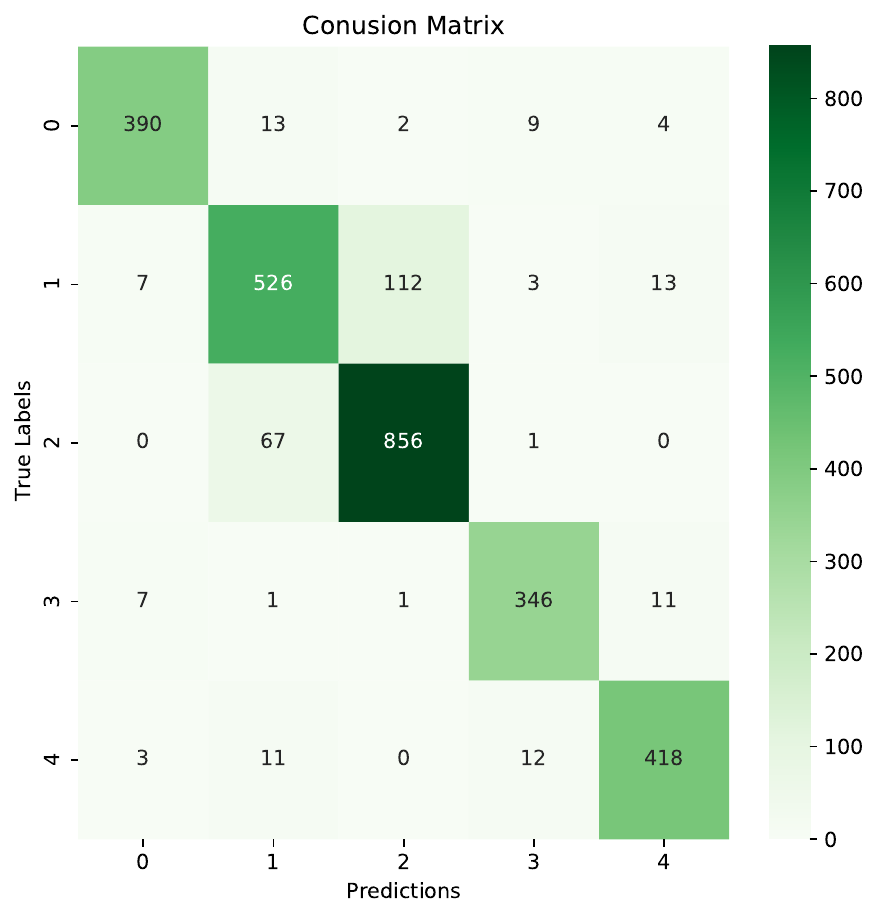} 
        \caption{ResNet50V2}
    \end{subfigure}
    \hspace{0.02\textwidth} 
    \begin{subfigure}[b]{0.3\textwidth}
        \centering
        \includegraphics[width=\textwidth, keepaspectratio, scale = 0.7]{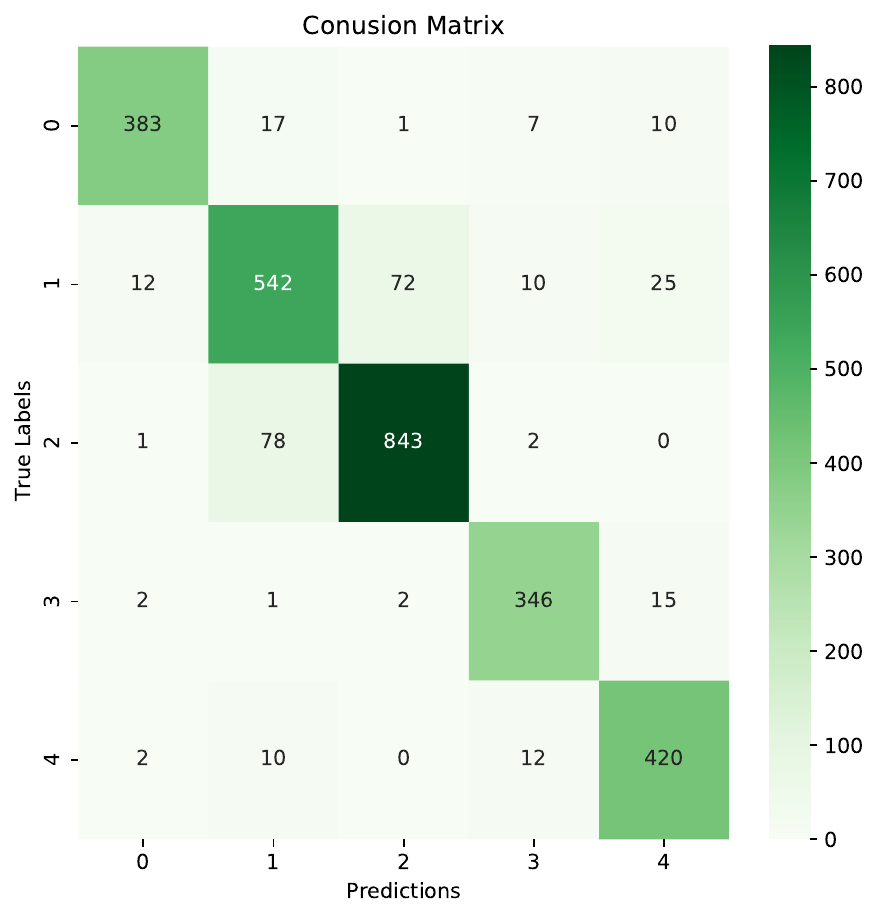} 
        \caption{MobileNetV2}
    \end{subfigure}
    \hspace{0.02\textwidth} 
    \begin{subfigure}[b]{0.3\textwidth}
        \centering
        \includegraphics[width=\textwidth, keepaspectratio, scale = 0.7]{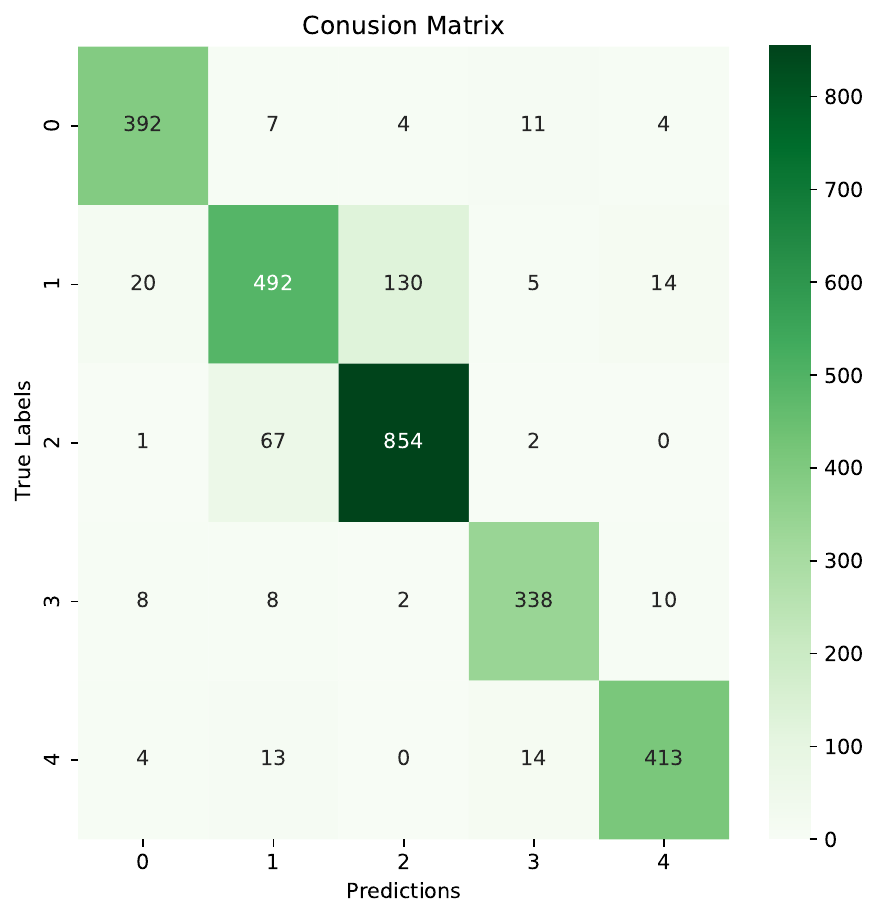} 
        \caption{Xception}
    \end{subfigure}
    \hspace{0.02\textwidth} 
    \begin{subfigure}[b]{0.3\textwidth}
        \centering
        \includegraphics[width=\textwidth, keepaspectratio]{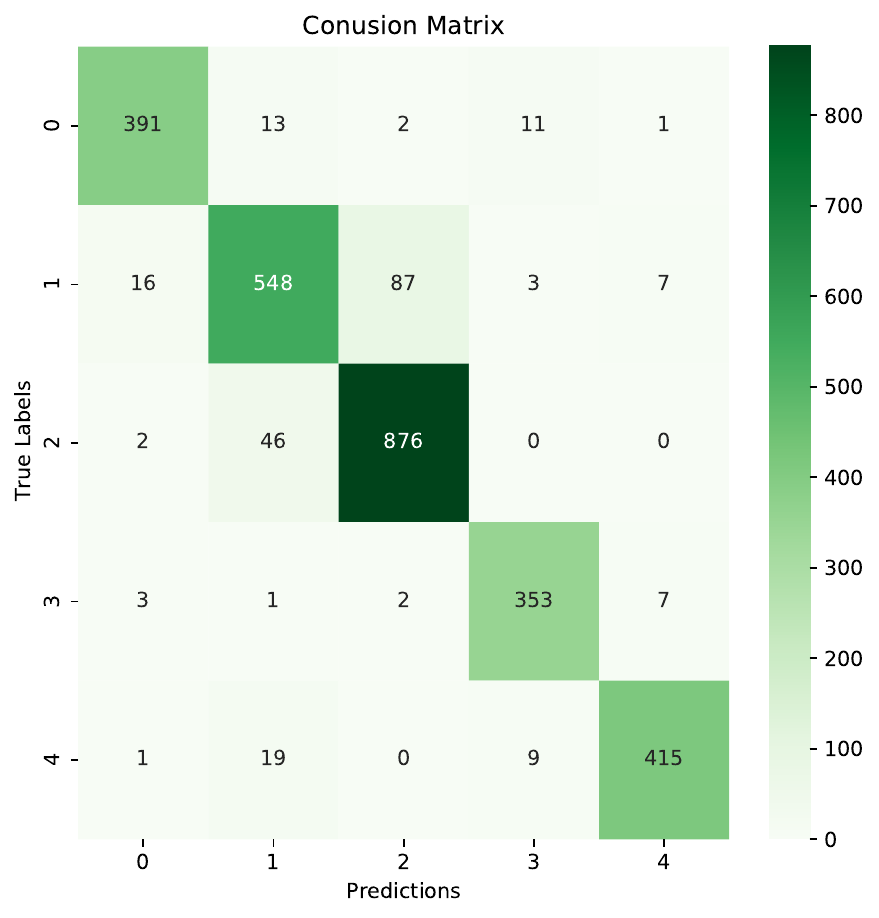} 
        \caption{VR-FuseNet}
    \end{subfigure}
    
    \caption{Hybrid Dataset Confusion Matrix}
    \label{Hybrid_Confusion_Matrix}
\end{figure}

\begin{figure}[htbp]
    \centering
    
    \begin{subfigure}[b]{0.3\textwidth}
        \centering
        \includegraphics[width=\textwidth, keepaspectratio, scale = 0.7]{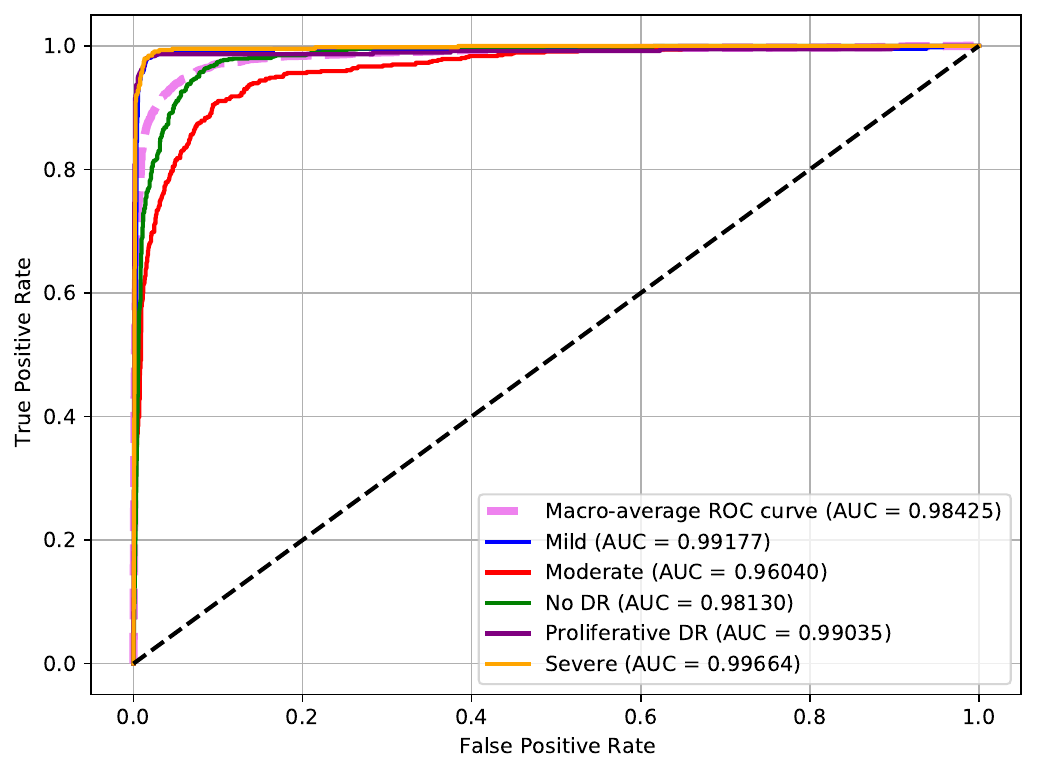} 
        \caption{VGG16}
    \end{subfigure}
    \hspace{0.02\textwidth} 
    \begin{subfigure}[b]{0.3\textwidth}
        \centering
        \includegraphics[width=\textwidth, keepaspectratio, scale = 0.7]{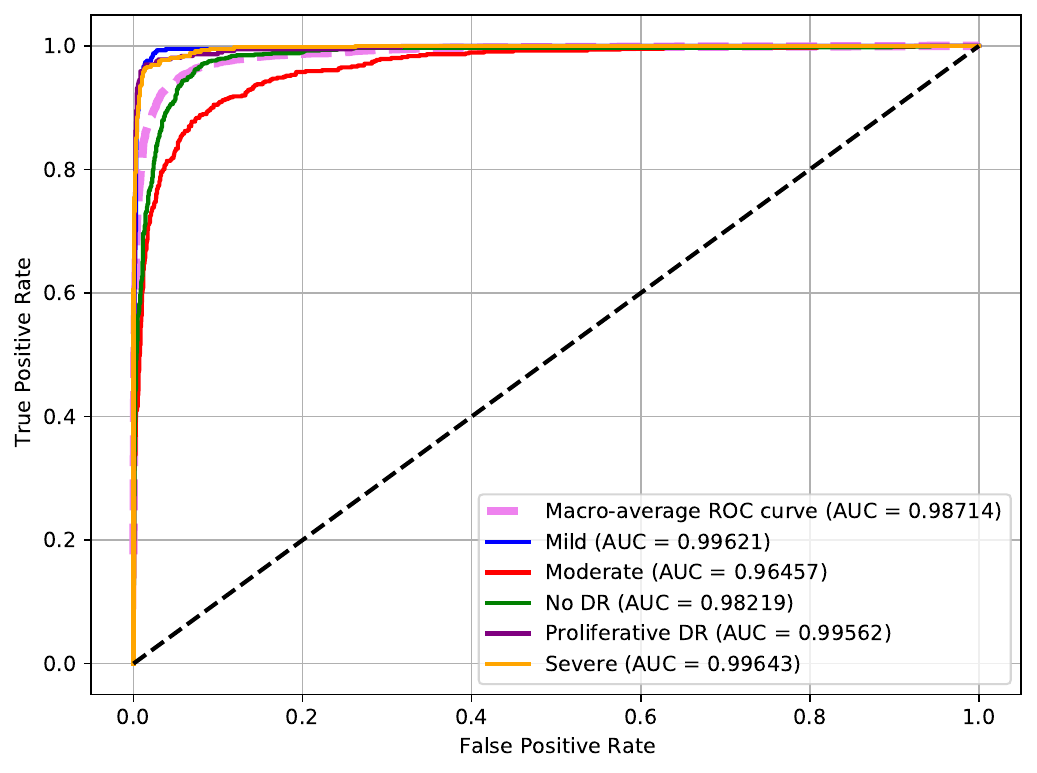} 
        \caption{VGG19}
    \end{subfigure}
    \hspace{0.02\textwidth} 
    \begin{subfigure}[b]{0.3\textwidth}
        \centering
        \includegraphics[width=\textwidth, keepaspectratio, scale = 0.7]{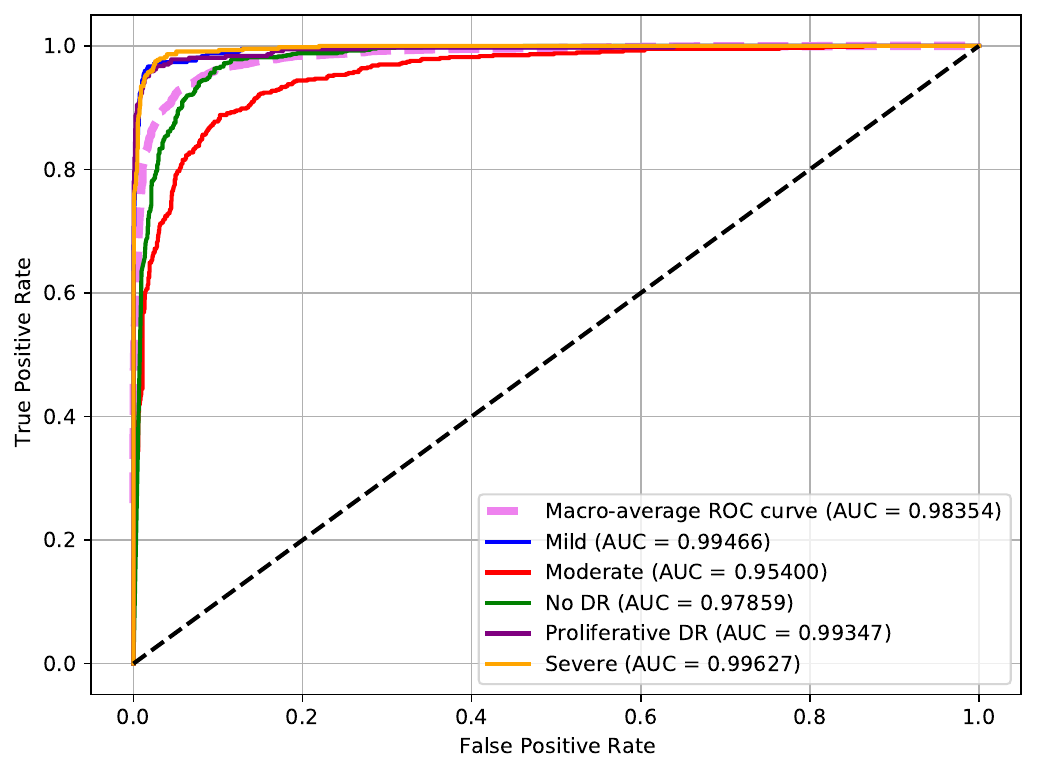} 
        \caption{ResNet50V2}
    \end{subfigure}
    \hspace{0.02\textwidth} 
    \begin{subfigure}[b]{0.3\textwidth}
        \centering
        \includegraphics[width=\textwidth, keepaspectratio, scale = 0.7]{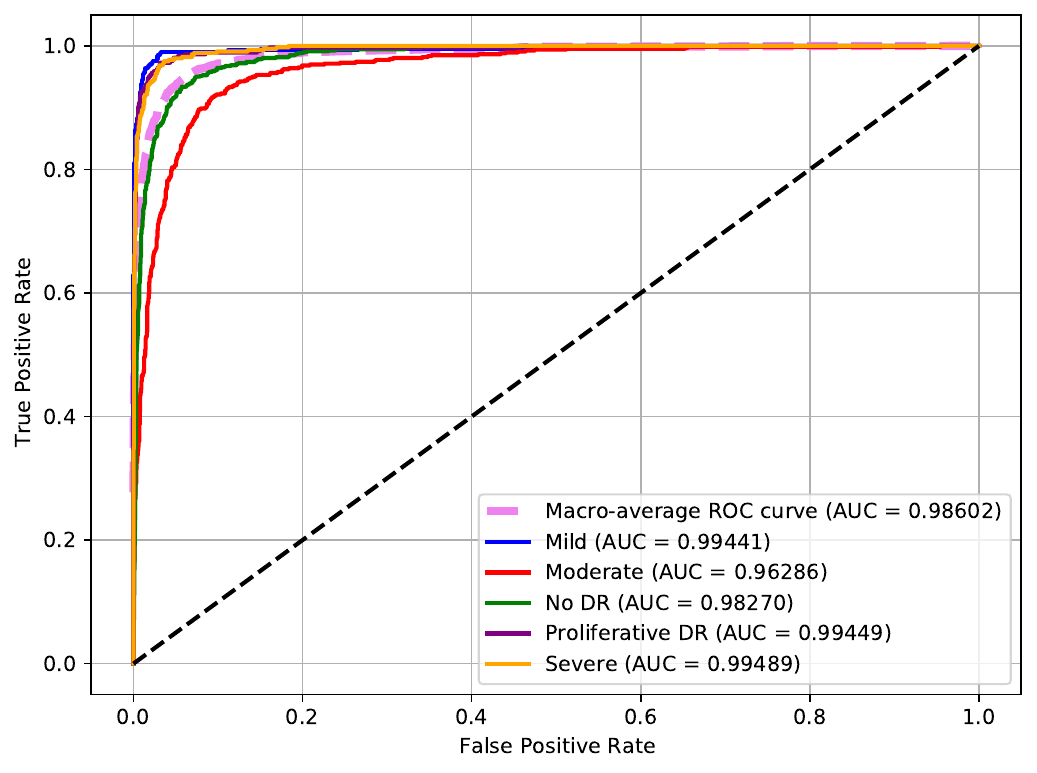} 
        \caption{MobileNetV2}
    \end{subfigure}
    \hspace{0.02\textwidth} 
    \begin{subfigure}[b]{0.3\textwidth}
        \centering
        \includegraphics[width=\textwidth, keepaspectratio, scale = 0.7]{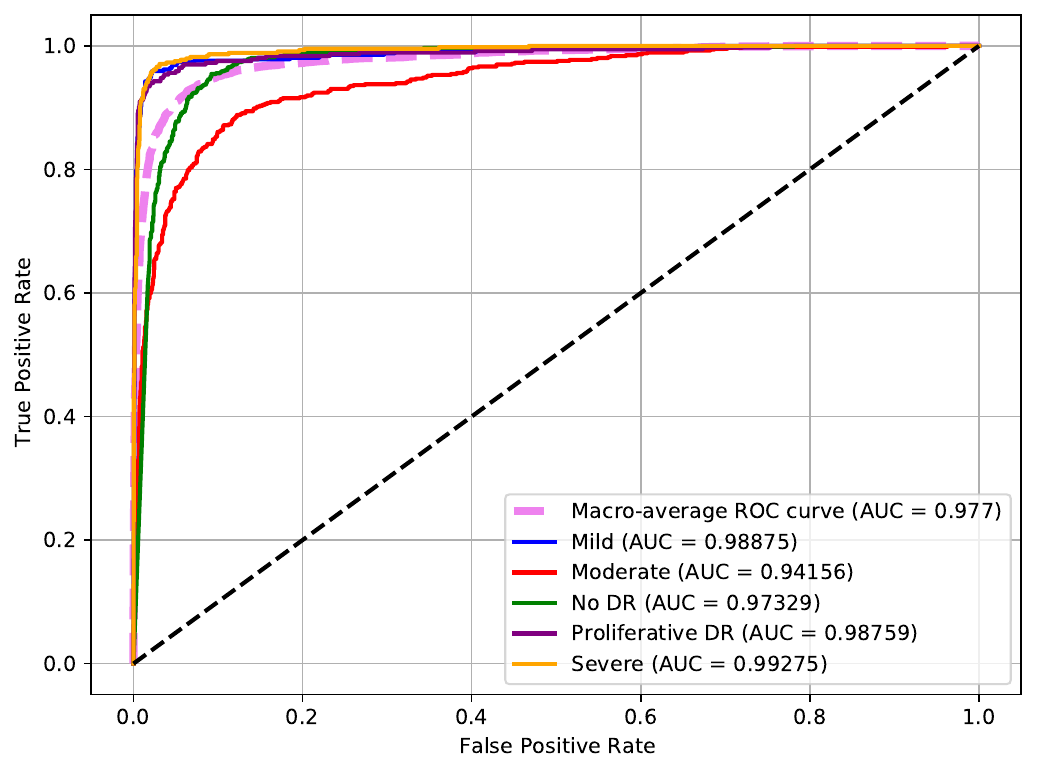} 
        \caption{Xception}
    \end{subfigure}
    \hspace{0.02\textwidth} 
    \begin{subfigure}[b]{0.3\textwidth}
        \centering
        \includegraphics[width=\textwidth, keepaspectratio]{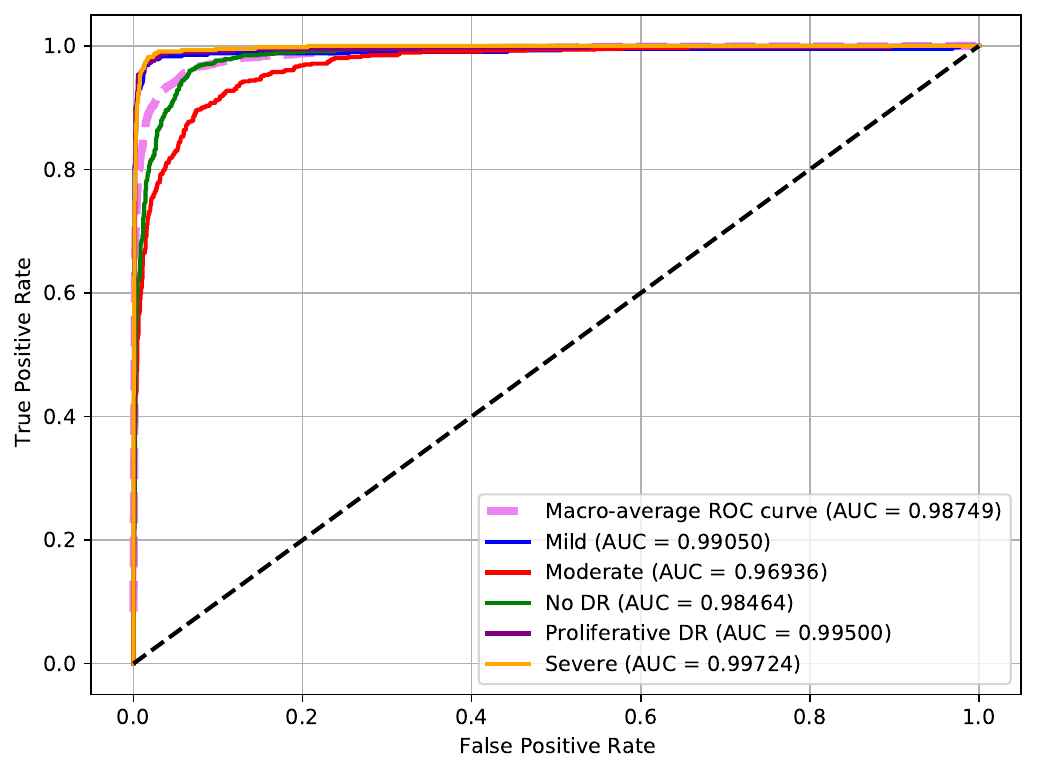} 
        \caption{VR-FuseNet}
    \end{subfigure}
    
    \caption{Hybrid Dataset ROC Curve}
    \label{Hybrid_ROC_Curve}
\end{figure}

In contrast, the proposed VR-FuseNet model, a combination of VGG19 and ResNet50V2, performs better on the hybrid dataset with an accuracy of 91.824\%, precision of 92.612\%, recall of 92.233\%, F1-score of 92.392\% and AUC of 98.749\%. VR-FuseNet combines the deep hierarchical representation of ResNet50V2 with the detailed and fine-grained spatial features of VGG19. This combination enables VR-FuseNet to handle the diversity in the hybrid dataset and outperforms individual models in all metrics. The combination in VR-FuseNet addresses the complexity and heterogeneity of the hybrid dataset and leverages the strengths of each architecture to demonstrate great generalization. Overall the hybrid dataset and VR-FuseNet model is an approach to highly accurate, reliable and generalized diabetic retinopathy detection, suitable and applicable in different imaging conditions and patient population.

\begin{table}[htbp]
\caption{Performance Metrics for Hybrid Dataset Models}
\label{tab:Table1}
\centering
\renewcommand{\arraystretch}{1.3}
\begin{tabular}{|c|c|c|c|c|c|}
\hline
\textbf{Model} & \textbf{Accuracy (\%)} & \textbf{Precision (\%)} & \textbf{Recall (\%)} & \textbf{F1-score (\%)} & \textbf{AUC (\%)} \\ \hline
VGG16 & 90.473 & 90.912 & 91.632 & 91.189 & 98.425 \\ \hline
VGG19 & 90.935 & 91.439 & 91.960 & 91.677 & 98.714 \\ \hline
ResNet50V2 & 90.153 & 91.215 & 90.840 & 90.993 & 98.354 \\ \hline
MobileNetV2 & 90.082 & 90.472 & 90.798 & 90.605 & 98.602 \\ \hline
Xception & 88.482 & 89.463 & 89.201 & 89.253 & 97.700 \\ \hline
\textbf{\begin{tabular}[c]{@{}c@{}}VR-FuseNet\\ (Proposed)\end{tabular}} & \textbf{91.824} & \textbf{92.612} & \textbf{92.233} & \textbf{92.392} & \textbf{98.749} \\ \hline
\end{tabular}
\end{table}

\par \Cref{Hybrid_Loss_Graph} and \Cref{Hybrid_Confusion_Matrix} Displays the training and validation accuracy and loss curves and confusion matrices for various deep learning models. Each subplot shows two graphs: one for loss (indicating how well the model is minimizing errors) and one for accuracy (indicating model performance over training epochs). The blue curves represent training data, while the orange curves represent validation data. The comparison of models reveals differences in learning rate, consistency, and ability to generalize effectively. Some models show fast convergence, while others require more epochs to stabilize. Certain models exhibit overfitting, where the training accuracy is significantly higher than validation accuracy, accompanied by fluctuations in validation loss.
On the other hand, each confusion matrix visualizes the model’s performance by displaying the number of correct and misclassified instances. At the bottom, the proposed VR-FuseNet model is evaluated using the hybrid dataset. This comparison provides insights into how different models classify various severity levels and highlights their advantages and limitations in diabetic retinopathy detection.

Also \Cref{Hybrid_ROC_Curve} shows Receiver Operating Characteristic (ROC) curves for different deep learning models trained on the hybrid dataset. Each ROC curve plots true positive rate (sensitivity) vs false positive rate (1-specificity) to evaluate the performance of different models. The dashed diagonal line is random classification, the closer to top left corner is better. The AUC (Area Under the Curve) values displayed in each figure is the performance, higher is better. The proposed VR-FuseNet aims to improve the accuracy. By comparing different models we can see how well they can differentiate between different diabetic retinopathy severity levels and what are their strengths and weaknesses.

\section{Explainable Artificial Intelligence (XAI)}
\label{Explainable Artificial Intelligence}
Explainable Artificial Intelligence (XAI) is a research area that makes the decision making process of artificial intelligence (AI) systems, especially deep learning models, more interpretable and transparent to users. Traditional deep neural networks (DNNs) are \textbf{\textit{black boxes}}, they make good predictions but provide no insight into how those decisions are made, which is a problem in high stakes applications like healthcare, finance and autonomous systems \cite{adadi2018peeking, faria2024explainable}. XAI techniques like Gradient-weighted Class Activation Mapping (Grad-CAM) \cite{selvaraju2017grad} help to address this by generating visual explanations that highlight the most relevant input features that contribute to the model’s output. In our work we have used five gradient based XAI methods to increase model explainability and get a deeper understanding of the decision making process. By providing human understandable insights XAI makes the model more transparent, helps in debugging and ensures fairness so stakeholders can trust and refine AI driven systems.

\subsection{Gradient-weighted Class Activation Mapping (Grad-CAM)}
Grad-CAM is a popular explainability technique to give visual explanations for the decision making process of deep convolutional neural networks (CNNs). Grad-CAM achieves this by using the gradient information flowing into the last convolutional layer to generate a class-discriminative localization map, highlighting the regions in the input image that contribute most to the model’s prediction. Unlike earlier visualization methods like Class Activation Mapping (CAM) which required architectural changes and retraining, Grad-CAM can be applied to all CNN architectures without any changes \cite{selvaraju2017grad}. The technique computes the gradients of the target class score $y^c$ (before the softmax) w.r.t the feature maps $A^k$ of the last convolutional layer to capture spatial importance. The gradient based importance weights $\alpha_k^c$ for each feature map $A^k$ is computed using global average pooling of the gradients as follows:
\begin{equation}
\alpha_k^c = \frac{1}{Z} \sum_i \sum_j \frac{\partial y^c}{\partial A^k_{ij}}
\end{equation}
where $Z$ is the total number of pixels in the feature map. These weights represent the importance of each feature map to the model’s decision for class $c$. The final Grad-CAM heatmap $L_{\text{Grad-CAM}}^c$ is obtained by combining the activation maps weighted by these scores and then applying ReLU to retain only positive contributions:
\begin{equation}
L_{\text{Grad-CAM}}^c = \text{ReLU} \left( \sum_k \alpha_k^c A^k \right)
\end{equation}
By applying ReLU, the method focuses on features that have a positive impact on the class of interest and ignore the negative ones. The resulting heatmap shows the most important regions in the input image for the given prediction. Grad-CAM is very useful in medical imaging such as diabetic retinopathy detection as it allows clinicians to interpret the model predictions by visualizing the regions of interest in retinal fundus images, to identify the pathological features such as microaneurysms, hemorrhages and exudates. To make it even more interpretable, Visual Question Answering (VQA) is integrated in Explainable AI (XAI) with Grad-CAM so humans can interact with AI models like they do with each other. VQA adds an extra layer of explainability by allowing users to ask natural language questions about model predictions and get answers with visual explanations. In medical imaging, VQA models use a mix of image features from CNNs and text features from transformers to provide meaningful answers. Given an image $I$ and a text question $T$, the VQA model extracts image features $V$ and text embeddings $Q$ as:
\begin{equation}
V = f_{img}(I), \quad Q = f_{text}(T)
\end{equation}

where $f_{img}$ is a vision model (e.g., ResNet, Vision Transformers) and $f_{text}$ is a text model (e.g., BERT, LLaMA). The answer $A$ is predicted as:

\begin{equation}
A = \arg\max p(A | V, Q)
\end{equation}
To integrate VQA with Grad-CAM, the model produces a heatmap-based explanation along with the text answer so the answer is backed by a highlighted region of interest in the image. For example, if a doctor asks, \textit{What features indicate diabetic retinopathy?}, the system not only gives a text answer but also overlays a Grad-CAM heatmap over the retinal image to show lesions, hemorrhages or exudates. By combining VQA with Grad-CAM, this makes deep learning models more transparent, interactive and interpretable in applications like medical diagnostics, autonomous decision making and security.

\subsection{Gradient-weighted Class Activation Mapping++ (Grad-CAM++)}
Grad-CAM++ is an enhanced version of Grad-CAM that improves the localisation accuracy and the detection of multiple instances of a class in an image. While Grad-CAM provides a class-discriminative heatmap by computing the gradients of the target class score with respect to the last convolutional layer feature maps, it has limitations in localizing object boundaries and identifying multiple objects of the same class \cite{chattopadhay2018grad}. Grad-CAM++ addresses these issues by reweighting the activation maps based on higher order derivatives so that the generated heatmaps capture finer spatial details. The formulation of Grad-CAM++ builds upon Grad-CAM but with a more refined weighting mechanism for the activation maps. Instead of using a global average pooling of the gradients, Grad-CAM++ uses the second and third order gradients to compute the importance of each feature map $A^k$ so that better object localisation is achieved. The class specific importance weight $\alpha_k^c$ is given by:
\begin{equation}
\alpha_k^c = \sum_i \sum_j \alpha^c_{ij} \cdot \text{ReLU} \left( \frac{\partial y^c}{\partial A^k_{ij}} \right)
\end{equation}

where $\alpha^c_{ij}$ is derived using the second-order and third-order derivatives:

\begin{equation}
\alpha^c_{ij} = \frac{\partial^2 y^c / \partial A_{ij}^{k2}}{2 (\partial^2 y^c / \partial A_{ij}^{k2}) + \sum_a \sum_b A_{ab}^k (\partial^3 y^c / \partial A_{ab}^{k3})}
\end{equation}

This ensures regions of the image that contribute most to the class prediction get more attention and irrelevant activations get suppressed. The final Grad-CAM++ heatmap $L_{\text{Grad-CAM}++}^c$ is calculated as:
\begin{equation}
L_{\text{Grad-CAM}++}^c = \text{ReLU} \left( \sum_k \alpha_k^c A^k \right)
\end{equation}
The main advantage of Grad-CAM++ over Grad-CAM is that it assigns different importance to different pixels in the same activation map, so it’s better at localizing multiple objects of the same class and capturing more details in the image. This is very useful for medical imaging tasks where you need to identify multiple lesions. By generating more precise and interpretable heatmaps, Grad-CAM++ increases model transparency and trust in deep learning based medical diagnosis systems.

\subsection{Fine-Grained Class Activation Mapping (Layer-CAM)}
Layer-CAM is an advanced class activation mapping (CAM) method that deepens the understanding of convolutional neural networks (CNNs) by generating fine-grained object localization maps across multiple layers. Unlike traditional methods like Grad-CAM which only rely on the final convolutional layer and produce coarse localization maps, Layer-CAM computes activation maps from both deep and shallow layers so it balances coarse spatial localization and fine-grained object details \cite{jiang2021layercam}. This is very useful for image classification tasks like retinal fundus image analysis where precise localization of critical regions is important. Layer-CAM works by computing activation maps from different layers of the CNN, considering the impact of each spatial location on the class score. Formally, the weight of each spatial location (i,j) in the k-th feature map Ak is defined as:
\begin{equation}
w_{i j}^{c k} = \text{ReLU} \left( \frac{\partial y^c}{\partial A_{i j}^k} \right)
\end{equation}
where $y^c$ is the predicted score of the target class $c$,, and $\frac{\partial y^c}{\partial A_{i j}^k}$ denotes the gradient of the target class score with respect to the feature map $A_k$. The class activation map $M^c$ is then computed as:
\begin{equation}
M^c = \text{ReLU} \left( \sum_k w_{i j}^{c k} \cdot A_{i j}^k \right)
\end{equation}
And a more general representation of Layer-CAM is:
\begin{equation}
S_c(x) = \text{ReLU} \sum_{c=1}^{C} \left( W_c \times \text{Score}(c, F(x)) \right)
\end{equation}
where $F(x)$ represents the output of the final convolutional layer, $W_c$ denotes the weight for each channel, and $\text{Score}(c, F(x))$ computes the contribution of each channel to the class score. By utilizing location-specific gradient weights rather than global channel-wise weights, Layer-CAM effectively highlights the most relevant image regions while filtering out irrelevant background information. Its ability to generate class activation maps from different CNN layers provides complementary insights, resulting in more precise and integral class-specific object regions. Furthermore, its ease of integration with off-the-shelf CNN-based image classifiers, without requiring architectural modifications or changes in back-propagation, makes it a versatile tool for various image analysis tasks, including weakly-supervised object localization, semantic segmentation, and medical image classification.

\subsection{Score-weighted Class Activation Mapping (Score-CAM)}
Score-CAM  is a way to show the parts of an input image that matter most for a neural network’s classification output. It builds upon Class Activation Mapping (CAM) and Grad-CAM and adds class specific information while not using gradients to compute importance of activations. Unlike Grad-CAM which uses the gradients of the target class score with respect to feature maps, Score-CAM derives the importance of activation maps based on their contribution to the model’s output score through forward pass. This gives more stable and noise free visual explanations especially in complex neural network architectures and improved class discriminative localization \cite{wang2020score}. In the Score-CAM method the input image goes through the network and produces final convolutional feature maps. These activation maps are extracted and upsampled to the size of the input image and used as spatial masks. Each activation map is used to generate modified inputs that highlight different regions and then forward passed through the model. The importance of each feature map is determined by how much it contributed to the model’s confidence score for a specific class. The class-specific importance score $\alpha_k^c$ for each feature map is defined as:
\begin{equation}
\alpha_k^c = f_c(X \circ s(Up(A^k))) - f_c(X_b)
\end{equation}
where $f_c(\cdot)$ is the model’s confidence score for class $c$, $X$ is the original image, $X_b$ is the baseline image (e.g., black or zero-input), $Up(A^k)$ denotes the upsampled activation map, and $s(\cdot)$ is a normalization function mapping values to $[0,1]$. The class scores are computed by linearly combining CAM values with weights learned from the fully connected layer. Unlike traditional gradient-based methods, Score-CAM doesn’t backpropagate gradients, so no gradient saturation and vanishing. After computing the contribution scores, a softmax is applied to rescale the scores so they sum to one. The final Score-CAM heatmap $L_{\text{Score-CAM}}^c$ is computed by combining the activation maps with their importance scores and then applying a ReLU:

\begin{equation}
L_{\text{Score-CAM}}^c = \text{ReLU} \left( \sum_k \alpha_k^c A^k \right)
\end{equation}
Using forward-pass confidence scores instead of gradients, Score-CAM is better than Grad-CAM as it avoids gradient related issues and provides class discriminative and less noisy saliency maps. This is very important in medical imaging applications where precise localization of pathological features is crucial. Its ability to produce fine grain explanations and be robust across different CNN architectures makes it a must have tool for Explainable AI in medical diagnostics.

\begin{figure}[htpb]
    \centering
    \includegraphics[width=\linewidth]{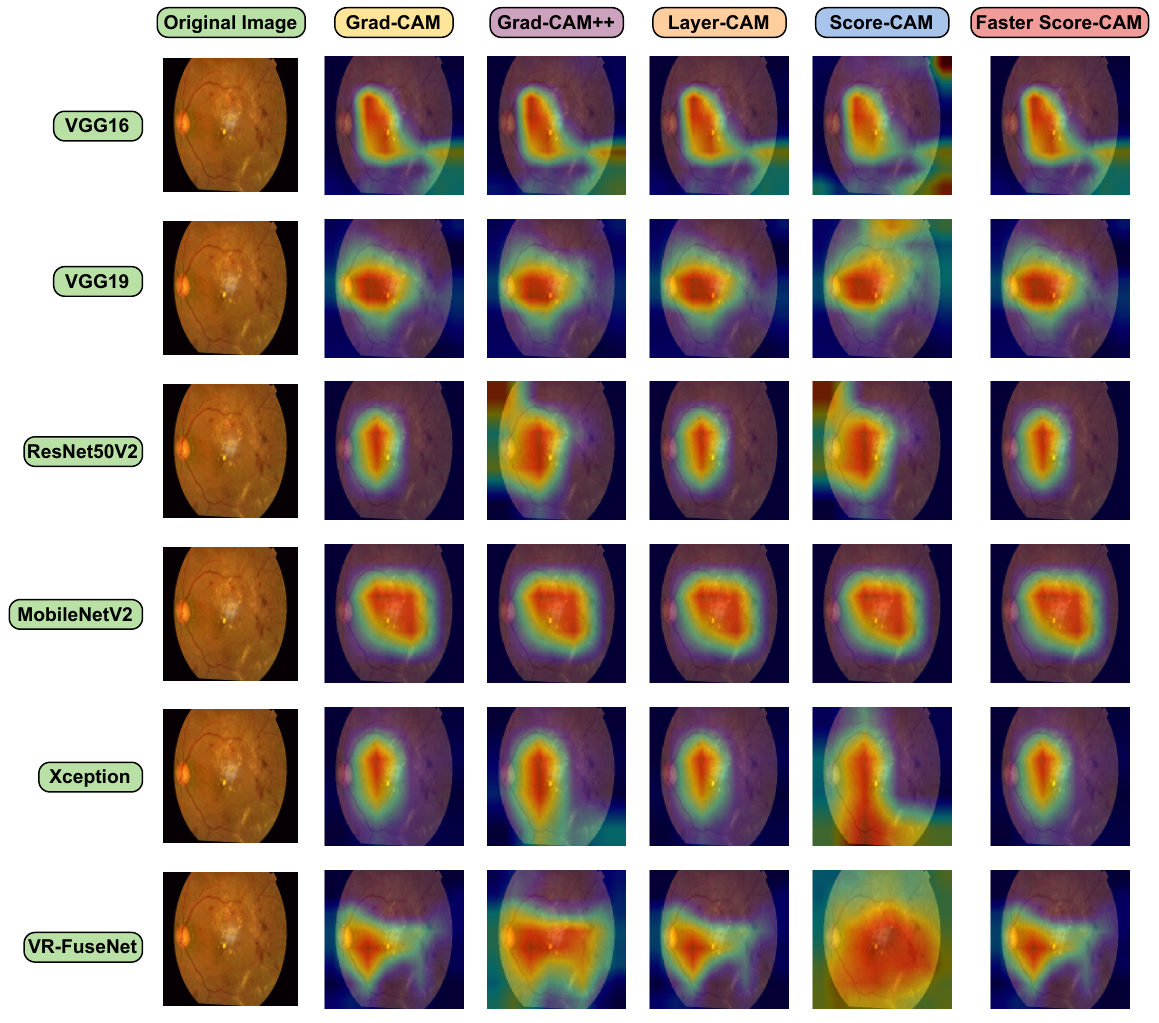}
    \caption{Comparative Analysis of Explainable AI Methods}
    \label{Figure XAI}
\end{figure}

\subsection{Optimized Score-weighted Class Activation Mapping (Faster Score-CAM)}
Faster Score-CAM is an optimization technique to improve the efficiency of convolutional neural networks (CNNs) in retinal fundus image classification. Traditional Score-CAM computes weights for all channels in a given feature map which can be computationally expensive \cite{li2023fimf}. To address this, Faster Score-CAM selects only the most informative channels by focusing on the ones with high variance, assuming they have more discriminative information. The process involves calculating the mean and variance for each channel, selecting the top $N$ channels with the highest variance and then use those channels to compute weights and generate a heatmap. This reduces the computational overhead while preserving the essential discriminative features making it more practical for large dataset analysis in medical imaging tasks like diabetic retinopathy and glaucoma detection. Mathematically the weight assigned to channel $k$ is given by:
\begin{equation}
\alpha_k = \frac{\text{Var}(A_k)}{\sum_{j=1}^{N} \text{Var}(A_j)}
\end{equation}
where $\text{Var}(A_k)$ is the variance of activations in channel $k$ and $N$ is the number of selected channels in the feature map. By reducing the number of channels Faster Score-CAM achieves a balance between computational efficiency and interpretability, the heatmap will highlight the relevant regions without requiring a lot of computational resources. However selecting the optimal value of $N$ is crucial, smaller $N$ may exclude useful features and larger $N$ may reintroduce redundancy. This tradeoff requires experimentation to maximize both speed and quality of the visual explanation especially in applications where model interpretability is critical like retinal disease classification

\subsubsection{Insights Derived from XAI Analysis}
\Cref{Figure XAI} shows the activation patterns of different Transfer Learning (TL) models for Diabetic Retinopathy using various XAI techniques. All models show consistent activation around the critical regions of the retina, especially around the lesions, microaneurysms or hemorrhages. Grad-CAM shows big and widespread activation across the retina, highlighting multiple affected areas but sometimes going beyond the most relevant regions. Grad-CAM++ generates more focused and precise heatmaps, concentrating on specific regions with higher pathological relevance. Layer-CAM shows broader activation but still focuses on the significant areas, so even peripheral signs of the disease are considered. Score-CAM spreads attention more evenly across the retinal surface, capturing general structural patterns while keeping a reasonable focus on disease-prone areas. Faster Score-CAM shows sharper and more confident activation in highly affected regions, pinpointing potential lesions with higher accuracy. Across the models, Grad-CAM and Grad-CAM++ highlight the key pathological regions, Layer-CAM has a wider but still targeted focus. Score-CAM and Faster Score-CAM are similar but have different focal points, Faster Score-CAM prioritizes the areas of higher confidence. The results show that despite the variations, the TL models recognize the same regions of interest, so they can consistently detect the signs of Diabetic Retinopathy. The variations in the heatmap distribution emphasize the importance of choosing the right XAI method to get the most reliable and interpretable insights from deep learning models in medical diagnosis.

\section{Limitations and Future Works}
\label{Limitations and Future Works}

The Kaggle EyePACS dataset is widely used in medical image classification tasks but has some challenges as mentioned in previous studies. These include large volume, noisy data and incorrect labeling which can affect the model’s accuracy and reliability \cite{tariq2022transfer}. Due to these limitations and the difficulty in annotating such a large dataset with accurate labels we didn’t use the EyePACS dataset in our research.

Despite the effectiveness of DR detection the model has some limitations to be addressed in future research. Computational cost is one of the main issues as the transformer model increases the number of parameters which leads to longer training and inference times. Due to these computational constraints we couldn’t implement Vision Transformers (ViTs) which have shown to be better in capturing long range dependencies and contextual information in medical imaging tasks. Future work will focus on integrating ViTs in the model architecture to leverage their ability to process global image features which could lead to better accuracy and interpretability of the model. However implementing ViTs will require optimized training strategies and computational resources which will be key moving forward.

Another limitation is imbalanced dataset which can affect the model’s ability to generalize well across different DR severity levels. Although the hybrid dataset combines multiple publicly available sources to increase diversity, some severity classes are underrepresented which can bias the model’s predictions. To address this future research will explore the use of Generative Adversarial Networks (GANs) to generate synthetic retinal images to balance the dataset and improve the model’s robustness. GANs are widely used in medical imaging to generate high quality synthetic data while maintaining clinical relevance. Their application in this case can reduce bias in the model’s predictions and improve its ability to generalize across different patient populations and imaging conditions.

Another area to improve is generalization to real world clinical data. The dataset used in this research is mainly composed of publicly available images which may not fully represent the variability seen in real world patient populations, different imaging devices or clinical settings. Domain adaptation techniques such as unsupervised domain adaptation will be explored in future work to make the model more adaptable to new environments. Adding more real world clinical images to the dataset and fine tuning the model to handle variations in lighting, resolution and noise levels will make the model more robust and reliable across different healthcare institutions.Finally, although the model is image based, combining multi-modal data such as patient demographics, clinical history and genetic factors could improve the accuracy. Many diseases including diabetic retinopathy are not just image based and incorporating additional clinical parameters could lead to more accurate diagnoses. Future work will explore methods to integrate image based deep learning models with structured clinical data to have a more complete AI driven diagnostic tool.

\section{Conclusion}
\label{Conclusion}
The VR-FuseNet model in this study does well in DR classification tasks by combining the strengths of two advanced convolutional neural network architectures, VGG19 and ResNet50V2. It uses VGG19’s ability to capture fine-grained spatial features and ResNet50V2’s deep hierarchical feature extraction. As a result, it scores high in accuracy, precision, recall and F1-score and outperforms individual models on the hybrid dataset of retinal images from various public sources. The careful fusion of these architectures helps the model to generalise across different clinical imaging conditions, hence more applicable in real-world healthcare settings. Moreover, the use of Explainable Artificial Intelligence (XAI) techniques, particularly gradient-based methods such as Grad-CAM, Grad-CAM++, Layer-CAM, Score-CAM and Faster Score-CAM adds value by providing interpretable visual explanations. These techniques allow medical professionals to see and understand which features in the retinal images contribute to the classification, improving model transparency, clinical trust and decision making. The model not only focuses on high accuracy but also ensures the predictions are interpretable and clinically meaningful, hence better validation and adoption by healthcare providers. However, there are still challenges to be addressed to fully utilise AI-driven DR detection systems. Specifically, computational complexity and dataset imbalance are big issues. Addressing these will be key in deploying these models in clinical settings. Future work should explore integrating Vision Transformers (ViTs) which have shown to capture long-range dependencies and global context in images. Also, using advanced data augmentation techniques including GANs can help balance the dataset and mitigate class imbalance issue, hence improve the model’s performance and robustness across different patient demographics and imaging modalities.

\bibliographystyle{unsrt}  
\bibliography{references.bib}

\end{document}